\documentclass[runningheads]{llncs}
\usepackage{graphicx}
\usepackage{tikz}
\usepackage{comment}
\usepackage{amsmath,amssymb} % define this before the line numbering.
\usepackage{color}
\usepackage{cite}
\usepackage{xspace}
\usepackage{rotating}
\usepackage{booktabs}
\usepackage{pifont}
\usepackage[abs]{overpic}
\usepackage[pagebackref=true,breaklinks=true,letterpaper=true,colorlinks,bookmarks=false]{hyperref}
\usepackage[accsupp]{axessibility}  % Improves PDF readability for those with disabilities.
\usepackage{caption}
\usepackage{subcaption}
\usepackage{soul}
\usepackage{colortbl}
\usepackage{microtype}
\newcommand{\ourmethod}{CoSMix\xspace}
\newcommand{\ie}{\textit{i.e.}\xspace}
\newcommand{\CC}[1]{\cellcolor{#1}}
\definecolor{sourcecolor}{rgb}{0.78, 0.78, 0.78}
\definecolor{bestcolor}{rgb}{0.5, 0.95, 0.5}

\begin{document}\sloppy
% \renewcommand\thelinenumber{\color[rgb]{0.2,0.5,0.8}\normalfont\sffamily\scriptsize\arabic{linenumber}\color[rgb]{0,0,0}}
% \renewcommand\makeLineNumber {\hss\thelinenumber\ \hspace{6mm} \rlap{\hskip\textwidth\ \hspace{6.5mm}\thelinenumber}}
% \linenumbers
\pagestyle{headings}
\mainmatter
\def\ECCVSubNumber{2050}  % Insert your submission number here

\title{CoSMix: Compositional Semantic Mix for Domain Adaptation in 3D LiDAR Segmentation} % Replace with your title

% INITIAL SUBMISSION
%\begin{comment}
% \titlerunning{ECCV-22 submission ID \ECCVSubNumber} 
% \authorrunning{ECCV-22 submission ID \ECCVSubNumber} 
% \author{Anonymous ECCV submission}
% \institute{Paper ID \ECCVSubNumber}
%\end{comment}
%******************

% CAMERA READY SUBMISSION
% \begin{comment}
\titlerunning{CoSMix: Compositional Semantic Mix for DA in 3D LiDAR Segmentation}
% If the paper title is too long for the running head, you can set
% an abbreviated paper title here
%
% \author{Cristiano Saltori\inst{1}\orcidlink{0000-0001-9583-4160
% } \and
% Fabio Galasso\inst{2}\orcidlink{0000-0003-1875-7813} \and
% Giuseppe Fiameni\inst{3}\orcidlink{0000-0001-8687-6609} \and \\
% Nicu Sebe\inst{1}\orcidlink{0000-0002-6597-7248} \and
% Elisa Ricci\inst{1,4}\orcidlink{0000-0002-0228-1147} \and
% Fabio Poiesi\inst{4}\orcidlink{0000-0002-9769-1279}
% }

\author{Cristiano Saltori\inst{1} \and
Fabio Galasso\inst{2} \and
Giuseppe Fiameni\inst{3} \and \\
Nicu Sebe\inst{1} \and
Elisa Ricci\inst{1,4} \and
Fabio Poiesi\inst{4}
}
\authorrunning{C. Saltori et al.}
% First names are abbreviated in the running head.
% If there are more than two authors, 'et al.' is used.
%
\institute{University of Trento, Trento, Italy \and
Sapienza University of Rome, Rome, Italy \and
NVIDIA AI Technology Center \and
Fondazione Bruno Kessler, Trento, Italy\\
\email{cristiano.saltori@unitn.it}}
% \end{comment}
%******************
\maketitle

%%%%%%%%%%%%%%%%%%%%%%%%%%%%%%%%%%%%%%%%%%%%%%%
%%%%%%%%%%%%%%%%%%%%%%%%%%%%%%%%%%%%%%%%%%%%%%%
%%%%%%%%%%%%%%%%%%%%%%%%%%%%%%%%%%%%%%%%%%%%%%%
\begin{abstract}
3D LiDAR semantic segmentation is fundamental for autonomous driving.
Several Unsupervised Domain Adaptation (UDA) methods for point cloud data have been recently proposed to improve model generalization for different sensors and environments.
Researchers working on UDA problems in the image domain have shown that sample mixing can mitigate domain shift.
We propose a new approach of sample mixing for point cloud UDA, namely Compositional Semantic Mix (\ourmethod), the first UDA approach for point cloud segmentation based on sample mixing.
\ourmethod consists of a two-branch symmetric network that can process labelled synthetic data (source) and real-world unlabelled point clouds (target) concurrently.
Each branch operates on one domain by mixing selected pieces of data from the other one, and by using the semantic information derived from source labels and target pseudo-labels.
We evaluate \ourmethod on two large-scale datasets, showing that it outperforms state-of-the-art methods by a large margin. 
Our code is available at \url{https://github.com/saltoricristiano/cosmix-uda}.

% \footnote{\scriptsize{Our code is available at \url{https://github.com/saltoricristiano/cosmix-uda}.}}
\end{abstract}

% Our work merges these two research directions and introduces Compositional Semantic Mix (\ourmethod), the first UDA approach for point cloud segmentation based on sample mixing.
%%%%%%%%%%%%%%%%%%%%%%%%%%%%%%%%%%%%%%%%%%%%%%%%%%%%%%%%%%%
%%%%%%%%%%%%%%%%%%%%%%%%%%%%%%%%%%%%%%%%%%%%%%%%%%%%%%%%%%%
%%%%%%%%%%%%%%%%%%%%%%%%%%%%%%%%%%%%%%%%%%%%%%%%%%%%%%%%%%%
\section{Introduction}\label{sec:introduction}
Point cloud semantic segmentation is the problem of assigning a finite set of semantic labels to a set of 3D points~\cite{choy20194d, zhu2021cylindrical}.
When deep learning-based approaches are employed to perform this task, large-scale datasets with point-level annotations are required to learn accurate models~\cite{behley2019iccv, pan2020semanticposs, nuscenes2019}.
This implies a costly and cumbersome data collection procedure, as point clouds need to be captured in the real world and manually annotated.
An alternative is to use synthetic data, which can be conveniently generated with simulators~\cite{Dosovitskiy17}.
However, deep neural networks are known to underpeform when trained and tested on data from different domains, due to \textit{domain shift}~\cite{csurka2017domain}.
Although significant effort has been invested to design simulators that can reproduce the acquisition sensor with high fidelity, further research is still needed to neutralize the domain shift between real and synthetic domains \cite{synlidar}.

%++++++++++++++++++++++++++++++++++++++
\begin{figure}[t]
    \centering
    \includegraphics[width=\textwidth]{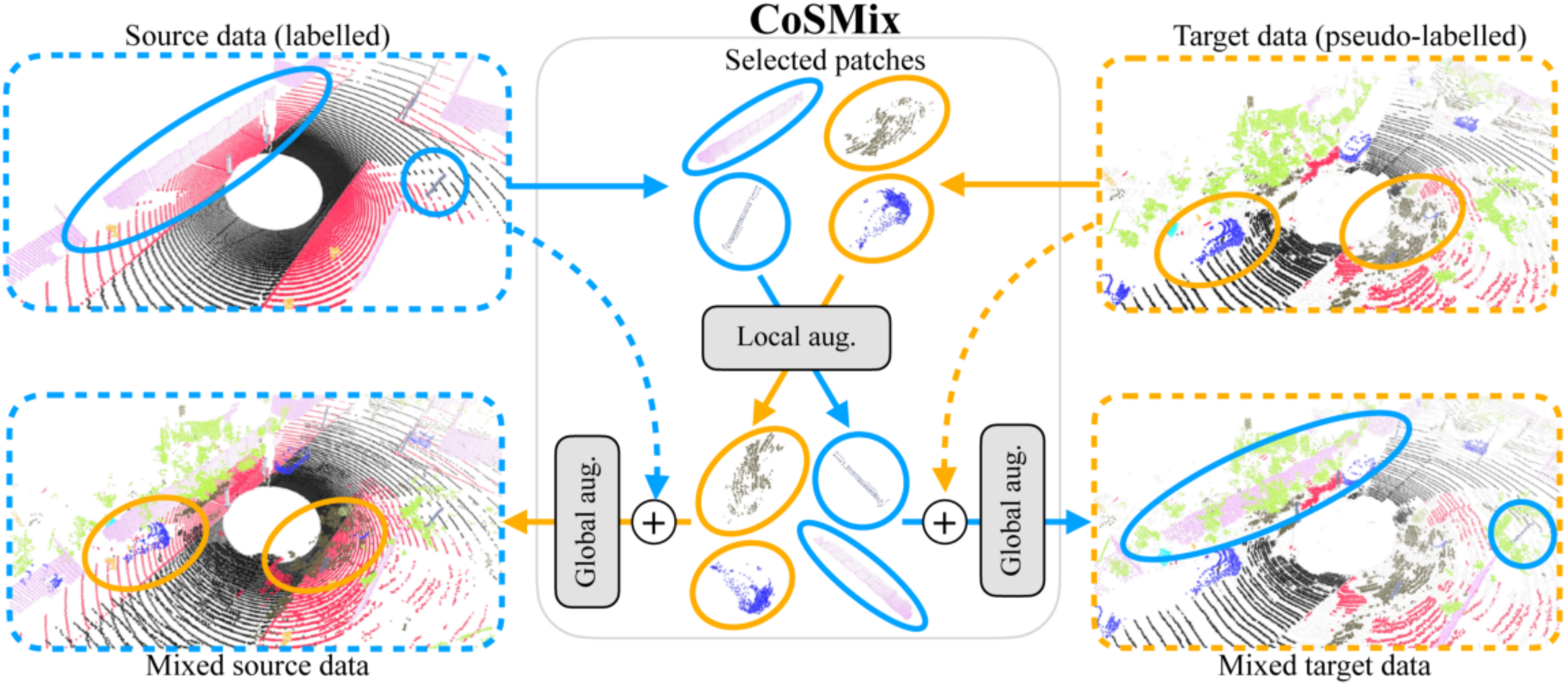}
    \vspace{-.5cm}
    \caption{\ourmethod applied to source and target data. 
    Given (labelled) source and (pseudo-labelled) target data, we select domain-specific patches with semantic information to be mixed across domains. 
    The resulting mixed data are a compositional semantic mix between the two domains, mixing source supervision in the target domain and target self-supervision (object and scene structure) in the source domain. 
    Augmentations are applied at both local and global levels.}
    \label{fig:teaser}
\end{figure}
%+++++++++++++++++++++++++++++++++++++

Unsupervised Domain Adaptation (UDA) for semantic segmentation has been widely studied for image data~\cite{wang2021domain, gao2021dsp, tranheden2021dacs, zou2018unsupervised, zou2019confidence}, however less attention has been paid to adaptation techniques for point clouds.
Approaches to address synthetic-to-real UDA for point clouds can operate in the input space~\cite{synlidar, zhao2020epointda} by using dropout rendering~\cite{zhao2020epointda}, or in the feature space through feature alignment~\cite{wu2019squeezesegv2}, or can use adversarial networks~\cite{synlidar}.
In the last few years, data augmentation approaches based on mixing of training samples and their labels, such as {Mixup}~\cite{mixup2018} or CutMix \cite{Yun2019}, have been proposed to promote generalization.
These techniques can be used for image classification~\cite{mixup2018, Yun2019}, image recognition \cite{xu2020adversarial, mancini2020towards}, and 2D semantic segmentation \cite{tranheden2021dacs, gao2021dsp}.
A few works proposed to exploit sample mixing for point cloud data~\cite{Nekrasov213DV,Zou2021,chen2020pointmixup, zhang2021pointcutmix}, but they are formulated for supervised applications.
We argue that the major challenge in extending 2D mix-based UDA approaches to point clouds lies in the application of these to geometric signals rather than photometric signals, e.g., the weighted alpha blending performed of labels in 2D~\cite{mixup2018, Yun2019} is still unclear how to extend it to 3D.

In this paper, we propose a novel UDA framework for 3D LiDAR segmentation, named \ourmethod, which can mitigate the domain shift by creating two new intermediate domains of composite point clouds obtained by applying a novel mixing strategy at input level (Fig.~\ref{fig:teaser}).
Our framework is based on a two-branch symmetric deep network structure that processes synthetic labelled point clouds (source) and real-world unlabelled point clouds (target).
Each branch is associated to a domain, \ie, on the source branch, a given source point cloud is mixed with parts of a target point cloud and vice versa for the target branch.
The mixing operation is implemented as a composition operation, which is similar to the concatenation operation proposed in \cite{Nekrasov213DV, Zou2021, chen2020pointmixup}, but unlike them, we account for the semantic information from source labels and target pseudo-labels to apply data augmentation both at local and global semantic level.
An additional key difference is the teacher-student learning scheme that we introduce to improve pseudo-label accuracy and, thus, point cloud composition.
We extensively evaluate our approach on recent and large scale segmentation benchmarks, \textit{i.e.}, considering SynLiDAR~\cite{synlidar} as source dataset, and SemanticPOSS~\cite{pan2020semanticposs} and SemanticKITTI~\cite{behley2019iccv} as target. 
Our results show that \ourmethod successfully alleviates the domain shift and outperforms state-of-the-art methods. We also perform an in-depth analysis of \ourmethod and an ablation study on each component, highlighting its strengths and discussing its main limitations.
To the best of our knowledge, this is the first work to have proposed a sample mixing scheme for adaptation in the context of 3D point cloud segmentation.

\vspace{.2cm}
\noindent Our main contributions can be summarised as follows:
\begin{itemize}
    % \item We introduce a novel scheme for mixing 3D point clouds leveraging semantic and geometric information.
    \item We introduce a novel scheme for mixing point clouds by leveraging semantic information and data augmentation.
    \item We show that the proposed mixing strategy can be used for reducing the domain shift and design \ourmethod, the first UDA method for 3D LiDAR semantic segmentation based on point cloud mixing.
    \item We conduct extensive experiments on two challenging synthetic-to-real 3D LiDAR semantic segmentation benchmarks demonstrating the effectiveness of \ourmethod, which outperforms state-of-the-art methods.
\end{itemize}

%By applying our mix strategy at input level, we can effectively align source and target data distributions to a common mixed space.
%Therefore, we aim to learn the target distribution by training on two different mixed point cloud sets.

% The problem of learning a 3D segmentation model from a large-scale \textit{source} dataset of synthetically generated and annotated point clouds, which is able to generalize to real-world point clouds (\textit{target}), has been addressed in the research community by introducing different techniques for Unsupervised Domain Adaptation (UDA)~\cite{yi2021complete,synlidar, zhao2020epointda}.
% \cris{The problem of learning a 3D segmentation model from a synthetic dataset (\textit{source}) that can generalize to real-world point clouds (\textit{target}) has been addressed by different Unsupervised Domain Adaptation (UDA) techniques.}

\section{Related works}\label{sec:related}

%%%%%%%%%%%%%%%%%%%%%%%%%%%%%%%%%%%%%%%%%%%%%%%%%%%%%%%%%%%%%%%%%%%%
\noindent\textbf{Point cloud semantic segmentation.}
Point cloud segmentation can be performed by using PointNet~\cite{qi2017pointnet} that is based on a series of multilayer perceptrons. 
PointNet++~\cite{qi2017pointnet++} improves PointNet by leveraging point aggregations performed at neighbourhood level and multi-scale sampling to encode both local features and global features.
RandLA-Net~\cite{hu2020randla} extends PoinNet++ \cite{qi2017pointnet++} by embedding local spatial encoding, random sampling and attentive pooling.
These methods are computationally inefficient when large-scale point clouds are processed.
Recent segmentation methods have improved the computational efficiency by projecting 3D points on 2D representations or by using 3D quantization approaches.
The former includes 2D projection based approaches that use 2D range maps and exploit standard 2D architectures~\cite{ronneberger2015u} to segment these maps prior to a re-projection in the 3D space. 
RangeNet++~\cite{milioto2019rangenet++}, SqueezeSeg networks~\cite{wu2018squeezeseg, wu2019squeezesegv2}, 3D-MiniNet~\cite{alonso2020MiniNet3D} and PolarNet~\cite{zhang2020polarnet} are examples of these approaches. 
Although these approaches are efficient, they tend to loose information when the input data are projected in 2D and re-projected in 3D.
The latter includes 3D quantization-based approaches that discretize the input point cloud into a 3D discrete representations and that employ 3D convolutions~\cite{zhou2018voxelnet} or 3D sparse convolutions~\cite{SubmanifoldSparseConvNet, choy20194d} to predict per-point classes. 
In this category, we find methods such as VoxelNet~\cite{zhou2018voxelnet}, SparseConv~\cite{SubmanifoldSparseConvNet, 3DSemanticSegmentationWithSubmanifoldSparseConvNet}, MinkowskiNet~\cite{choy20194d} and, Cylinder3D~\cite{zhu2021cylindrical}. 
In our work, we use the MinkowskiNet~\cite{choy20194d} which provides a trade off between accuracy and efficiency.

% \vspace{-.1cm}
%%%%%%%%%%%%%%%%%%%%%%%%%%%%%%%%%%%%%%%%%%%%%%%%%%%%%%%%%%%%%%%%%%%%
\noindent\textbf{Unsupervised domain adaptation for point cloud segmentation.}
Unsupervised Domain Adaptation (UDA) for point cloud segmentation can be used in the case of real-to-real~\cite{jaritz2019xmuda, yi2021complete, langer2020domain} and synthetic to real scenarios~\cite{wu2018squeezeseg, wu2019squeezesegv2, zhao2020epointda}.
Real-to-real adaptation can be used when a deep network is trained with data of real-world scenes captured with a LiDAR sensors and then tested on unseen scenes captured with a different LiDAR sensor~\cite{yi2021complete, langer2020domain}. 
Therein, domain adaptation can be formulated as a 3D surface completion task \cite{yi2021complete} or by transferring the sensor pattern of the target domain to the source domain through ray casting \cite{langer2020domain}.
Synthetic-to-real domain adaptation can be used when the source data are acquired with a simulated LiDAR sensor~\cite{Dosovitskiy17} and the target data are obtained with a real LiDAR sensor.
In this case, domain shift occurs due to differences in (i) sampling noise, (ii) structure of the environment and (iii) class distributions \cite{wu2019squeezesegv2, zhao2020epointda}.
Attention models can be used to aggregate contextual information~\cite{wu2018squeezeseg,wu2019squeezesegv2} and geodesic correlation alignment with progressive domain calibration can be adopted to improve domain adaptation~\cite{wu2019squeezesegv2}. 
In \cite{zhao2020epointda}, real dropout noise is simulated on synthetic data through a generative adversarial network.
Similarly, in \cite{synlidar} domain shift is disentangled into appearance difference and sparsity difference and a generative network is applied to mitigate each difference.
In our work, we do not use a learning-based approach to perturb the input data, but we formulate a novel compositional semantic point cloud mixing approach that enables the deep network to improve its performance on the unlabelled target domain self-supervisedly.

%%%%%%%%%%%%%%%%%%%%%%%%%%%%%%%%%%%%%%%%%%%%%%%%%%%%%%%%%%%%%%%%%%%%
\noindent\textbf{Sample Mixing for UDA.}
Deep neural networks often exhibit undesired behaviours such as memorization and overfitting. 
To alleviate this problem, mixing strategies~\cite{mixup2018, Yun2019} train a network on additional data derived from the convex combination of paired samples and labels, which are obtained either mixing the whole samples~\cite{mixup2018} or cutting and pasting their patches~\cite{Yun2019}.
Mixing strategies showed their effectiveness also in reducing domain shift in UDA for image classification~\cite{xu2020adversarial, wu2020dual} and semantic segmentation~\cite{tranheden2021dacs, gao2021dsp, Yang2020fourier}.
In DACS~\cite{tranheden2021dacs}, mixed samples are created by mixing pairs of images from different domains by using source ground-truth annotations pasted on pseudo-labelled target images. 
In DSP~\cite{gao2021dsp}, authors adopt a strategy that prioritize the selection of long-tail classes from the source domain images, and to paste their corresponding image patches on other source images and on target images.
The first point cloud mixing strategies~\cite{Nekrasov213DV, chen2020pointmixup, zhang2021pointcutmix} showed that point cloud pairs and their point-level annotations can be mixed for improving accuracy in semantic segmentation~\cite{Nekrasov213DV} and classification~\cite{chen2020pointmixup, zhang2021pointcutmix}.
Zou \textit{et al.}~\cite{Zou2021} propose to use Mix3D~\cite{Nekrasov213DV} as a pretext task for classification by predicting the rotation angle of mixed pairs.
Apply mixing strategy to address UDA in 3D semantic segmentation has not been previously investigated.
We fill this gap by introducing a novel compositional semantic mixing strategy that goes beyond the standard concatenation of two point clouds~\cite{Nekrasov213DV, zhang2021pointcutmix} or of randomly selected crops~\cite{zhang2021pointcutmix}. 
% Our approach selects patches of point clouds to be mixed based on the distribution of source classes and on the confidence of target pseudo-labels and generates new mixed point clouds for both source and target domains.

%%%%%%%%%%%%%%%%%%%%%%%%%%%%%%%%%%%%%%%%%%%%%%%%%%%%%%%%%%%%%%%%%%
%%%%%%%%%%%%%%%%%%%%%%%%%%%%%%%%%%%%%%%%%%%%%%%%%%%%%%%%%%%%%%%%%%
%%%%%%%%%%%%%%%%%%%%%%%%%%%%%%%%%%%%%%%%%%%%%%%%%%%%%%%%%%%%%%%%%%
\section{Our approach}\label{sec:method}

\begin{figure}[t]
    \centering
    \includegraphics[width=\textwidth]{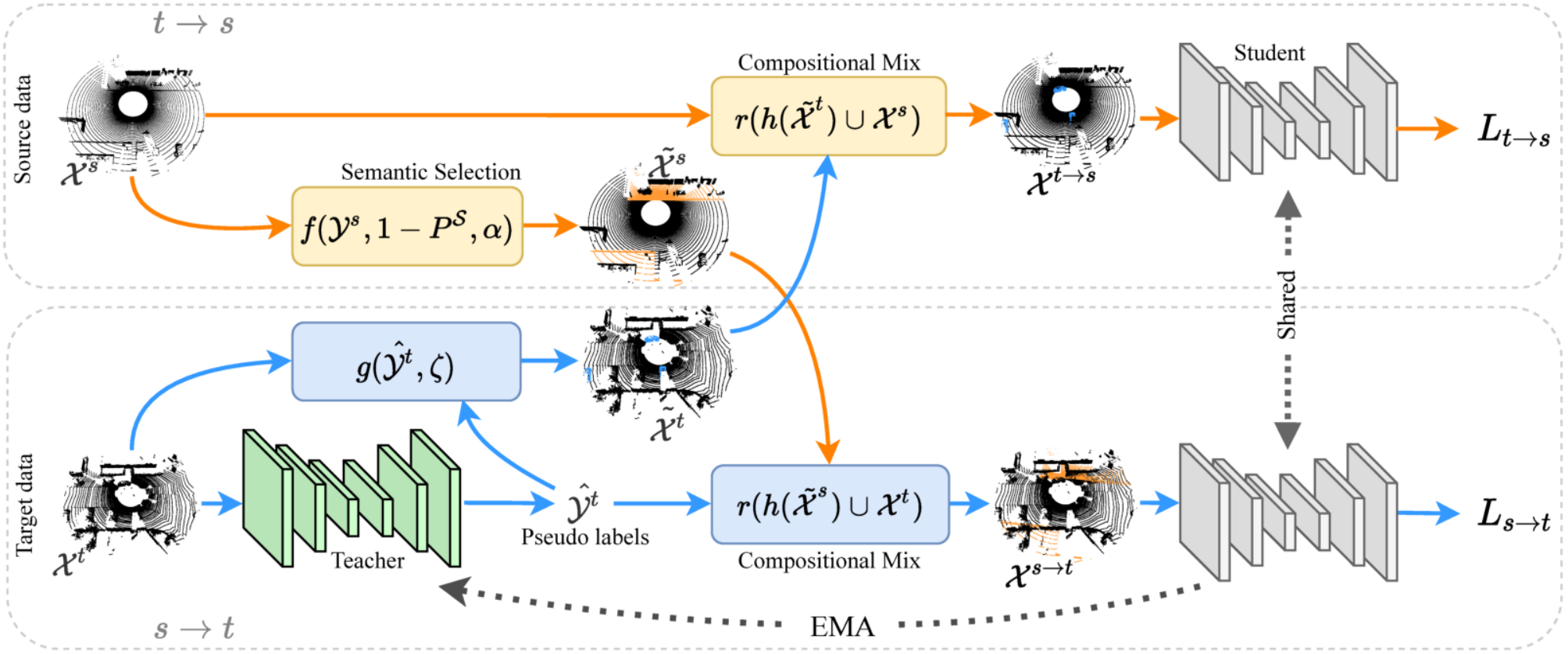}
    \vspace{-.4cm}
    \caption{Block diagram of \ourmethod. In the top branch, the input source point cloud $\mathcal{X}^s$ is mixed with the target point cloud $\mathcal{X}^t$ obtaining $\mathcal{X}^{t\rightarrow s}$. In the bottom branch, the input target point cloud $\mathcal{X}^t$ is mixed with the source point cloud $\mathcal{X}^s$ obtaining $\mathcal{X}^{s\rightarrow t}$. A teacher-student learning architecture is used to improve pseudo-label accuracy while adapting over target domain. Semantic Selection ($f$ and $g$) selects subsets of points (patches) to be mixed based on the source labels $\mathcal{Y}^s$ and target pseudo-labels $\hat{\mathcal{Y}^t}$ information. Compositional Mix applies local $h$ and global $r$ augmentations and mixes the selected patches among domains.}
    \label{fig:block_diagram}
\end{figure}

%%%%%%%%%%%%%%%%%%%%%%%%%%%%%%%%%%%%%%%%%%%%%%%%%%%%%%%%%%%%%%%%%%
%%%%%%%%%%%%%%%%%%%%%%%%%%%%%%%%%%%%%%%%%%%%%%%%%%%%%%%%%%%%%%%%%%
% \subsection{Overview and notation}

% \ourmethod is the first UDA approach for point cloud segmentation based on sample mixing.
\ourmethod implements a teacher-student learning scheme that exploits the supervision from the source domain and the self-supervision from the target domain to improve the semantic segmentation on the target domain.
Our method is trained on two different mixed point cloud sets.
The first is the composition of the source point cloud with pseudo-labelled pieces, or \textit{patches}, of the target point cloud. Target patches bring the target modality in the source domain pulling the source domain closer to the target domain.
The second is the composition of the target point cloud with randomly selected patches of the source point cloud. 
Source patches pull the target domain closer to the source domain, preventing overfitting from noisy pseudo-labels.
The teacher-student network enables the iterative improvement of pseudo labels, progressively reducing the domain gap.

Fig.~\ref{fig:block_diagram} shows the block diagram of \ourmethod.
Let $\mathcal{S} = \{(\mathcal{X}^s,\mathcal{Y}^s)\}$ be the source dataset that is composed of $N^s = |\mathcal{S}|$ labelled point clouds, where $\mathcal{X}^s$ is a point cloud and $\mathcal{Y}^s$ is its point-level labels, and $|.|$ is the cardinality of a set.
Labels take values from a set of semantic classes $\mathcal{C} = \{ c \}$, where $c$ is a semantic class.
Let $\mathcal{T} = \{\mathcal{X}^t\}$ be the target dataset composed of $N^t = |\mathcal{T}|$ unlabelled point clouds.
On the top branch, the source point cloud $\mathcal{X}^s$ is mixed with selected patches of the target point cloud $\mathcal{X}^t$.
The target patches are subsets of points that correspond to the most confident pseudo-labels $\hat{\mathcal{Y}^t}$ that the teacher network produces during training.
On the bottom branch, the target point cloud $\mathcal{X}^t$ is mixed with the selected patches of the source point cloud $\mathcal{X}^s$.
The source patches are subsets of points that are randomly selected based on their class frequency distribution in the training set.
Let $\mathcal{X}^{t \rightarrow s}$ be the mixed point cloud obtained from the top branch, and $\mathcal{X}^{s \rightarrow t}$ be the mixed point cloud obtained from the bottom branch.
We define the branch that mixes target point cloud patches to the source point cloud as $t \rightarrow s$ and the branch that does the vice versa as $s \rightarrow t$.
Lastly, let $\Phi_\mathcal{\theta}$ and $\Phi_\mathcal{\theta'}$ be the student and teacher deep networks with learnable parameters $\theta$ and $\theta'$, respectively.

We explain how the semantic selection operates on the source and target point clouds in Sec.~\ref{sec:semantic_selection}.
We detail the modules in charge of mixing the point clouds coming from the different domains in Sec.~\ref{sec:compositional_mix}.
Then, we describe how the teacher network is updated during training and the loss functions that we use to train the student networks in Sec.~\ref{sec:network_update}.

%%%%%%%%%%%%%%%%%%%%%%%%%%%%%%%%%%%%%%%%%%%%%%%%%%%%%%%%%%%%%%%%%%
%%%%%%%%%%%%%%%%%%%%%%%%%%%%%%%%%%%%%%%%%%%%%%%%%%%%%%%%%%%%%%%%%%
\subsection{Semantic selection}\label{sec:semantic_selection}

In order to train the student networks with balanced data, we select reliable and informative point cloud patches prior to mixing points and labels across domains.
A point cloud patch corresponds to a subset of points of the same semantic class.
To select patches from the source point cloud, we rely on the class frequency distribution by counting the number of points for each semantic class within $\mathcal{S}$.
Unlike DSP~\cite{gao2021dsp}, we do not select long-tail classes in advance, but we instead exploit the source distribution and the semantic classes available to dynamically sample classes at each iteration.

We define the class frequency distribution of $\mathcal{S}$ as $P_\mathcal{Y}^s$ and create a function $f$ that randomly selects a subset of classes based on the labels $\tilde{\mathcal{Y}}^s \subset \mathcal{Y}^s$ for supervision at each iteration. 
The likelihood that $f$ selects a class $c$ is inversely proportional to its class frequency in $\mathcal{S}$.
Specifically, 
%----------------------------------
\begin{equation}
    \tilde{\mathcal{Y}}^s = f(\mathcal{Y}^s, 1-P_\mathcal{Y}^s, \alpha),
\end{equation}
%----------------------------------
where $\alpha$ is an hyperparameter that regulates the ratio of selected classes for each point cloud.
For example, by setting $\alpha=0.5$, the algorithm will select a number of patches corresponding to the 50\% of the classes available by sampling them based on their class frequency distribution, i.e.,~long-tailed classes will have a higher likelihood to be selected.
We define the set of points that correspond to $\tilde{\mathcal{Y}}^s$ as $\tilde{\mathcal{X}}^s$, and a patch as the set of points $\tilde{\mathcal{X}}^s_c \subset \tilde{\mathcal{X}}^s$ that belong to class $c \in \mathcal{C}$.

To select patches from the target point clouds, we apply the same set of operations but using the pseudo-labels produced by the teacher network based on their prediction confidence.
Specifically, we define a function $g$ that selects reliable pseudo-labels based on their confidence value. 
The selected pseudo-labels are defined as 
%----------------------------------
\begin{equation}
    \tilde{\mathcal{Y}}^t = g(\Phi_{\theta'}(\mathcal{X}^t), \zeta),
\end{equation}
%----------------------------------
where $\Phi_{\theta'}$ is the teacher network, $\zeta$ is the confidence threshold used by the function $g$ and $\tilde{\mathcal{Y}}^t \subset \hat{\mathcal{Y}}^t$.
We define the set of points that correspond to $\tilde{\mathcal{Y}}^t$ as $\tilde{\mathcal{X}}^t$.

%%%%%%%%%%%%%%%%%%%%%%%%%%%%%%%%%%%%%%%%%%%%%%%%%%%%%%%%%%%%%%%%%%
%%%%%%%%%%%%%%%%%%%%%%%%%%%%%%%%%%%%%%%%%%%%%%%%%%%%%%%%%%%%%%%%%%
\subsection{Compositional mix}\label{sec:compositional_mix}

The goal of our compositional mixing module is to create mixed point clouds based on the selected semantic patches.
The compositional mix involves three consecutive operations:
\textit{local random augmentation}, patches are augmented randomly and independently from each other;
\textit{concatenation}, the augmented patches are concatenated to the point cloud of the other domain to create the mixed point cloud;
\textit{global random augmentation}, the mixed point cloud is randomly augmented.
This module is applied twice, once for the $t \rightarrow s$ branch (top of Fig.~\ref{fig:block_diagram}), where target patches are mixed within the source point cloud, and once for the $s \rightarrow t$ branch (bottom of Fig.~\ref{fig:block_diagram}), where source patches are mixed within the target point cloud.
Unlike Mix3D~\cite{Nekrasov213DV}, our mixing strategy embeds data augmentation at local level and global level.

In the $s \rightarrow t$ branch, we apply the local random augmentation $h$ to all the points $\tilde{\mathcal{X}}^s_c \subset \tilde{\mathcal{X}}^s$.
We repeat this operation for all $c \in \tilde{\mathcal{Y}}^s$.
Note that $h$ is a random augmentation that produces a different result each time it is applied to a set of points.
% and by exploiting the prior knowledge about the egocentric setup of LiDAR data to position the selected classes in geometrically meaningful locations.
Therefore, we define the result of this operation as
%----------------------------------
\begin{equation}
    h(\tilde{\mathcal{X}}^s) = \left \{ h(\tilde{\mathcal{X}}^s_c), \forall c \in \tilde{\mathcal{Y}}^s \right \}.
\end{equation}
%----------------------------------

Then, we concatenate $h(\tilde{\mathcal{X}}^s)$ with the source point cloud and apply the global random augmentation.
Their respective labels are concatenated accordingly, such as
%----------------------------------
\begin{equation}
    \mathcal{X}^{s \rightarrow t} = r(h(\tilde{\mathcal{X}}^s) \cup \mathcal{X}^t),  \,\,\,\,\,\, \mathcal{Y}^{s \rightarrow t} = \tilde{\mathcal{Y}}^s \cup \mathcal{Y}^t,
\end{equation}
%----------------------------------
where $r$ is the global augmentation function.
The same operations are also performed in the $t \rightarrow s$ branch by mixing target patches within the source point cloud. Instead of using source labels, we use the teacher network's pseudo-labels obtained from the target data and concatenate them with the labels of the source data.
This results in $\mathcal{X}^{t \rightarrow s}$ and $\mathcal{Y}^{t \rightarrow s}$.

%%%%%%%%%%%%%%%%%%%%%%%%%%%%%%%%%%%%%%%%%%%%%%%%%%%%%%%%%%%%%%%%%%
%%%%%%%%%%%%%%%%%%%%%%%%%%%%%%%%%%%%%%%%%%%%%%%%%%%%%%%%%%%%%%%%%%
\subsection{Network update}\label{sec:network_update}

We leverage the teacher-student learning scheme to facilitate the transfer of knowledge acquired during the course of the training with mixed domains.
We use the teacher network $\Phi_{\theta'}$ to produce target pseudo-labels $ \hat{\mathcal{Y}}^t$ for the student network $\Phi_{\theta}$, and train $\Phi_{\theta}$ to segment target point clouds by using the mixed point clouds $\mathcal{X}^{s \rightarrow t}$ and $\mathcal{X}^{t \rightarrow s}$ based on their mixed labels and pseudo-labels (Sec.~\ref{sec:compositional_mix}).

At each batch iteration, we update the student parameters $\Phi_\theta$ to minimize a total objective loss $\mathcal{L}_{tot}$ defined as
%------------------------------------------
\begin{equation}
    \mathcal{L}_{tot} = \mathcal{L}_{s \rightarrow t} + \mathcal{L}_{t \rightarrow s},
\end{equation}
%------------------------------------------
where $\mathcal{L}_{s \rightarrow t}$ and $\mathcal{L}_{t \rightarrow s}$ are the $s \rightarrow t$ and $t \rightarrow s$ branch losses, respectively.
Given $\mathcal{X}^{s \rightarrow t}$ and  $\mathcal{Y}^{s \rightarrow t}$, we define the segmentation loss for the $s \rightarrow t$ branch as
%------------------------------------------
\begin{equation}
    \mathcal{L}_{s \rightarrow t} = \mathcal{L}_{seg}(\Phi_{\theta}(\mathcal{X}^{s \rightarrow t}), \mathcal{Y}^{s \rightarrow t}),
\end{equation}
%------------------------------------------
the objective of which is to minimize the segmentation error over $\mathcal{X}^{s \rightarrow t}$, thus learning to segment source patches in the target domain.
Similarly, given $\mathcal{X}^{t \rightarrow s}$ and  $\mathcal{Y}^{t \rightarrow s}$, we define the segmentation loss for the $t \rightarrow s$ branch as
%------------------------------------------
\begin{equation}
    \mathcal{L}_{t \rightarrow s} = \mathcal{L}_{seg}(\Phi_{\theta}(\mathcal{X}^{t \rightarrow s}), \mathcal{Y}^{t \rightarrow s}),
\end{equation}
%------------------------------------------
whose objective is to minimize the segmentation error over $\mathcal{X}^{t \rightarrow s}$ where target patches are composed with source data. 
We implement $\mathcal{L}_{seg}$ as the Dice segmentation loss~\cite{Jadon2020}, which we found effective for the segmentation of large-scale point clouds as it can cope with long-tail classes well.

Lastly, we update the teacher parameters $\theta'$ every $\gamma$ iterations following the exponential moving average (EMA)\cite{gao2021dsp} approach
%------------------------------------------
\begin{equation}
    \theta'_i = \beta\theta'_{i-1} + (1-\beta)\theta,
\end{equation}
%------------------------------------------
where $i$ indicates the training iteration and $\beta$ is a smoothing coefficient hyperparamenter.

\section{Experiments}\label{sec:experimental}

% \fp{provide a high-level description of what you aim to show and demonstrate with the results you are presenting, by motivating the choice of the datasets you have chosen.}
We evaluate our method in the synthetic-to-real UDA scenario for LiDAR segmentation. 
We use the SynLiDAR dataset~\cite{synlidar} as (synthetic) source domain, and the SemanticKITTI~\cite{behley2019iccv, geiger2012we, geiger2013vision} and SemanticPOSS~\cite{pan2020semanticposs} datasets as (real) target domains (more details in Sec.~\ref{sec:dataset}).
We describe \ourmethod implementation in Sec.~\ref{sec:implementation}.
We compare \ourmethod with five state-of-the-art UDA methods: two general purpose adaptation methods (ADDA~\cite{tzeng2017adversarial}, Ent-Min~\cite{vu2019advent}), one image segmentation method (ST~\cite{zou2019confidence}) and, two point cloud segmentation methods (PCT~\cite{synlidar}, ST-PCT~\cite{synlidar}) (Sec.~\ref{sec:results}).
Like~\cite{synlidar}, we compare \ourmethod against methods working on 3D point clouds for synthetic to real, such as PCT~\cite{synlidar} and ST-PCT~\cite{synlidar}. 
These are the only two state-of-the-art methods for synthetic-to-real UDA that use 360$^\circ$ LiDAR point clouds. 
Results of baselines are taken from~\cite{synlidar}.

%%%%%%%%%%%%%%%%%%%%%%%%%%%%%%%%%%%%%%%%%%%%%%%%%%%%%%%%%%%%%%%
%%%%%%%%%%%%%%%%%%%%%%%%%%%%%%%%%%%%%%%%%%%%%%%%%%%%%%%%%%%%%%%
\subsection{Datasets and metrics}\label{sec:dataset}

\noindent \textbf{SynLiDAR}~\cite{synlidar} is a large-scale synthetic dataset that is captured with the Unreal Engine~\cite{Dosovitskiy17}.
It is composed of 198,396 LiDAR scans with point-level segmentation annotations over 32 semantic classes. 
We follow the authors' instructions~\cite{synlidar}, and use 19,840 point clouds for training and 1,976 point clouds for validation.

\noindent \textbf{SemanticPOSS}~\cite{pan2020semanticposs} consists of 2,988 real-world scans with point-level annotations over 14 semantic classes.
Based on the official benchmark guidelines~\cite{pan2020semanticposs}, we use the sequence $03$ for validation and the remaining sequences for training.

\noindent \textbf{SemanticKITTI}~\cite{behley2019iccv} is a large-scale segmentation dataset consisting of LiDAR acquisitions of the popular KITTI dataset~\cite{geiger2012we, geiger2013vision}.
It is composed of 43,552 scans captured in Karlsruhe (Germany) and point-level annotations over 19 semantic classes. 
Based on the official protocol~\cite{behley2019iccv}, we use sequence $08$ for validation and the remaining sequences for training.

\noindent \textbf{Class mapping.} Like~\cite{synlidar}, we make source and target labels compatible across our datasets, i.e., SynLiDAR $\rightarrow$ SemanticPOSS and SynLiDAR $\rightarrow$ SemanticKITTI. 
We map SynLiDAR labels into 14 segmentation classes for SynLiDAR $\rightarrow$ SemanticPOSS and 19 segmentation classes for SynLiDAR $\rightarrow$ SemanticKITTI~\cite{synlidar}.

\noindent \textbf{Metrics.} We follow the typical evaluation protocol for UDA in 3D semantic segmentation~\cite{synlidar} and evaluate the segmentation performance before and after adaptation. 
We compute the Intersection over the Union (IoU)~\cite{rahman2016optimizing} for each segmentation class and report the per-class IoU. 
Then, we average the IoU over all the segmented classes and report the mean Intersection over the Union (mIoU).

%%%%%%%%%%%%%%%%%%%%%%%%%%%%%%%%%%%%%%%%%%%%%%%%%%%%%%%%%%%%%%%
%%%%%%%%%%%%%%%%%%%%%%%%%%%%%%%%%%%%%%%%%%%%%%%%%%%%%%%%%%%%%%%
\subsection{Implementation details}\label{sec:implementation}
We implemented \ourmethod in PyTorch and run our experiments on 4$\times$NVIDIA A100 (40GB SXM4).
We use MinkowskiNet as our point cloud segmentation network~\cite{choy20194d}. 
For a fair comparison, we use MinkUNet32 as in~\cite{synlidar}. 
We use warm-up, i.e., our network is pre-trained on the source domain for 10 epochs with Dice loss~\cite{Jadon2020} starting from randomly initialized weights.
During the adaptation step, we initialize student and teacher networks with the parameters obtained after warm-up.
The warm-up and adaptation stage share the same hyperparameters.
In both the warm-up and adaptation steps, we use Stochastic Gradient Descent (SGD) with a learning rate of $0.001$.
We set $\alpha$ by analyzing the long-tailed classes in the source domain during adaptation. 
We experimentally found $\alpha = 50\%$ to be a good value in each task.
In the target semantic selection function $g$, we set $\zeta$ such that about $80\%$ of pseudo-labelled points per scene can be selected.
On SynLiDAR$\rightarrow$SemanticPOSS, we use a batch size of $12$ and perform adaptation for $10$ epochs. 
We set source semantic selection $f$ with $\alpha=0.5$ while target semantic selection $g$ with a confidence threshold $\zeta=0.85$ (Sec.~\ref{sec:semantic_selection}).
On SynLiDAR$\rightarrow$SemanticKITTI, we use a batch size of $16$, adapting for $3$ epochs.
During source semantic selection $f$ we set $\alpha=0.5$ while in target semantic selection $g$ we use a confidence threshold of $\zeta=0.90$.
We use the same compositional mix (Sec.~\ref{sec:compositional_mix}) parameters for both the adaptation directions. 
% We set local augmentation $h$ to apply rigid rotation around the $z$-axis and scaling along all the axis.
We implement the local augmentation $h$ as rigid rotation around the $z$-axis, scaling along all the axes and random point downsampling.
We bound rotations between $[-\pi/2, \pi/2]$ and scaling between $[0.95, 1.05]$, and perform random downsampling for 50\% of the patch points.
For global augmentation $r$, we use a rigid rotation, translation and scaling along all the three axes. 
We set $r$ parameters to the same used in~\cite{choy20194d}.
During the network update step (Sec.~\ref{sec:network_update}), we obtain the teacher parameters $\theta'_i$ with $\beta=0.99$ every $\gamma=1$ steps on SynLiDAR$\rightarrow$SemanticPOSS and every $\gamma=500$ steps on SynLiDAR$\rightarrow$SemanticKITTI.

%%%%%%%%%%%%%%%%%%%%%%%%%%%%%%%%%%%%%%%%%%%%%%%%%%%%%%%%%%%%%%%
%%%%%%%%%%%%%%%%%%%%%%%%%%%%%%%%%%%%%%%%%%%%%%%%%%%%%%%%%%%%%%%
\subsection{Quantitative comparisons}\label{sec:results}

\begin{table}[t]
    \centering
    \caption{Adaptation results on SynLiDAR $\rightarrow$ SemanticPOSS. Source corresponds to the model trained on the source synthetic dataset (lower bound in gray). Results are reported in terms of mean Intersection over the Union (mIoU).}
    \vspace{.1cm}
    \label{tab:adaptation_poss}
    \tabcolsep 4pt
    \resizebox{\textwidth}{!}{%
    \begin{tabular}{l|ccccccccccccc|c}
        \toprule
        %\begin{turn}{70}Source Free\end{turn}& \begin{turn}{70}Supervised\end{turn}
        \textbf{Model} & \textbf{pers.} & \textbf{rider} & \textbf{car} & \textbf{trunk} & \textbf{plants} & \textbf{traf.} & \textbf{pole} & \textbf{garb.} & \textbf{buil.} & \textbf{cone.} & \textbf{fence} & \textbf{bike} & \textbf{grou.} & \textbf{mIoU} \\
        \midrule
        
        \CC{sourcecolor}Source & \CC{sourcecolor}3.7 & \CC{sourcecolor}25.1 & \CC{sourcecolor}12.0 & \CC{sourcecolor}10.8 & \CC{sourcecolor}53.4 & \CC{sourcecolor}0.0 & \CC{sourcecolor}19.4 & \CC{sourcecolor}12.9 & \CC{sourcecolor}49.1 & \CC{sourcecolor}3.1 & \CC{sourcecolor}20.3 & \CC{sourcecolor}0.0 & \CC{sourcecolor}59.6 & \CC{sourcecolor}20.7 \\
        \midrule
        ADDA~\cite{tzeng2017adversarial} & 27.5 & 35.1 & 18.8 & 12.4 & 53.4 & 2.8 & 27.0 & 12.2 & 64.7 & 1.3 & 6.3 & 6.8 & 55.3 & 24.9 \\
        Ent-Min~\cite{vu2019advent} & 24.2 & 32.2 & 21.4 & 18.9 & 61.0 & 2.5 & 36.3 & 8.3 & 56.7 & 3.1 & 5.3 & 4.8 & 57.1 & 25.5 \\
        ST~\cite{zou2019confidence} & 23.5 & 31.8 & 22.0 & 18.9 & 63.2 & 1.9 & \textbf{41.6} & 13.5 & 58.2 & 1.0 & 9.1 & 6.8 & 60.3 & 27.1\\
        PCT~\cite{synlidar} & 13.0 & 35.4 & 13.7 & 10.2 & 53.1 & 1.4 & 23.8 & 12.7 & 52.9 & 0.8 & 13.7 & 1.1 & 66.2 & 22.9\\
        ST-PCT~\cite{synlidar} & 28.9 & 34.8 & 27.8 & 18.6 & 63.7 & 4.9 & 41.0 & 16.6 & 64.1 & 1.6 & 12.1 & 6.6 & 63.9 & 29.6\\
        \midrule
        \ourmethod (Ours) & \textbf{55.8} & \textbf{51.4} & \textbf{36.2} & \textbf{23.5} & \textbf{71.3} & \textbf{22.5} & 34.2 & \textbf{28.9} & \textbf{66.2} & \textbf{20.4} & \textbf{24.9} & \textbf{10.6} & \textbf{78.7} & \textbf{40.4}\\
        \bottomrule
    \end{tabular}
    }
\end{table}

\begin{table}[t]
    \centering
    \caption{Adaptation results on SynLiDAR $\rightarrow$ SemanticKITTI. Source corresponds to the model trained on the source synthetic dataset (lower bound in gray). Results are reported in terms of mean Intersection over the Union (mIoU).}
    \vspace{.1cm}
    \label{tab:adaptation_kitti}
    \resizebox{\textwidth}{!}{%
    \begin{tabular}{l|ccccccccccccccccccc|c}
        \toprule
        \textbf{Model} & \rotatebox{90}{\textbf{car}} & \rotatebox{90}{\textbf{bi.cle}} & \rotatebox{90}{\textbf{mt.cle}} & \rotatebox{90}{\textbf{truck}} & \rotatebox{90}{\textbf{oth-v.}} & \rotatebox{90}{\textbf{pers.}} & \rotatebox{90}{\textbf{b.clst}} & \rotatebox{90}{\textbf{m.clst}} & \rotatebox{90}{\textbf{road}} & \rotatebox{90}{\textbf{park.}} & \rotatebox{90}{\textbf{sidew.}} & \rotatebox{90}{\textbf{oth-g.}} & \rotatebox{90}{\textbf{build.}} & \rotatebox{90}{\textbf{fence}} & \rotatebox{90}{\textbf{veget.}} & \rotatebox{90}{\textbf{trunk}} & \rotatebox{90}{\textbf{terra.}} & \rotatebox{90}{\textbf{pole}} & \rotatebox{90}{\textbf{traff.}} & \textbf{mIoU} \\
        \midrule
        
        \CC{sourcecolor} Source & \CC{sourcecolor} 42.0 & \CC{sourcecolor}5.0 & \CC{sourcecolor}4.8 & \CC{sourcecolor}0.4 & \CC{sourcecolor}2.5 & \CC{sourcecolor}12.4 & \CC{sourcecolor}43.3 & \CC{sourcecolor}1.8 & \CC{sourcecolor}48.7 & \CC{sourcecolor}4.5 & \CC{sourcecolor}31.0 & \CC{sourcecolor}0.0 & \CC{sourcecolor}18.6 & \CC{sourcecolor}11.5 & \CC{sourcecolor}60.2 & \CC{sourcecolor}30.0 & \CC{sourcecolor}48.3 & \CC{sourcecolor}19.3 & \CC{sourcecolor}3.0 & \CC{sourcecolor}20.4 \\
        \midrule

        ADDA~\cite{tzeng2017adversarial} & 52.5 & 4.5 & 11.9 & 0.3 & 3.9 & 9.4 & 27.9 & 0.5 & 52.8 & 4.9 & 27.4 & 0.0 & 61.0 & 17.0 & 57.4 & 34.5 & 42.9 & 23.2 & 4.5 & 23.0\\
        Ent-Min~\cite{vu2019advent} & 58.3 & 5.1 & 14.3 & 0.3 & 1.8 & 14.3 & \textbf{44.5} & 0.5 & 50.4 & 4.3 & 34.8 & 0.0 & 48.3 & 19.7 & 67.5 & 34.8 & \textbf{52.0} & 33.0 & 6.1 & 25.8 \\
        ST~\cite{zou2019confidence} & 62.0 & 5.0 & 12.4 & 1.3 & 9.2 & 16.7 & 44.2 & 0.4 & 53.0 & 2.5 & 28.4 & 0.0 & 57.1 & 18.7 & 69.8 & \textbf{35.0} & 48.7 & 32.5 & 6.9 & 26.5 \\
        PCT~\cite{synlidar} & 53.4 & 5.4 & 7.4 & 0.8 & 10.9 & 12.0 & 43.2 & 0.3 & 50.8 & 3.7 & 29.4 & 0.0 & 48.0 & 10.4 & 68.2 & 33.1 & 40.0 & 29.5 & 6.9 & 23.9 \\
        ST-PCT~\cite{synlidar} & 70.8 & \textbf{7.3} & 13.1 & 1.9 & 8.4 & 12.6 & 44.0 & 0.6 & 56.4 & 4.5 & 31.8 & 0.0 & \textbf{66.7} & \textbf{23.7} & \textbf{73.3} & 34.6 & 48.4 & \textbf{39.4} & 11.7 & 28.9 \\
        \midrule
        \ourmethod (Ours) & \textbf{75.1} & 6.8 & \textbf{29.4} & \textbf{27.1} & \textbf{11.1} & \textbf{22.1} & 25.0 & \textbf{24.7} & \textbf{79.3} & \textbf{14.9} & \textbf{46.7} & \textbf{0.1} & 53.4 & 13.0 & 67.7 & 31.4 & 32.1 & 37.9 & \textbf{13.4} & \textbf{32.2}\\
        \bottomrule

    \end{tabular}
    }
\end{table}

Tab.~\ref{tab:adaptation_poss} and Tab.~\ref{tab:adaptation_kitti} reports the adaptation results on SynLiDAR$\rightarrow$SemanticPOSS, and on SynLiDAR$\rightarrow$SemanticKITTI, respectively.
The Source model is the lower bound of each scenario with $20.7$ mIoU on SynLiDAR$\rightarrow$SemanticPOSS and $22.2$ mIoU on SynLiDAR$\rightarrow$SemanticKITTI. 
We highlight in gray the associated results in both tables.
In SynLiDAR$\rightarrow$SemanticPOSS (Tab.~\ref{tab:adaptation_poss}), \ourmethod outperforms the baselines on all the classes, with the exception of \textit{pole} where ST achieves better results.
On average, we achieve $40.4$ mIoU surpassing ST-PCT by $+10.8$ mIoU and improving over the Source of $+19.7$ mIoU. Interestingly, \ourmethod improves also on difficult classes as in the case of \textit{person}, \textit{traffic-sign}, \textit{cone} and, \textit{bike}, whose performance were low before adaptation.
SemanticKITTI is a more challenging domain as the validation sequence includes a wide range of different scenarios with a large number of semantic classes.
In SynLiDAR$\rightarrow$SemanticKITTI (Tab.~\ref{tab:adaptation_kitti}), \ourmethod improves on all the classes when compared to Source, with the exception of \textit{bicyclist} and \textit{terrain}. 
We relate this behaviour to the additional noise introduced by pseudo labels on these classes and in related classes such as $sidewalk$. 
Compared to the other baselines, \ourmethod improves on $11$ out of $19$ classes, with a large margin in the classes \textit{car}, \textit{motorcycle}, \textit{truck}, \textit{person}, \textit{road}, \textit{parking} and \textit{sidewalk}. 
On average, also in this more challenging scenario, we achieve the new state-of-the-art performance of $32.2$ mIoU, outperforming ST-PCT by $+3.3$ mIoU and improving over Source of about $+11.8$ mIoU.

%%%%%%%%%%%%%%%%%%%%%%%%%%%%%%%%%%%%%%%%%%%%%%%%%%%%%%%%%%%%%%%
%%%%%%%%%%%%%%%%%%%%%%%%%%%%%%%%%%%%%%%%%%%%%%%%%%%%%%%%%%%%%%%
\subsection{Qualitative results}\label{sec:qualitative}

\begin{figure}[t]
\centering
    \setlength\tabcolsep{1.pt}
    \begin{tabular}{ccc}
    \raggedright
        \begin{overpic}[width=0.33\textwidth]{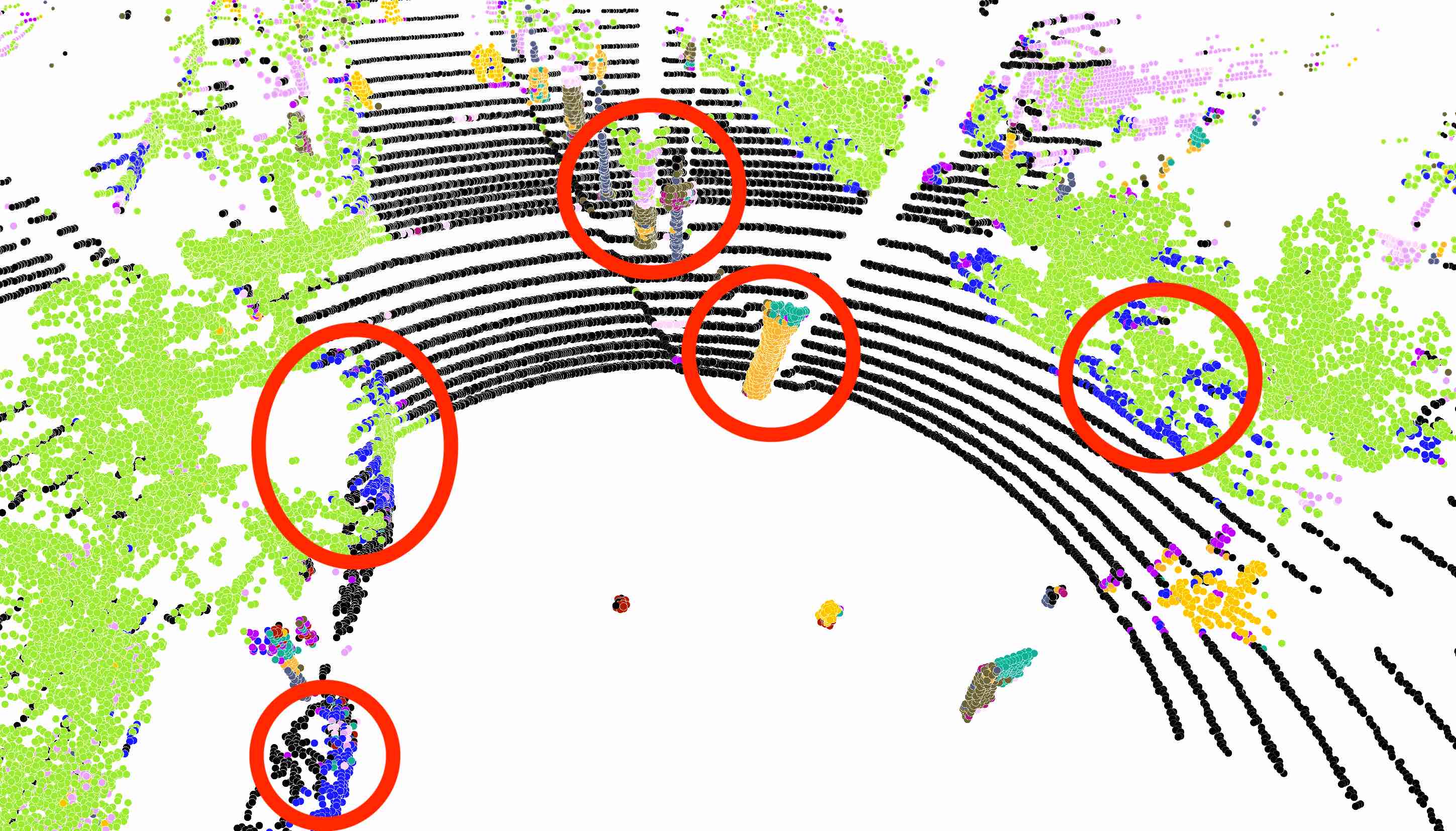}
        \put(40,67){\color{black}\footnotesize \textbf{source}}
        \put(160,67){\color{black}\footnotesize \textbf{ours}}
        \put(285,68){\color{black}\footnotesize \textbf{gt}}
        \end{overpic} &  
        \begin{overpic}[width=0.33\textwidth]{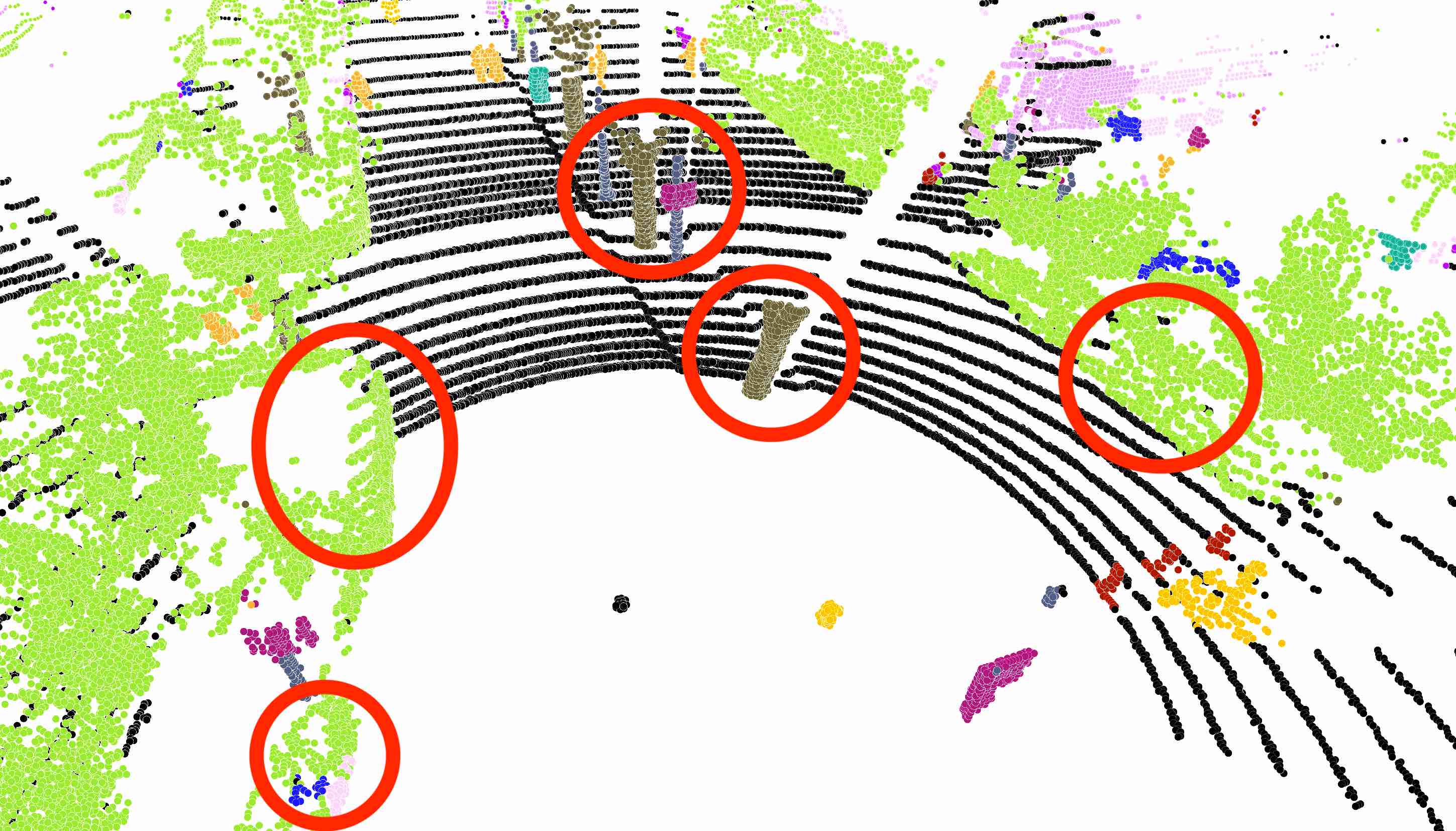}
        \end{overpic} &
        \begin{overpic}[width=0.33\textwidth]{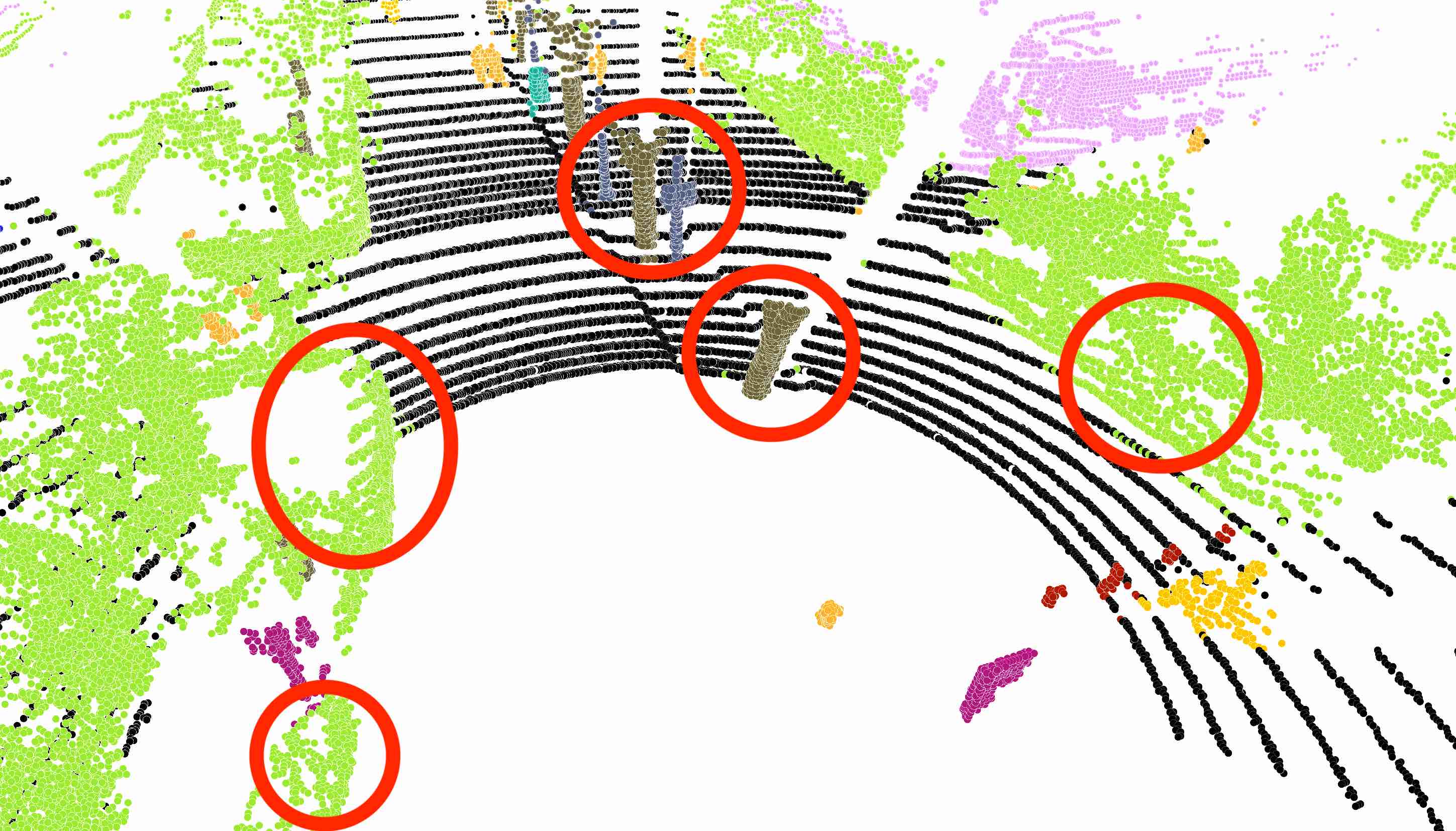}
        \end{overpic}\\
        \begin{overpic}[width=0.33\textwidth]{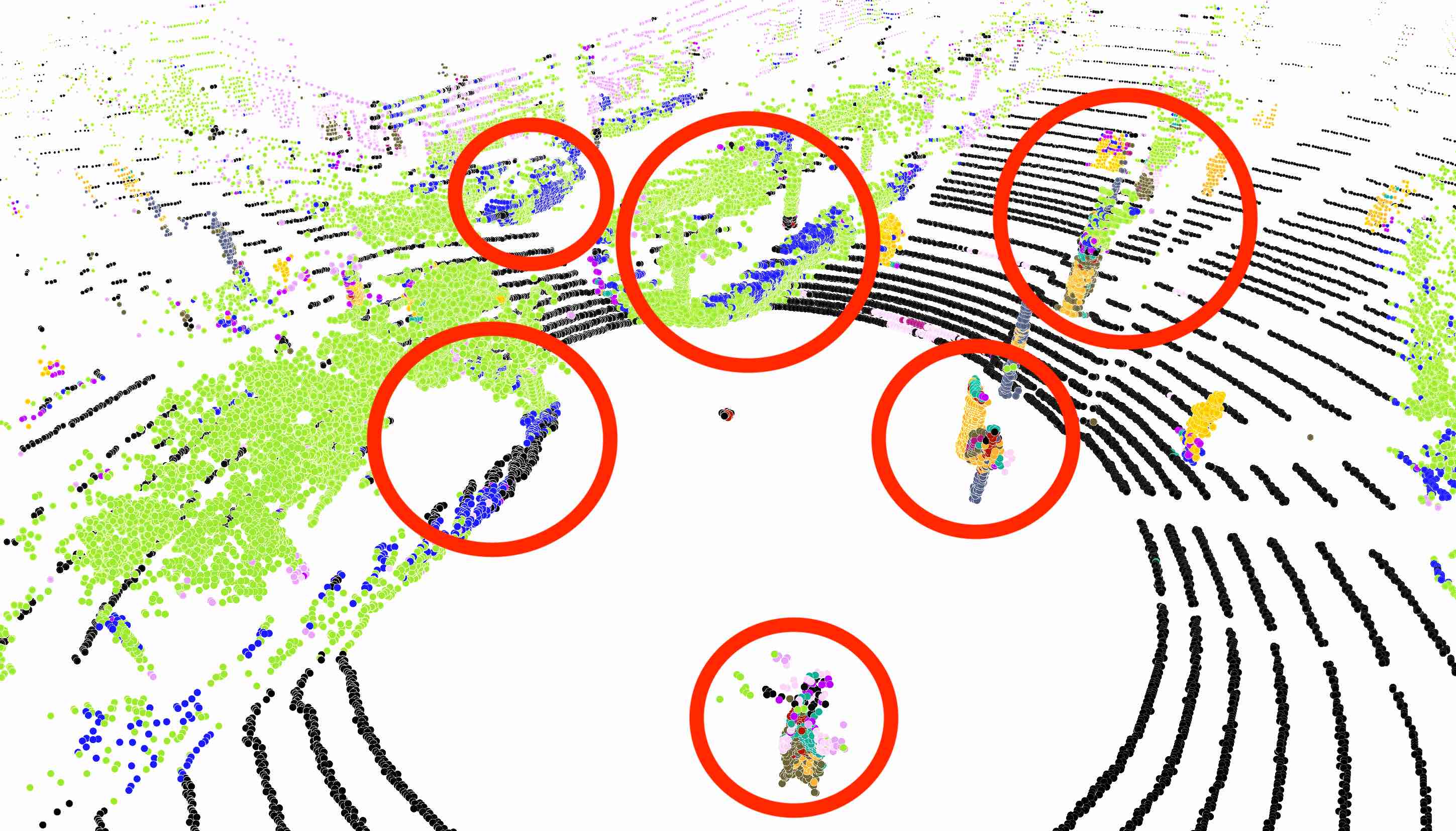}
        % \put(40,55){\color{black}\footnotesize \textbf{source}}
        % \put(-7,20){\color{black}\footnotesize \rotatebox{90}{\textbf{large}}}
        \end{overpic} &  
        \begin{overpic}[width=0.33\textwidth]{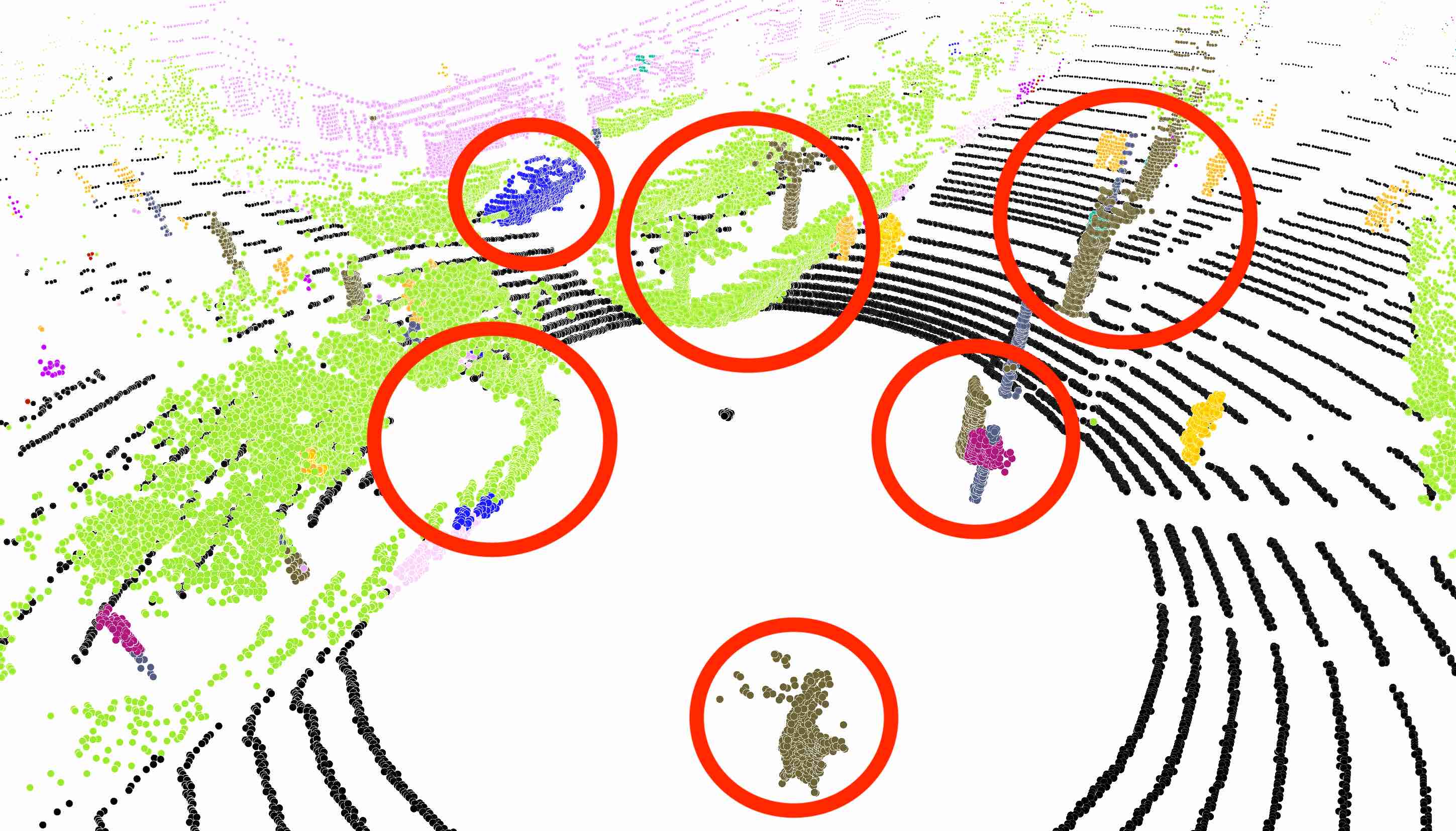}
        \end{overpic} &
        \begin{overpic}[width=0.33\textwidth]{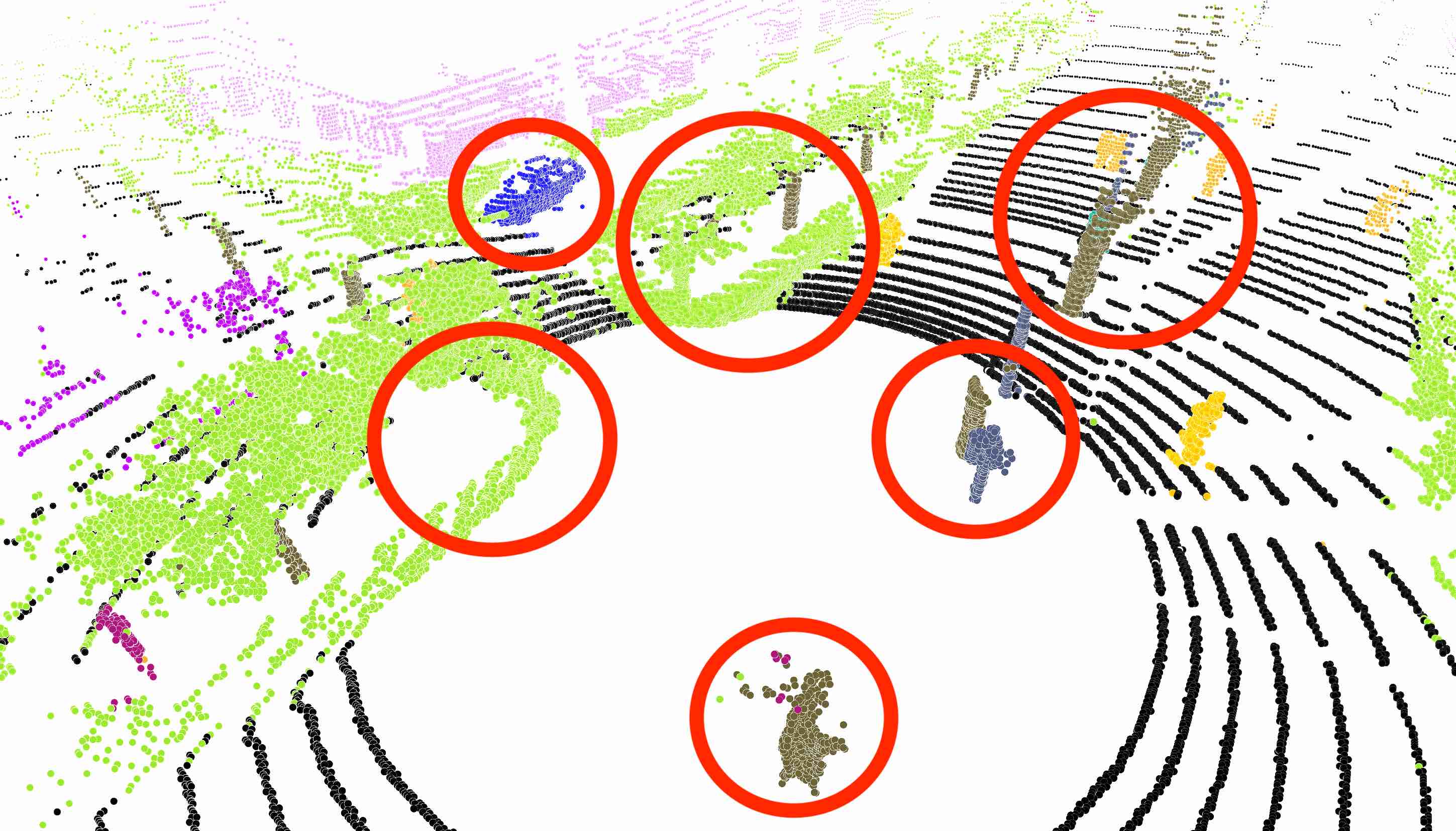}
        \end{overpic}\\
        \begin{overpic}[width=0.33\textwidth]{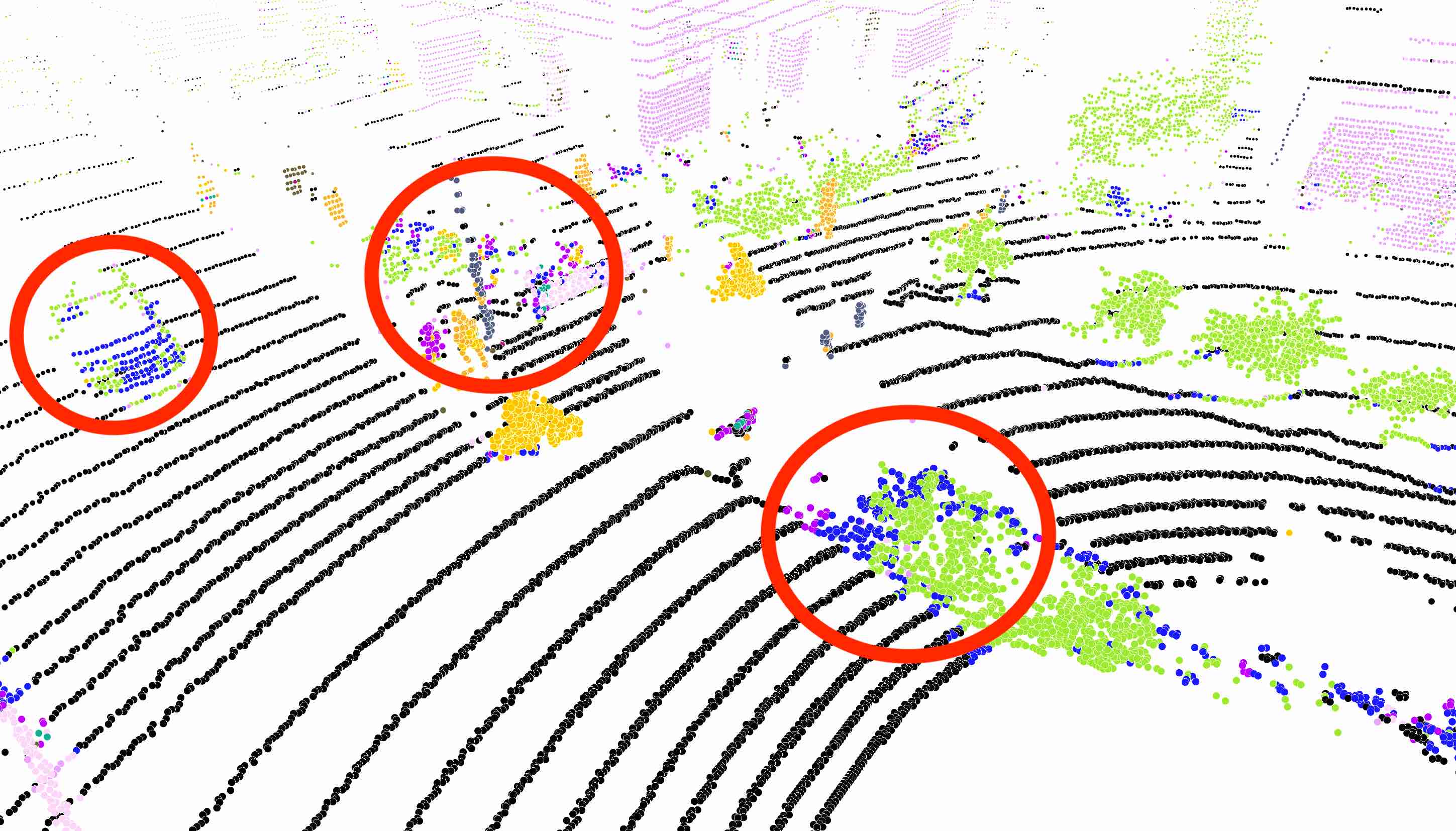}
        % \put(40,55){\color{black}\footnotesize \textbf{source}}
        % \put(-7,20){\color{black}\footnotesize \rotatebox{90}{\textbf{large}}}
        \end{overpic} &  
        \begin{overpic}[width=0.33\textwidth]{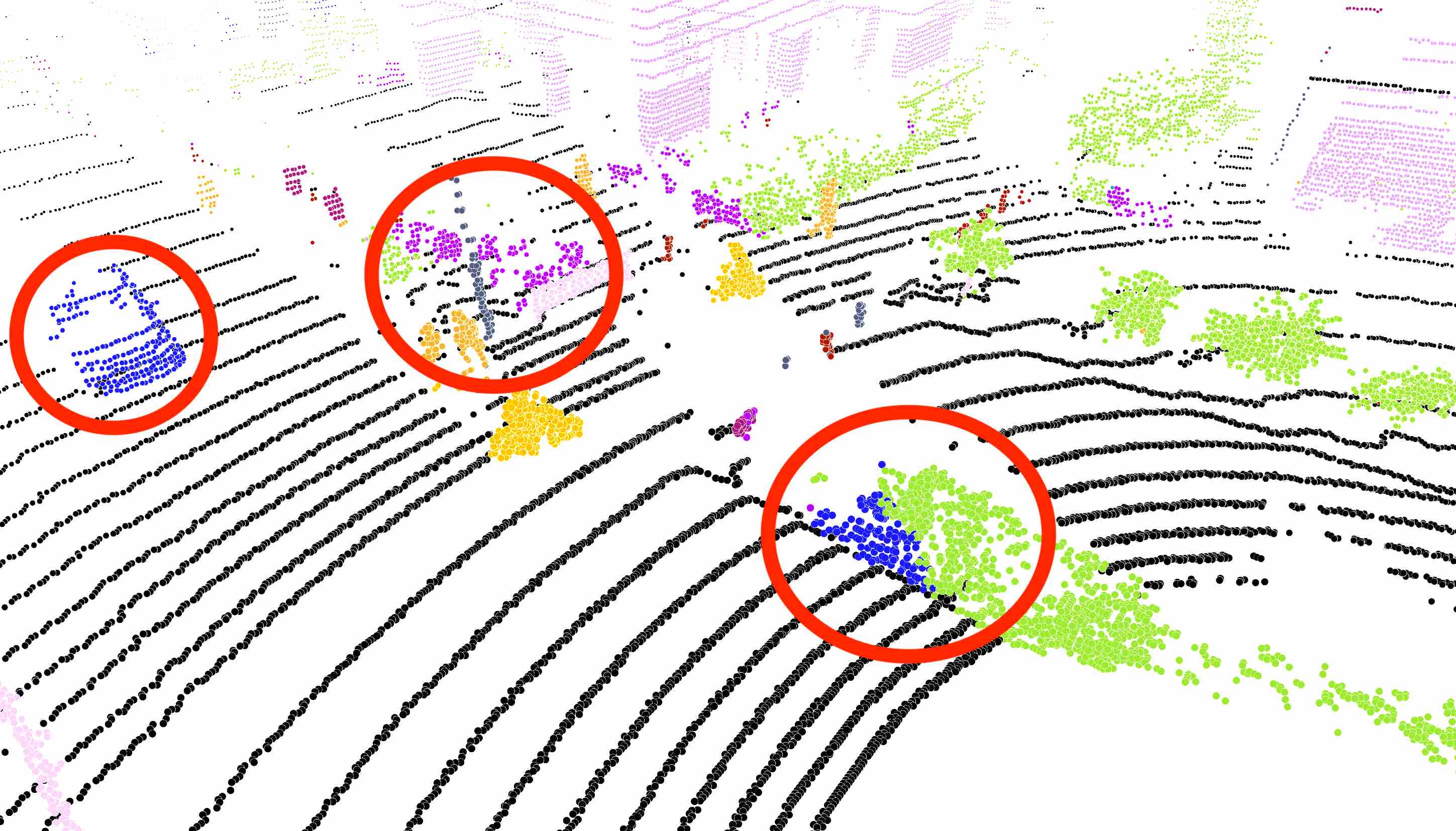}
        \end{overpic} &
        \begin{overpic}[width=0.33\textwidth]{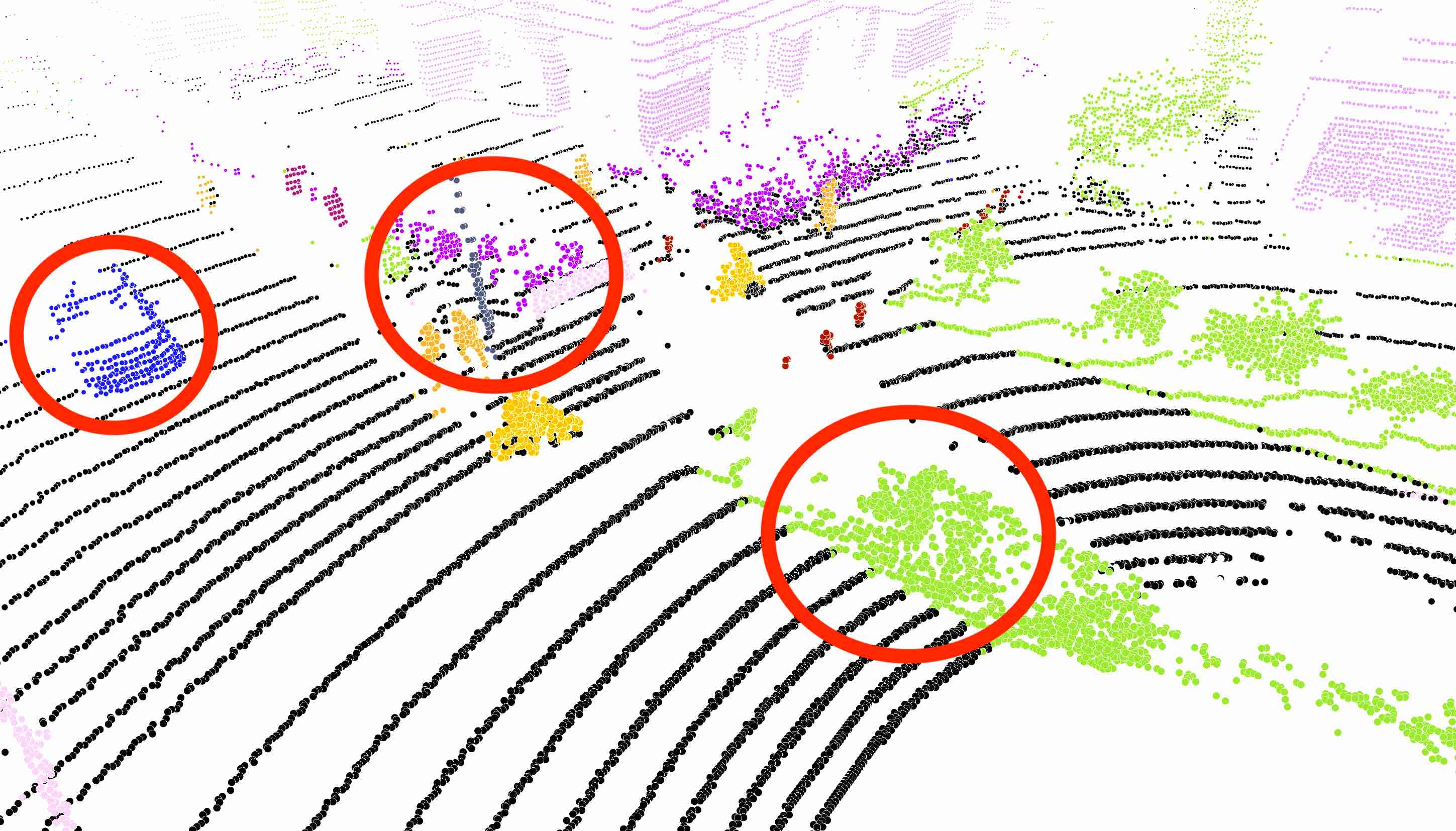}
        \end{overpic}\\
        \multicolumn{3}{c}{
        \begin{overpic}[width=0.99\textwidth]{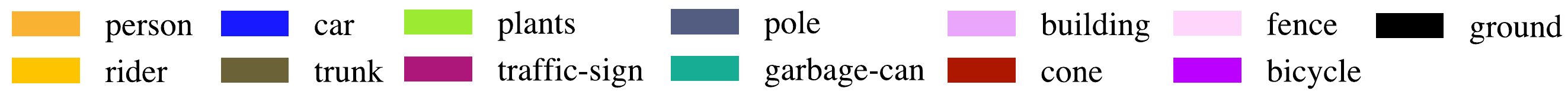}
        \end{overpic}}
    \end{tabular}
    \vspace{-.3cm}
    \caption{Results on SynLiDAR$\rightarrow$SemanticPOSS. Source predictions are often wrong and mingled in the same region. After adaptation, \ourmethod improves segmentation with homogeneous predictions and correctly assigned classes. The red circles highlight regions with interesting results.}
    \label{fig:qualitative_poss}
\end{figure}

\begin{figure}[t]
\centering
    \setlength\tabcolsep{1.pt}
    \begin{tabular}{ccc}
    \raggedright
        \begin{overpic}[width=0.33\textwidth]{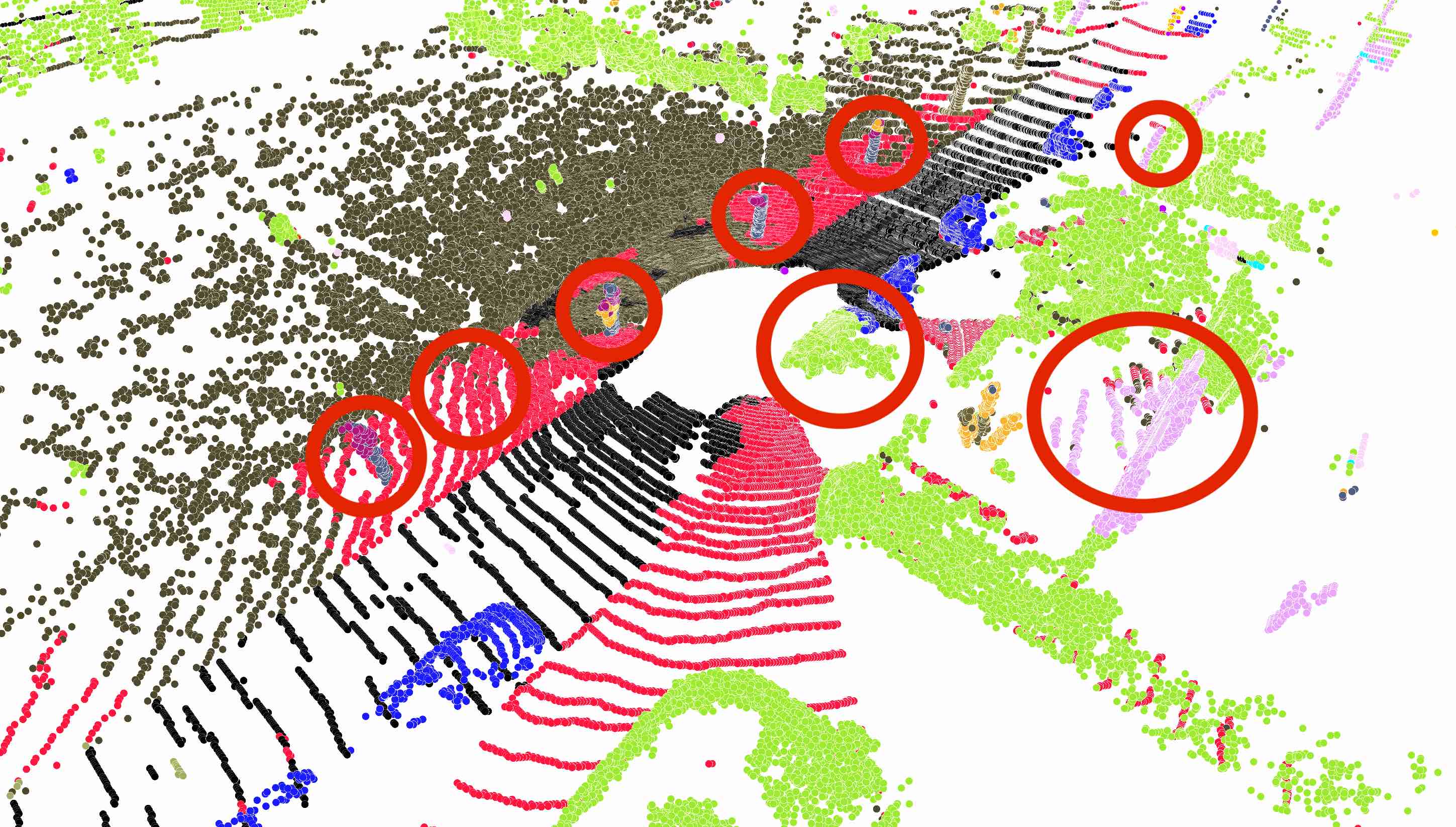}
        \put(40,67){\color{black}\footnotesize \textbf{source}}
        \put(160,67){\color{black}\footnotesize \textbf{ours}}
        \put(285,68){\color{black}\footnotesize \textbf{gt}}
        \end{overpic} &  
        \begin{overpic}[width=0.33\textwidth]{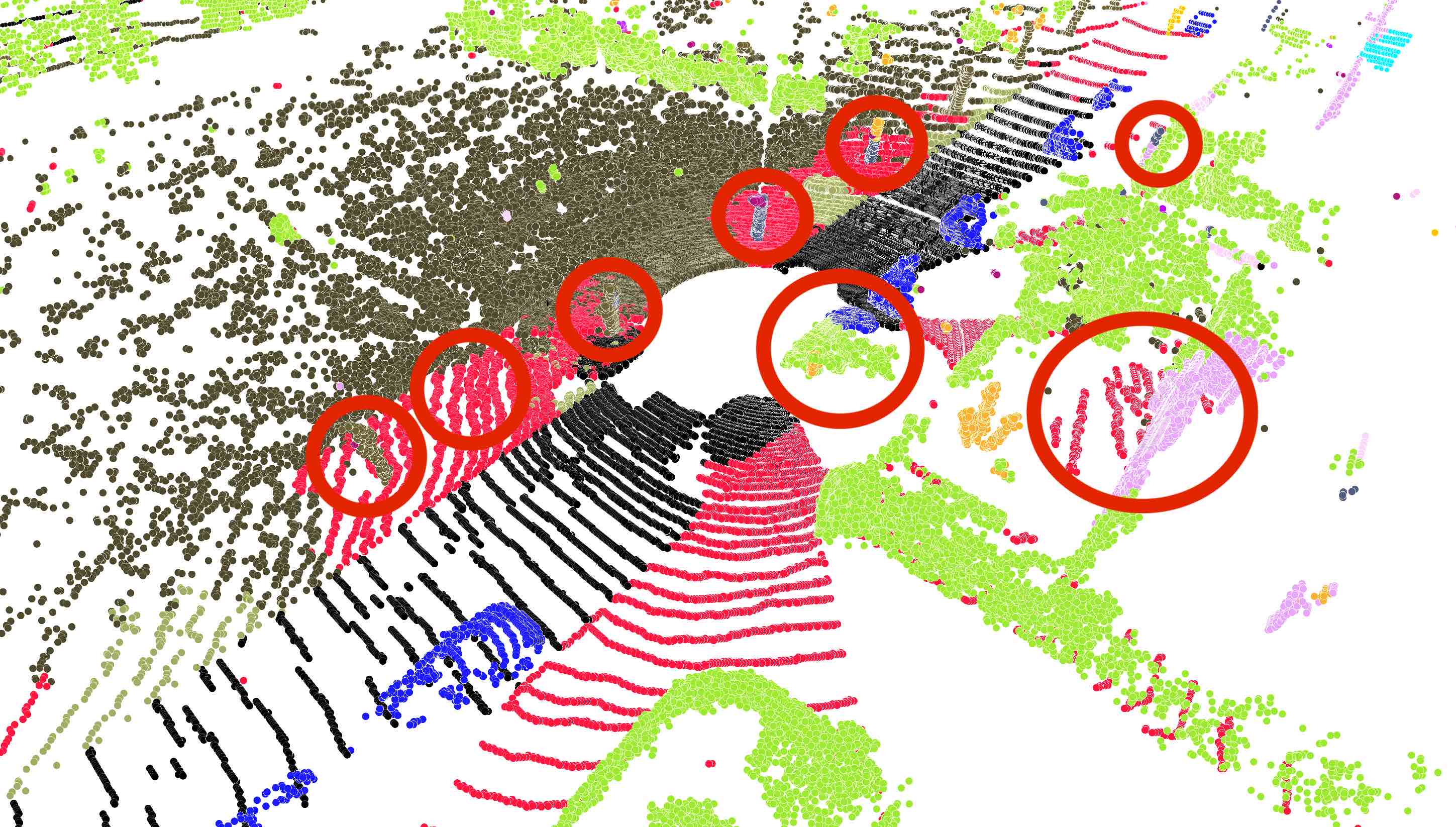}
        \end{overpic} &
        \begin{overpic}[width=0.33\textwidth]{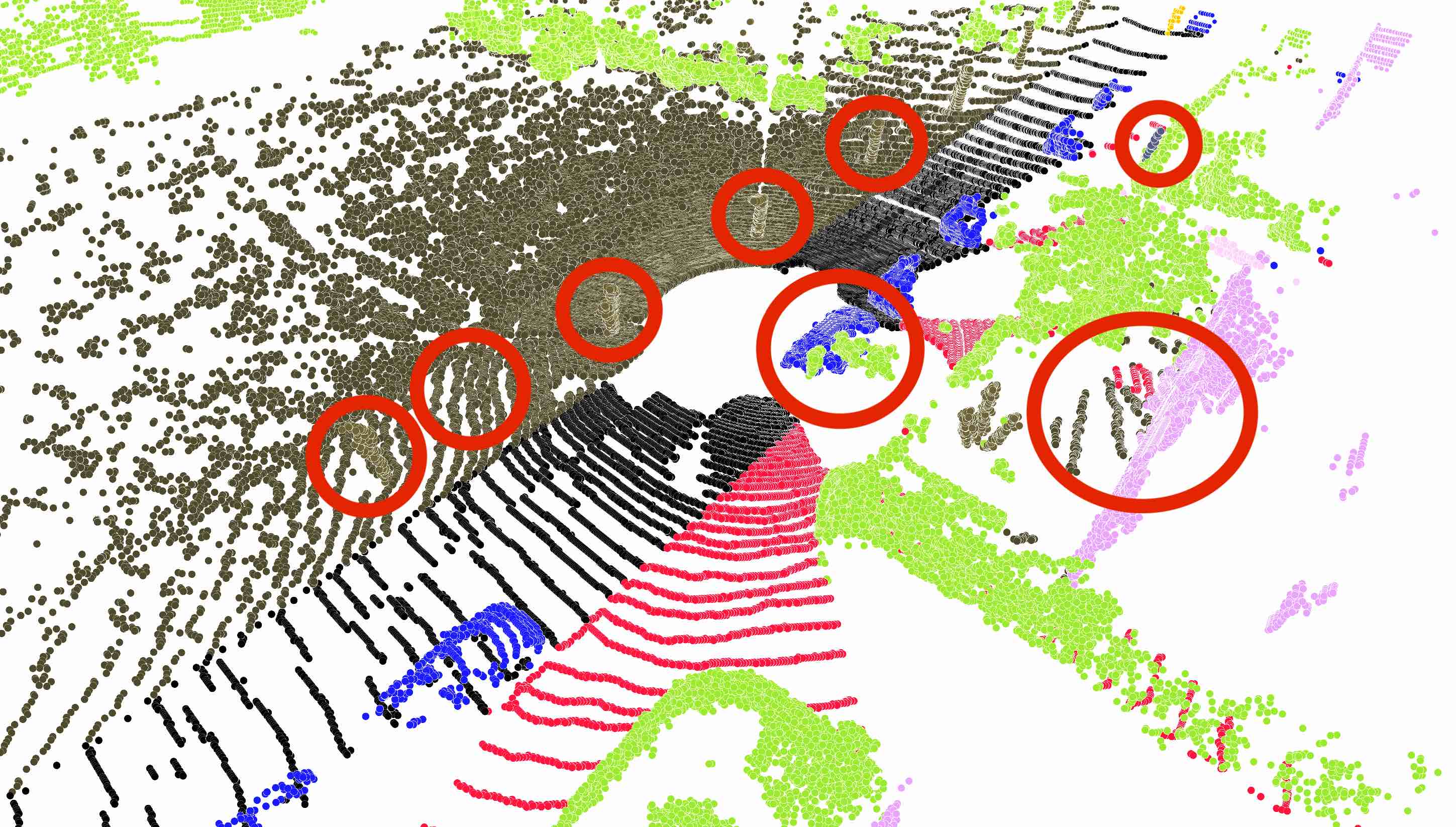}
        \end{overpic}\\
        \begin{overpic}[width=0.33\textwidth]{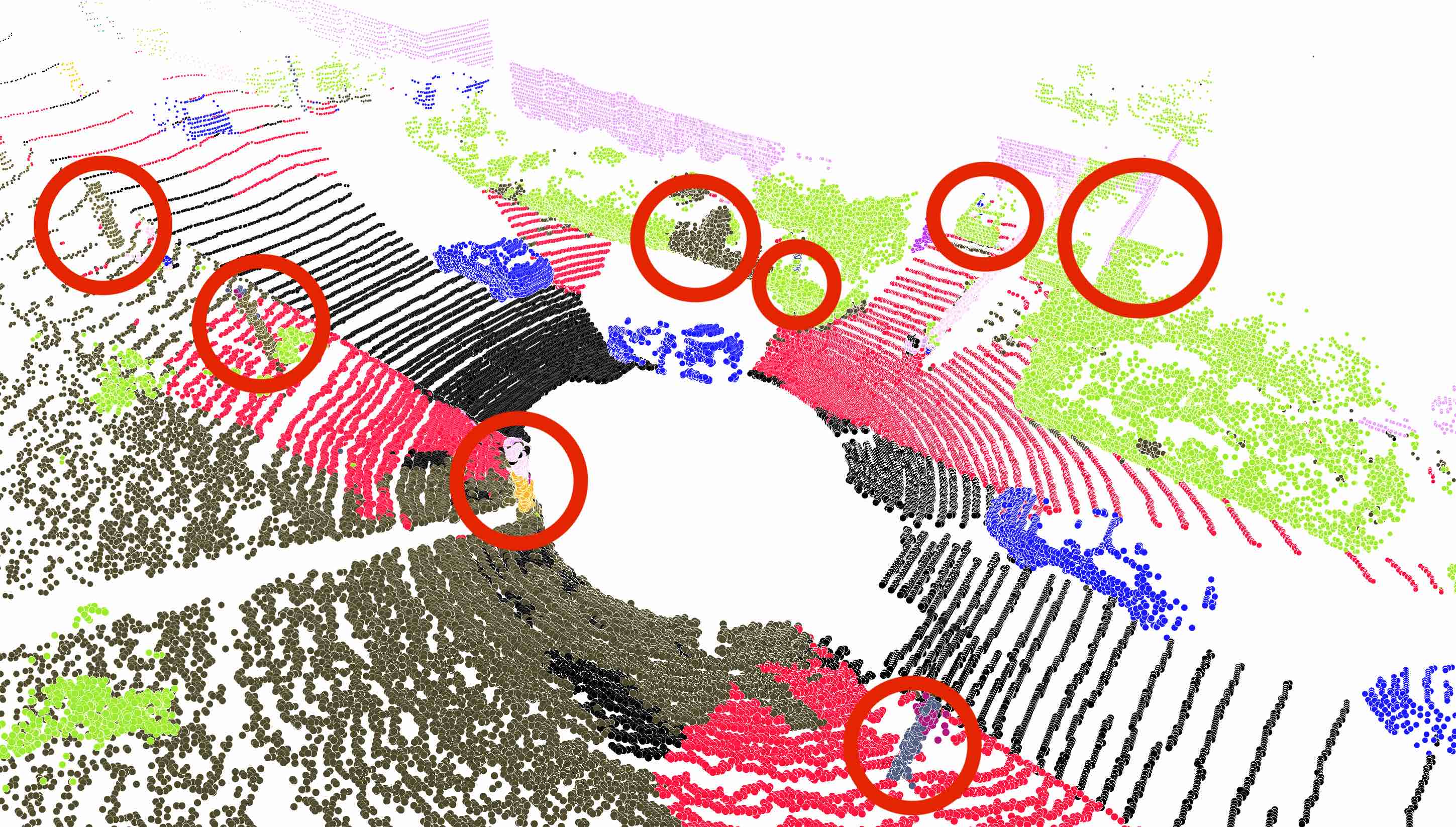}
        \end{overpic} &  
        \begin{overpic}[width=0.33\textwidth]{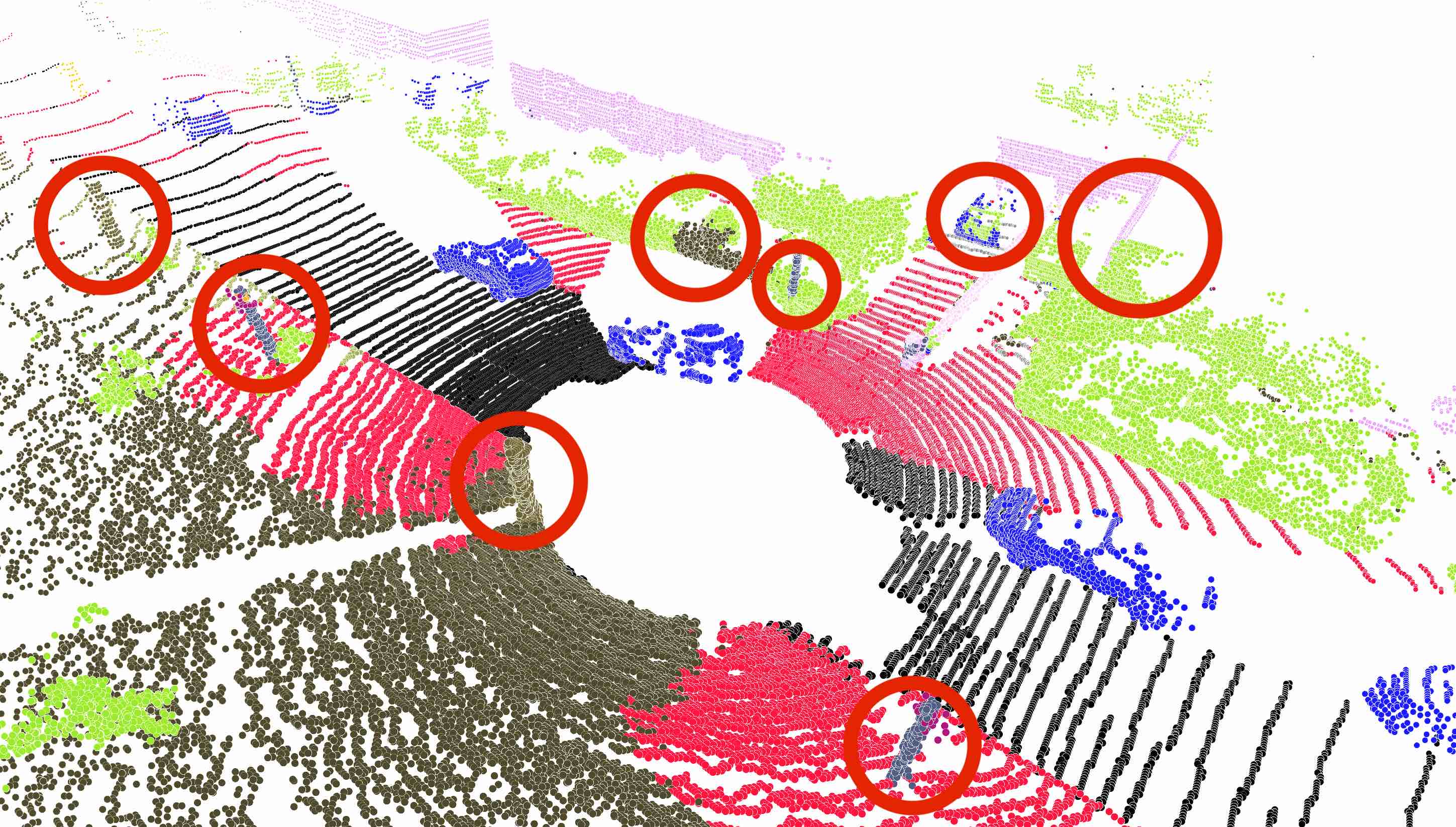}
        \end{overpic} &
        \begin{overpic}[width=0.33\textwidth]{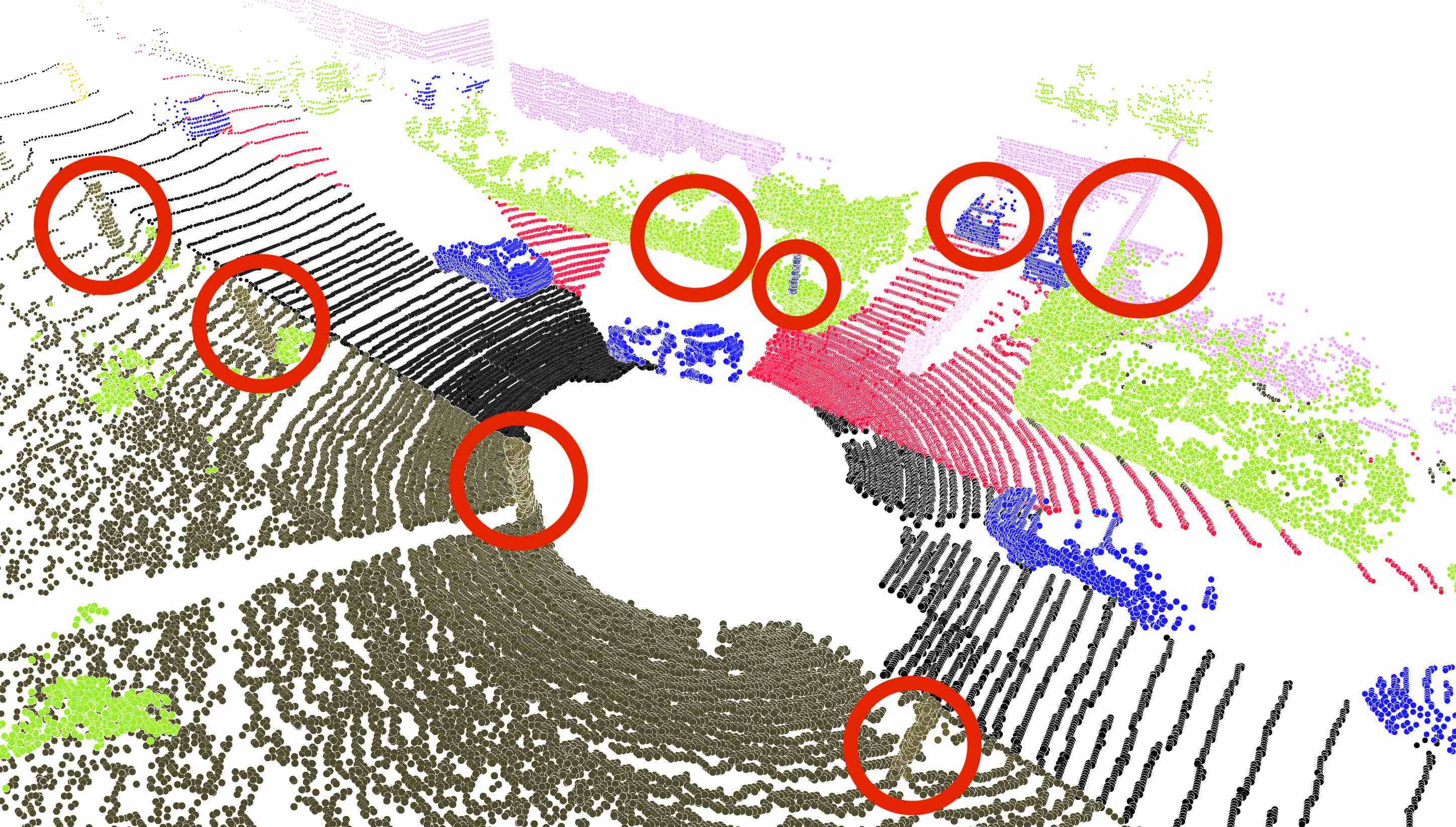}
        \end{overpic}\\
        \begin{overpic}[width=0.33\textwidth]{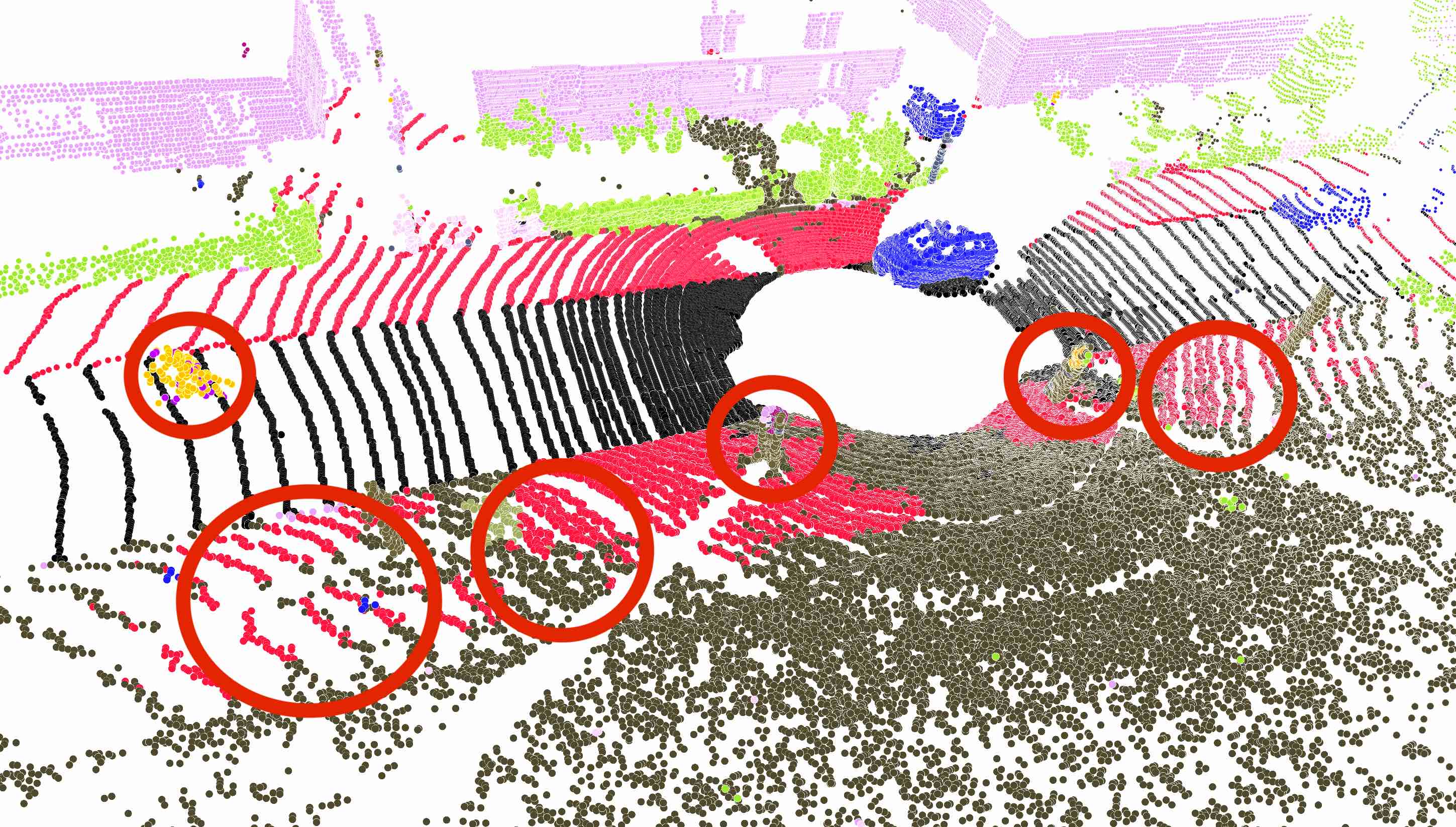}
        \end{overpic} &  
        \begin{overpic}[width=0.33\textwidth]{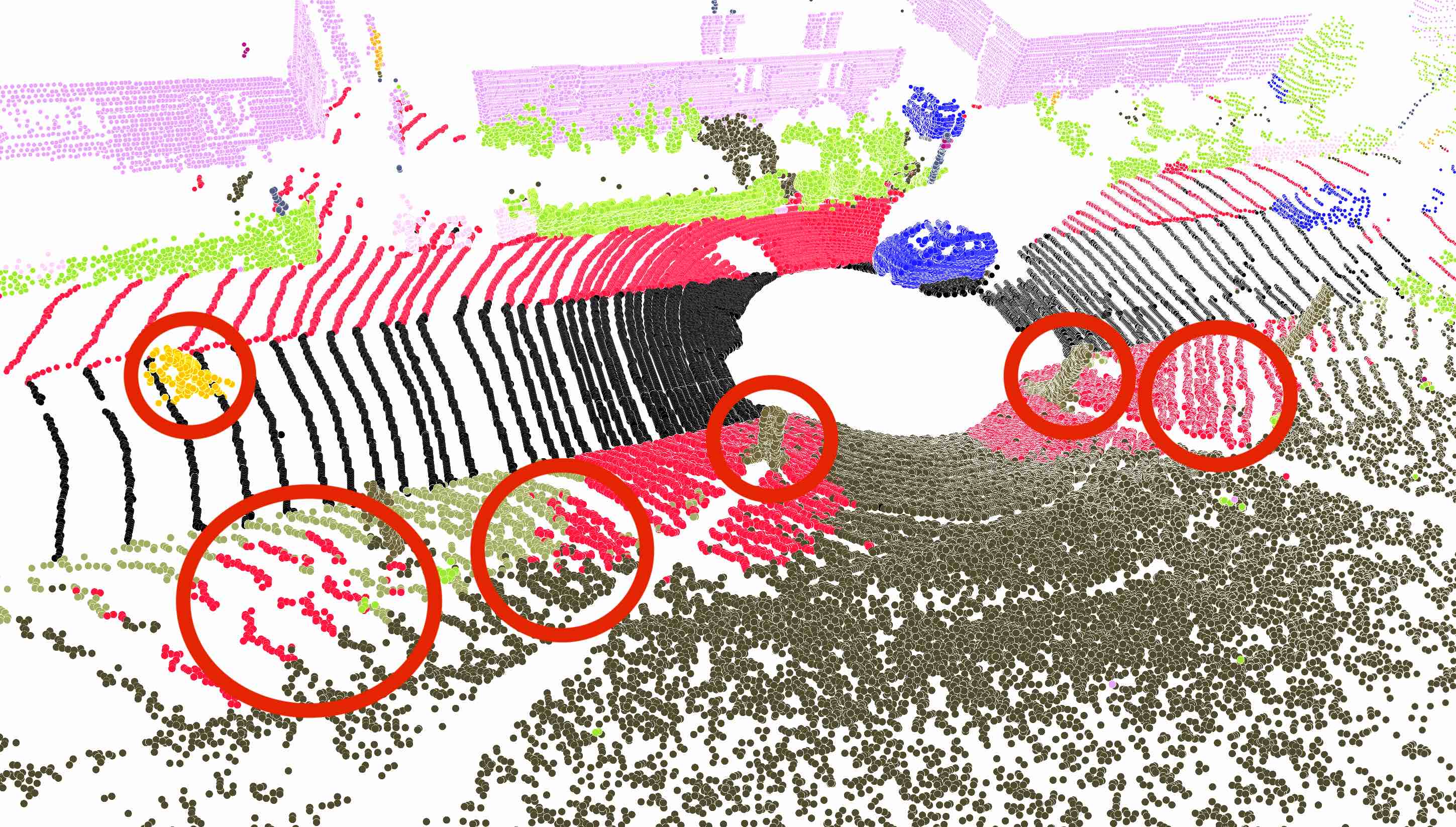}
        \end{overpic} &
        \begin{overpic}[width=0.33\textwidth]{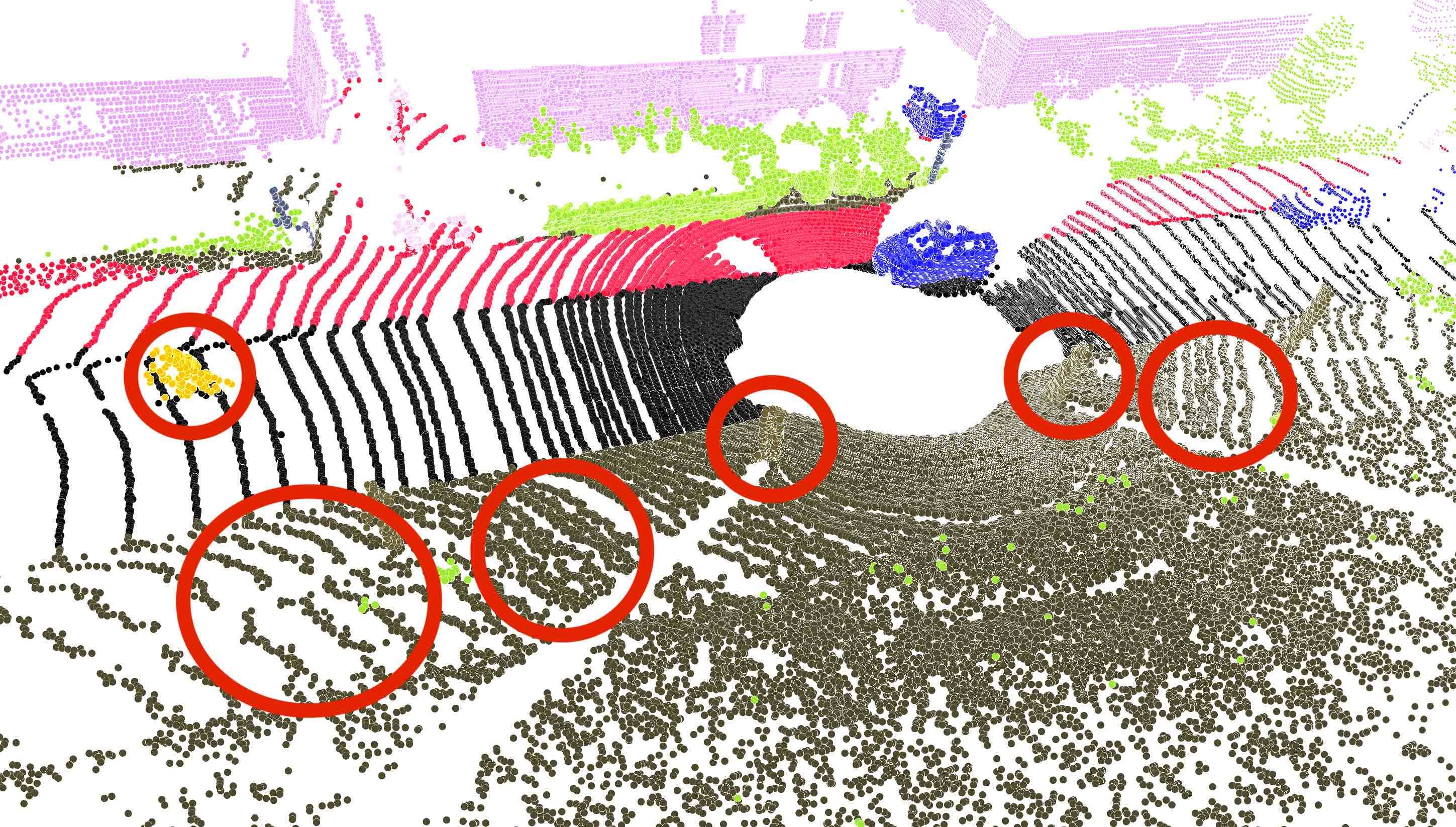}
        \end{overpic}\\
        \multicolumn{3}{c}{
        \begin{overpic}[width=0.98\textwidth]{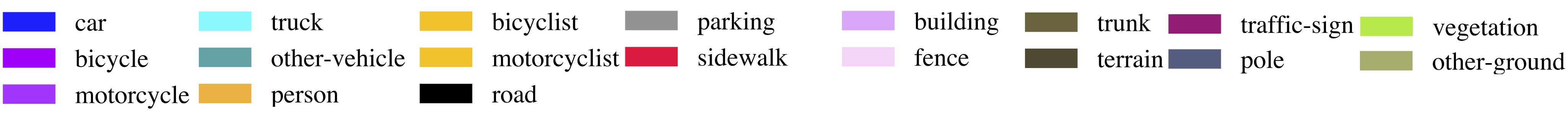}
        \end{overpic}}
    \end{tabular}
    \caption{Results on SynLiDAR$\rightarrow$SemanticKITTI. Source predictions are often wrong and mingled in the same region. After adaptation, \ourmethod improves segmentation with homogeneous predictions and correctly assigned classes. The red circles highlight regions with interesting results.}
    \label{fig:qualitative_kitti}
\end{figure}

We report qualitative examples of the adaptation performance before (source) and after \ourmethod adaptation (ours), and compare them to ground-truth annotations (gt). 
Fig.~\ref{fig:qualitative_poss} shows the adaptation results on SynLiDAR$\rightarrow$SemanticPOSS, while Fig.~\ref{fig:qualitative_kitti} show the results on SynLiDAR$\rightarrow$SemanticKITTI. 
Red circles highlight regions with interesting results.
In Fig.~\ref{fig:qualitative_poss}, improvements are visible in multiple regions of the examples. Source predictions are often not homogeneous with completely wrong regions. After adaptation, \ourmethod improves segmentation with more homogeneous regions and correctly assigned classes.
In Fig.~\ref{fig:qualitative_kitti}, source predictions are less sparse but wrong for several spatial regions. After adaptation, \ourmethod allows better and correct predictions. 
Additional examples can be found in the Supplementary Material.

%%%%%%%%%%%%%%%%%%%%%%%%%%%%%%%%%%%%%%%%%%
%%%%%%%%%%%%%%%%%%%%%%%%%%%%%%%%%%%%%%%%%%
%%%%%%%%%%%%%%%%%%%%%%%%%%%%%%%%%%%%%%%%%%
%*******************************************
\begin{table}[t]
    \centering
    \caption{Ablation study of the \ourmethod components: mixing strategy ($t\rightarrow s$ and $s\rightarrow t$), compositional mix augmentations (local $h$ and global $r$), mean teacher update ($\beta$) and, weighted class selection in semantic selection ($f$). Each combination is named with a different version (a-h). Source performance are added as lower bound and highlighted in gray to facilitate the reading.}
    \label{tab:ablation_components}
    \vspace{.2cm}
    \tabcolsep 5pt
    \resizebox{.55\columnwidth}{!}{%
    \begin{tabular}{c|cc|cc|cc|c}
        \toprule
        \ourmethod & \multicolumn{2}{c|}{mix} & \multicolumn{2}{c|}{augs} & \multicolumn{2}{c|}{others} & \\
         version & $t \rightarrow s$ & $s \rightarrow t$ & $h$ & $r$ & $\beta$ & $f$ & \textbf{mIoU}\\
        \midrule
        \CC{sourcecolor}Source & \CC{sourcecolor}- & \CC{sourcecolor}- & \CC{sourcecolor}- & \CC{sourcecolor}- & \CC{sourcecolor}- & \CC{sourcecolor}- & \CC{sourcecolor}20.7\\
        \midrule
        (a) & \ding{51} & & & & & & 31.6\\
        (b) & \ding{51} & & & & \ding{51} & & 31.9\\
        (c) & \ding{51} & \ding{51} &  & & & & 35.0\\
        (d) & \ding{51} & \ding{51} & & &\ding{51} & & 35.4\\
        (e) & \ding{51} & \ding{51} &\ding{51} & &\ding{51} & & 36.8\\
        (f) & \ding{51} & \ding{51} & & \ding{51} &\ding{51} & & 37.3\\
        (g) & \ding{51} & \ding{51} &\ding{51} & \ding{51} & \ding{51} & & 39.0\\
        (h) & \ding{51} & \ding{51} &\ding{51} & \ding{51} & & \ding{51} & 39.1\\
        \midrule
        Full & \ding{51} & \ding{51} & \ding{51} & \ding{51} & \ding{51} & \ding{51} & \textbf{40.4}\\
        \bottomrule
    \end{tabular}
    }
\end{table}
%*******************************************

\section{Ablation study}\label{sec:ablations}

We perform an ablation study of \ourmethod by using the SynLiDAR $\rightarrow$ SemanticPOSS setup.
We compare our mixing approach with a recent point cloud mixing strategy~\cite{Nekrasov213DV} by applying it to the synthetic-to-real setting (Sec.~\ref{sec:mix_up}).
In Sec.~\ref{sec:confidence}, we investigate the importance of confidence threshold in \ourmethod.

%%%%%%%%%%%%%%%%%%%%%%%%%%%%%%%%%%%%%%%%%%
%%%%%%%%%%%%%%%%%%%%%%%%%%%%%%%%%%%%%%%%%%
\subsection{Method components}\label{sec:components}

We analyze \ourmethod by organizing its components into three groups: mixing strategies (\textit{mix}), augmentations (\textit{augs}) and other components (\textit{others}).
In the \textit{mix} group, we assess the importance of the mixing strategies ($t\rightarrow s$ and $s\rightarrow t$) used in our compositional mix (Sec.~\ref{sec:compositional_mix}) after semantic selection. 
In the \textit{augs} group, we assess the importance of the local $h$ and global $r$ augmentations that are used in the compositional mix (Sec.~\ref{sec:compositional_mix}). 
In the \textit{others} group, we assess the importance of the mean teacher update ($\beta$) (Sec.~\ref{sec:network_update}) and of the long-tail weighted sampling $f$ (Sec.~\ref{sec:semantic_selection}).
When the $t \rightarrow s$ branch is active, also the pseudo-label filtering $g$ is utilized, while when $f$ is not active, $\alpha=0.5$ source classes are selected randomly.
With different combinations of components, we obtain different versions of \ourmethod which we name \ourmethod (a-h).
The complete version of our method is named \textit{Full}, where all the components are activated. 
The Source performance (Source) is also added as a reference for the lower bound.
See Tab.~\ref{tab:ablation_components} for the definition of these different versions.

When the $t \rightarrow s$ branch is used, \ourmethod (a) achieves an initial $31.6$ mIoU showing that the $t \rightarrow s$ branch provides a significant adaptation contribution over the Source. 
When we also use the $s \rightarrow t$ branch and the mean teacher $\beta$, \ourmethod (b-d) further improve performance achieving a $35.4$ mIoU. 
By introducing local and global augmentations in \ourmethod (e-h), we can improve performance up to $39.1$ mIoU. 
The best performance of $40.4$ mIoU is achieved with \ourmethod Full where all the components are activated.

%%%%%%%%%%%%%%%%%%%%%%%%%%%%%%%%%%%%%%%%%%
%%%%%%%%%%%%%%%%%%%%%%%%%%%%%%%%%%%%%%%%%%
\subsection{Point Cloud Mix}\label{sec:mix_up}
We compare \ourmethod with Mix3D~\cite{Nekrasov213DV} and PointCutMix~\cite{zhang2021pointcutmix} to show the effectiveness of the different mixing designs. 
As per our knowledge, Mix3D~\cite{Nekrasov213DV} is the only mixup strategy designed for 3D semantic segmentation, while PointCutMix is the only strategy for mixing portions of different point clouds.
We implement Mix3D~\cite{Nekrasov213DV} and PointCutMix~\cite{zhang2021pointcutmix} based on authors descriptions: we concatenate point clouds (random crops for PointCutMix) of the two domains, i.e., $\mathcal{X}^s$ and $\mathcal{X}^t$, as well as their labels and pseudo-labels, i.e., $\mathcal{Y}^s$ and $\hat{\mathcal{Y}}^t$, respectively.
\ourmethod double is our two-branch network with sample mixing.
For a fair comparison, we deactivate the weighted sampling and the mean teacher update.
We keep local and global augmentations ($h$ and $r$) activated.

\begin{figure}[t]
     \centering
     \begin{subfigure}[b]{0.45\textwidth}
         \centering
         \includegraphics[width=\textwidth]{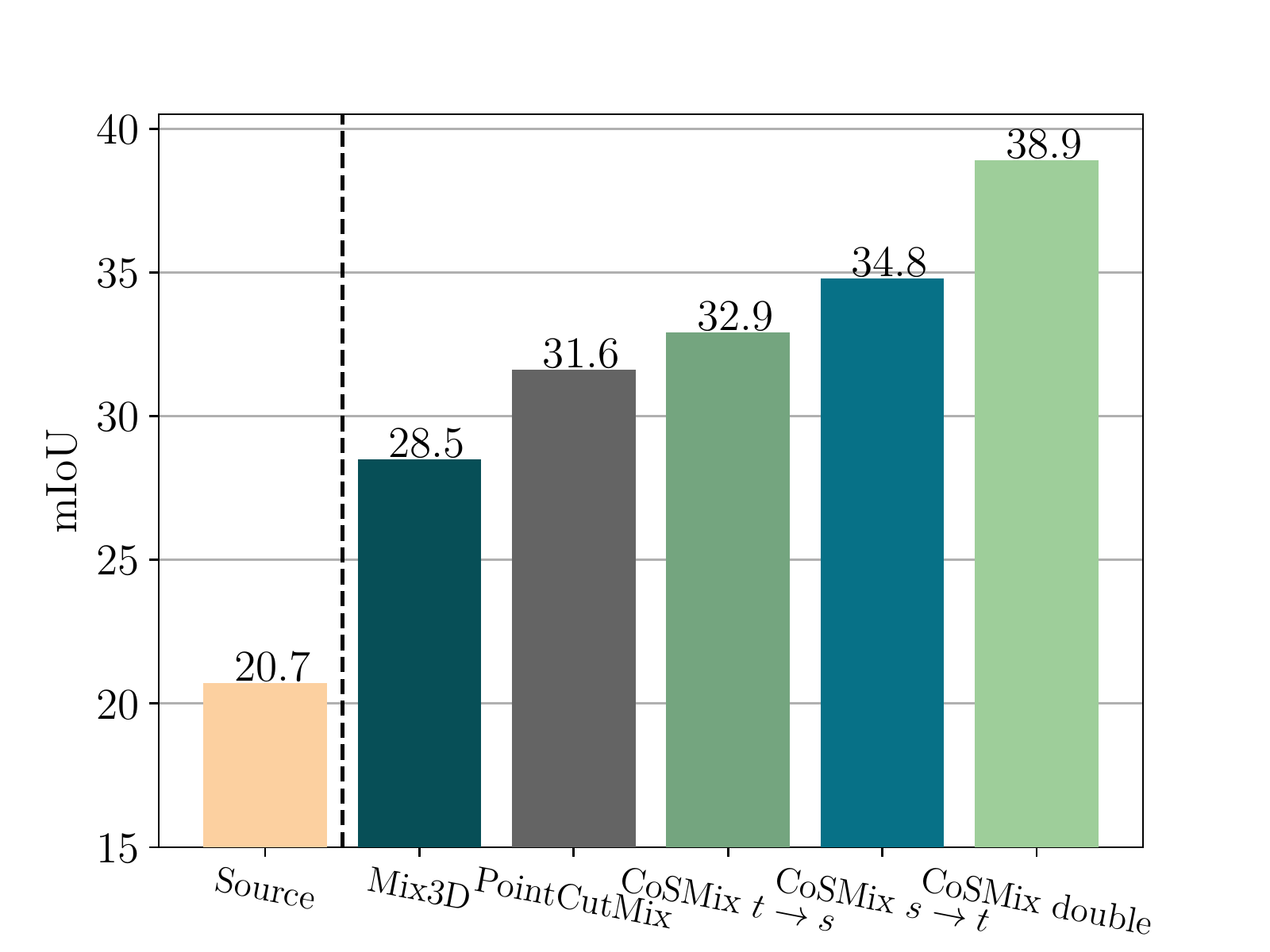}
         \caption{}
        %  \caption{Mix up strategies.}
         \label{fig:ablation_mix}
     \end{subfigure}
     \hfill
     \begin{subfigure}[b]{0.45\textwidth}
         \centering
         \includegraphics[width=\textwidth]{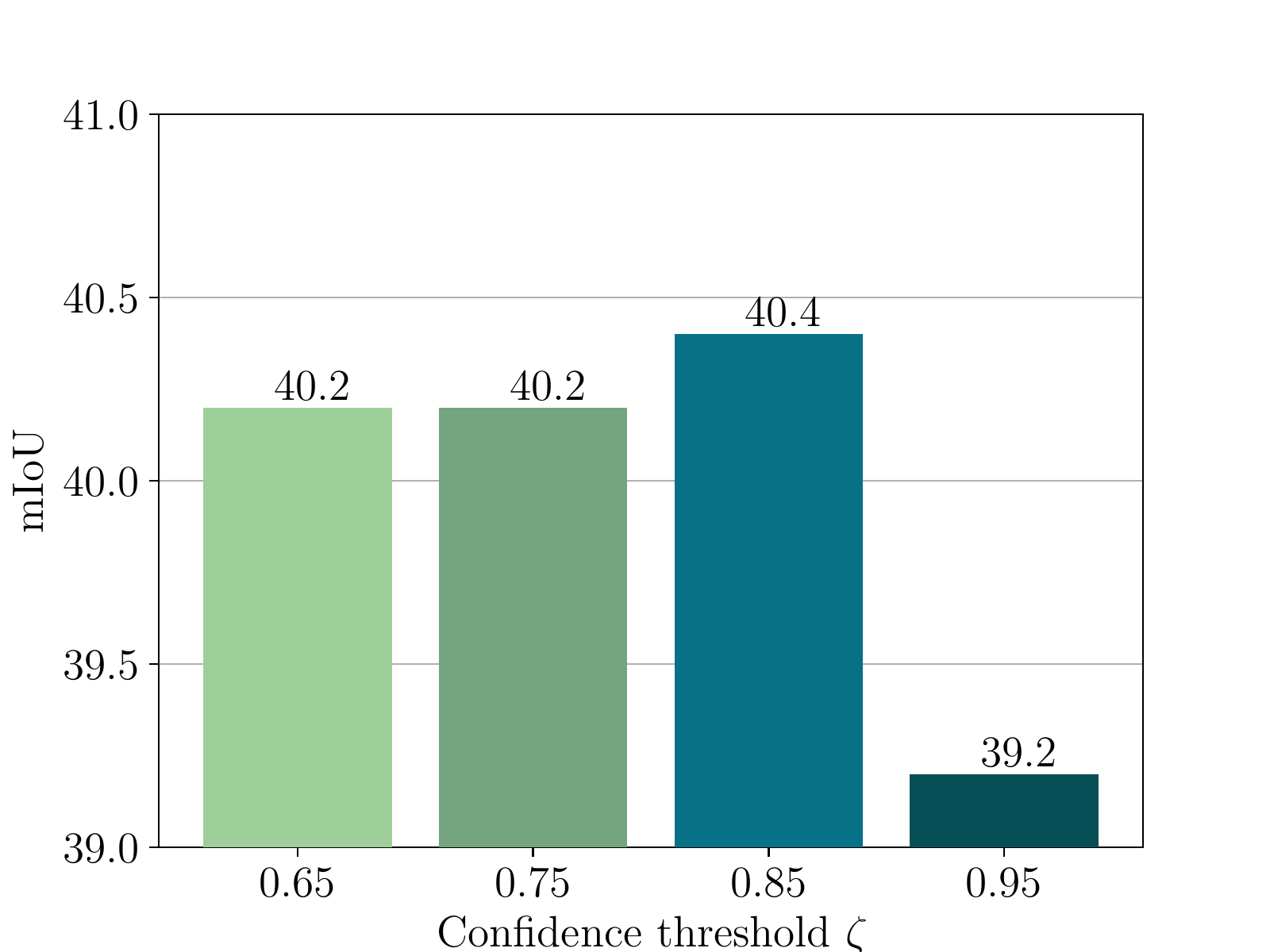}
        %  \caption{Confidence threshold.}
        \caption{}
         \label{fig:ablation_confidence}
     \end{subfigure}
     \vspace{-.3cm}
    \caption{Comparison of the adaptation performance with (a) different point cloud mix up strategies and (b) on confidence threshold values. (a) Compared to the recent mixing strategy Mix3D \cite{Nekrasov213DV}, our mixing strategy and its variations achieve superior performance. (b) Adaptation results show that $\zeta$ should be set such that to achieve a trade-off between pseudo-label correctness and object completeness.}
    \label{fig:three graphs}
\end{figure}

Fig.~\ref{fig:three graphs} shows that Mix3D~\cite{Nekrasov213DV} outperforms the Source model, achieving $28.5$ mIoU, while PointCutMix~\cite{chen2020pointmixup} achieves $31.6$ mIoU. 
When we use the $t \rightarrow s$ branch alone we can achieve $32.9$ mIoU and when we use the $s \rightarrow t$ branch alone, \ourmethod can further improve the results, achieving $34.8$ mIoU.
This shows that the supervision from the source to target is effective for adaptation on the target domain.
When we use the contribution from both branches simultaneously, \ourmethod achieves the best result with $38.9$ mIoU.

%++++++++++++++++++++++++++++++++++++++++++++++
%++++++++++++++++++++++++++++++++++++++++++++++

%%%%%%%%%%%%%%%%%%%%%%%%%%%%%%%%%%%%%%%%%%
%%%%%%%%%%%%%%%%%%%%%%%%%%%%%%%%%%%%%%%%%%
\subsection{Pseudo label filtering}\label{sec:confidence}

We investigate the robustness of \ourmethod to increasingly noisier pseudo-labels and study the importance of setting the correct confidence threshold $\zeta$ for pseudo-label distillation in $g$ (Sec.~\ref{sec:semantic_selection}). We repeat the experiments with a confidence threshold from $0.65$ to $0.95$ and report the obtained adaptation performance in Fig.~\ref{fig:three graphs}.
\ourmethod is robust to noisy pseudo-labels reaching a $40.2$ mIoU with the low threshold of $0.65$. 
The best adaptation performance of $40.4$ mIoU is achieved with a confidence threshold of $0.85$.
By using a high confidence threshold of $0.95$ performance is affected reaching $39.2$ mIoU. 
With this configuration, too few pseudo-labels are selected to provide an effective contribution for the adaptation.

%%%%%%%%%%%%%%%%%%%%%%%%%%%%%%%%%%%%%%%%%%%%%%%%%%%%%%
%%%%%%%%%%%%%%%%%%%%%%%%%%%%%%%%%%%%%%%%%%%%%%%%%%%%%%
%%%%%%%%%%%%%%%%%%%%%%%%%%%%%%%%%%%%%%%%%%%%%%%%%%%%%%
\section{Conclusions}
In this paper, we proposed the first UDA method for 3D semantic segmentation based on a novel 3D point cloud mixing strategy that exploits semantic and structural information concurrently.
We performed an extensive evaluation in the synthetic-to-real UDA scenario by using large-scale publicly available LiDAR datasets.
Experiments showed that our method outperforms all the compared state-of-the-art methods by a large margin. 
Furthermore, in-depth studies highlighted the importance of each \ourmethod component and that our mixing strategy is beneficial for solving domain shift in 3D LiDAR segmentation.
Future research directions may include the introduction of self-supervised learning tasks and the extension of \ourmethod to source-free adaptation tasks.

\vspace{.2cm}
\noindent\textbf{Acknowledgements.} This work was partially supported by OSRAM GmbH, by the Italian Ministry of Education, Universities and Research (MIUR) ``Dipartimenti di Eccellenza 2018-2022'', by the EU JPI/CH SHIELD project, by the PRIN project PREVUE (Prot. 2017N2RK7K), the EU ISFP PROTECTOR (101034216) project and the EU H2020 MARVEL (957337) project and, it was carried out in the Vision and Learning joint laboratory of FBK and UNITN.

\clearpage
% ---- Bibliography ----
%
% BibTeX users should specify bibliography style 'splncs04'.
% References will then be sorted and formatted in the correct style.
%
\bibliographystyle{splncs04}
\bibliography{egbib}
\end{document}

\sloppy
% \renewcommand\thelinenumber{\color[rgb]{0.2,0.5,0.8}\normalfont\sffamily\scriptsize\arabic{linenumber}\color[rgb]{0,0,0}}
% \renewcommand\makeLineNumber {\hss\thelinenumber\ \hspace{6mm} \rlap{\hskip\textwidth\ \hspace{6.5mm}\thelinenumber}}
% \linenumbers
\pagestyle{headings}
\mainmatter
\def\ECCVSubNumber{2050}  % Insert your submission number here

\title{Supplementary Material\\
CoSMix: Compositional Semantic Mix for Domain Adaptation in 3D LiDAR Segmentation} % Replace with your title

% INITIAL SUBMISSION
%\begin{comment}
% \titlerunning{ECCV-22 submission ID \ECCVSubNumber} 
% \authorrunning{ECCV-22 submission ID \ECCVSubNumber} 
% \author{Anonymous ECCV submission}
% \institute{Paper ID \ECCVSubNumber}
%\end{comment}
%******************
\titlerunning{CoSMix: Compositional Semantic Mix for DA in 3D LiDAR Segmentation}
% If the paper title is too long for the running head, you can set
% an abbreviated paper title here
%
% \author{Cristiano Saltori\inst{1}\orcidlink{0000-0001-9583-4160
% } \and
% Fabio Galasso\inst{2}\orcidlink{0000-0003-1875-7813} \and
% Giuseppe Fiameni\inst{3}\orcidlink{0000-0001-8687-6609} \and \\
% Nicu Sebe\inst{1}\orcidlink{0000-0002-6597-7248} \and
% Elisa Ricci\inst{1,4}\orcidlink{0000-0002-0228-1147} \and
% Fabio Poiesi\inst{4}\orcidlink{0000-0002-9769-1279}
% }

\author{Cristiano Saltori\inst{1} \and
Fabio Galasso\inst{2} \and
Giuseppe Fiameni\inst{3} \and \\
Nicu Sebe\inst{1} \and
Elisa Ricci\inst{1,4} \and
Fabio Poiesi\inst{4}
}
\authorrunning{C. Saltori et al.}
% First names are abbreviated in the running head.
% If there are more than two authors, 'et al.' is used.
%
\institute{University of Trento, Trento, Italy \and
Sapienza University of Rome, Rome, Italy \and
NVIDIA AI Technology Center \and
Fondazione Bruno Kessler, Trento, Italy\\
\email{cristiano.saltori@unitn.it}}
%******************
\maketitle
\section{Introduction}
We provide the supplementary material in support of our main paper.
This document is organized as follows:
\begin{itemize}
    \item Sec.~\ref{sec:real2real} reports preliminary results on the real$\rightarrow$real UDA setup of SemanticPOSS$\rightarrow$SemanticKITTI with \ourmethod.
    \item Sec.~\ref{sec:mixed_samples} provides qualitative examples of our mixed point clouds $\mathcal{X}^{s\rightarrow t}$ and $\mathcal{X}^{t\rightarrow s}$ from SynLiDAR\cite{synlidar} to SemanticKITTI~\cite{behley2019iccv}.
    \item Sec.~\ref{sec:supplementary_qualitative} reports additional qualitative results on SemanticPOSS~\cite{pan2020semanticposs} and SemanticKITTI~\cite{behley2019iccv}.

\end{itemize}

\section{Real to real adaptation performance}
\label{sec:real2real}
Domain adaptation between different real datasets requires their classes to be compatible.
SemanticPOSS and SemanticKITTI are labelled neither by using the same semantic classes nor by following the same annotation protocol. We created a mapping between SemanticPOSS and SemanticKITTI classes to train the source model and to adapt it with \ourmethod. Specifically, during the experiment we consider only the semantic classes \textit{person}, \textit{car/vehicle}, \textit{trunk}, \textit{plants}, \textit{traffic-sign}, \textit{pole}, \textit{building}, \textit{fence} and \textit{ground}.
Tab.~\ref{tab:real2real} reports some preliminary results of this experiment.
%-----------------------------------
\begin{table}
\caption{Preliminary results of \ourmethod on SemanticPOSS$\rightarrow$SemanticKITTI.}
\label{tab:real2real}
\centering
\begin{tabular}{l|c|c}
\toprule
Method & \textbf{Source} & \textbf{\ourmethod} \\
\midrule
mIoU & 22.5 & 26.4\\
\bottomrule
\end{tabular}
\end{table}
%-----------------------------------

\section{Qualitative mixed samples}\label{sec:mixed_samples}
\begin{figure}[t]
\centering
    \setlength\tabcolsep{1.pt}
    \begin{tabular}{ccc}
    \raggedright
        
        \begin{overpic}[width=0.32\textwidth]{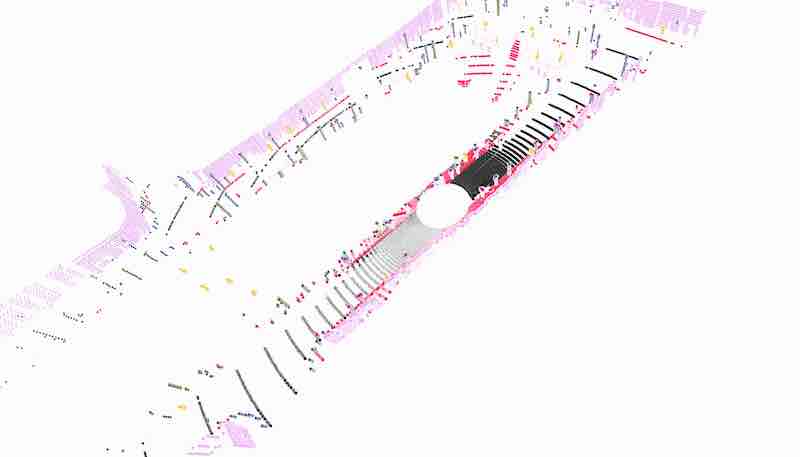}
        \put(40,67){\color{black}\footnotesize \textbf{source}}
        \put(160,67){\color{black}\footnotesize \textbf{$s \rightarrow t$}}
        \put(285,68){\color{black}\footnotesize \textbf{target}}
        \end{overpic} &  
        \begin{overpic}[width=0.32\textwidth]{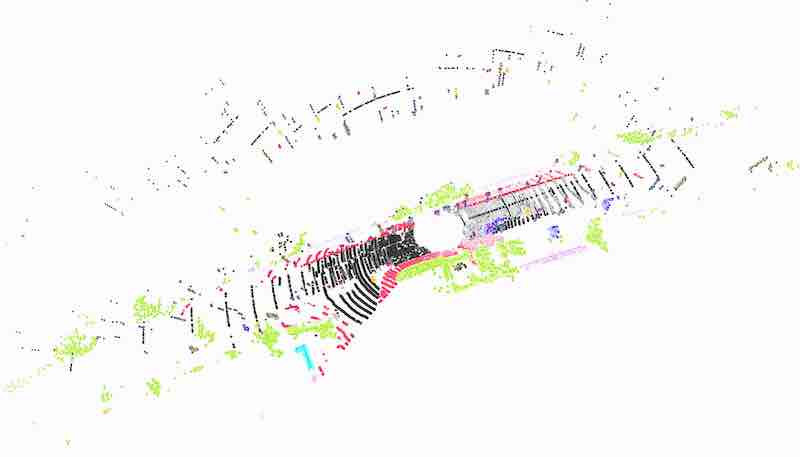}
        \end{overpic} &
        \begin{overpic}[width=0.32\textwidth]{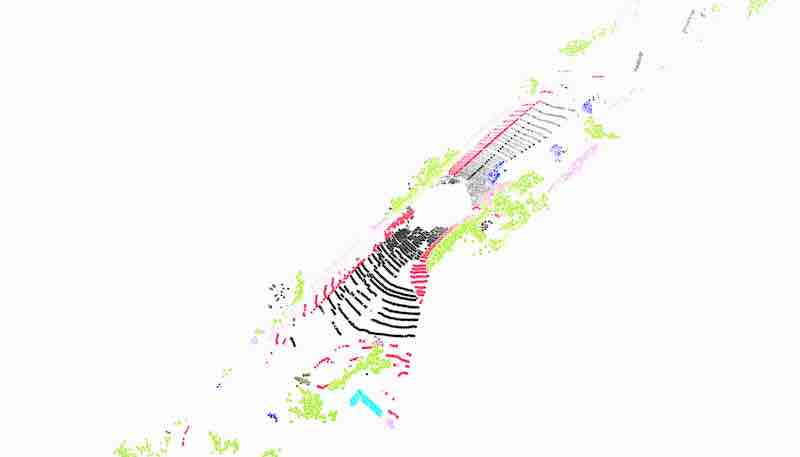}
        \end{overpic}\\
        \begin{overpic}[width=0.32\textwidth]{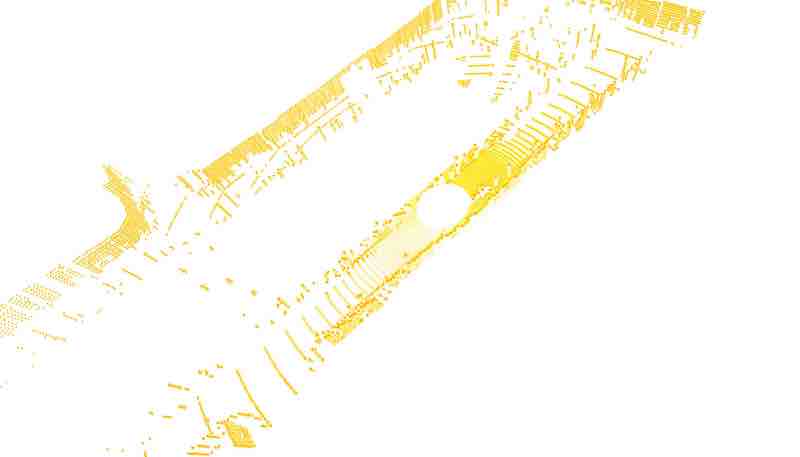}
        \end{overpic} &  
        \begin{overpic}[width=0.32\textwidth]{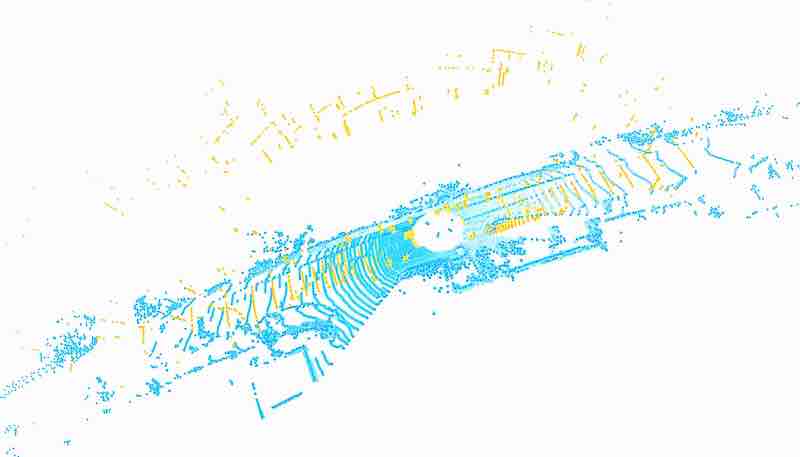}
        \end{overpic} &
        \begin{overpic}[width=0.32\textwidth]{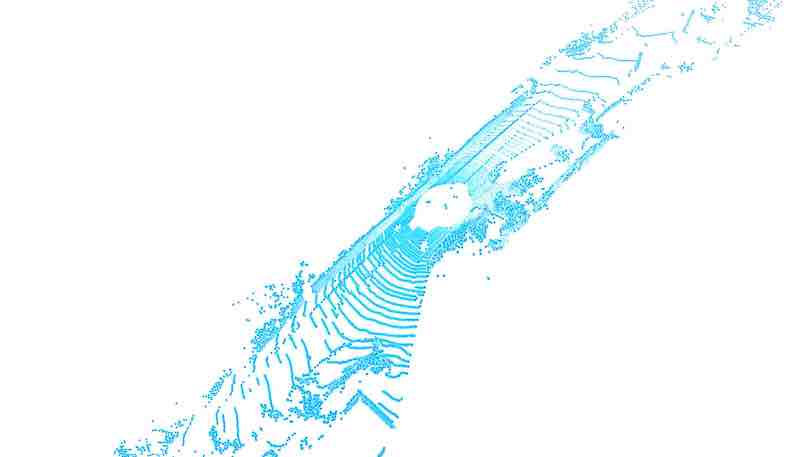}
        \end{overpic}\\
        \midrule
        \begin{overpic}[width=0.32\textwidth]{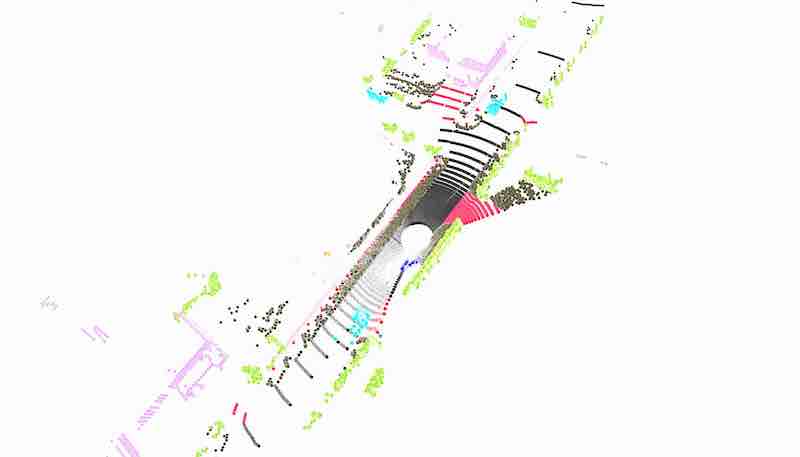}
        \end{overpic} &  
        \begin{overpic}[width=0.32\textwidth]{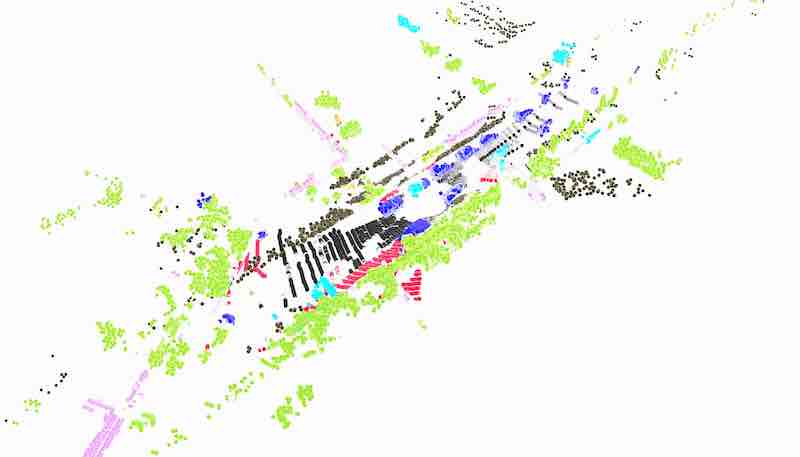}
        \end{overpic} &
        \begin{overpic}[width=0.32\textwidth]{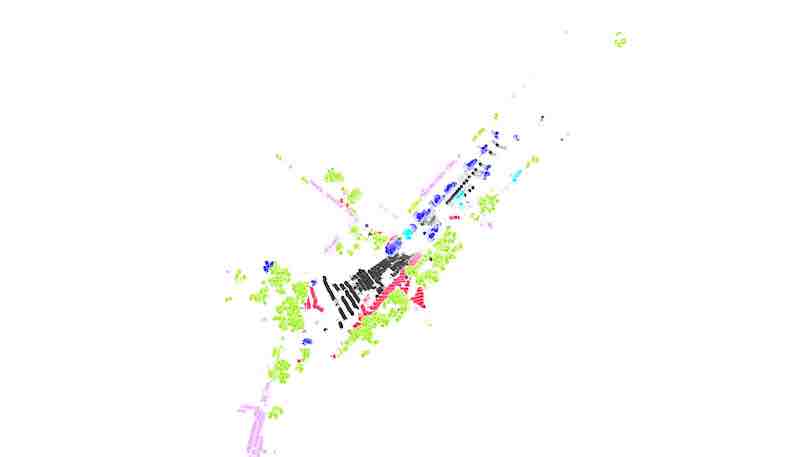}
        \end{overpic}\\
        \begin{overpic}[width=0.32\textwidth]{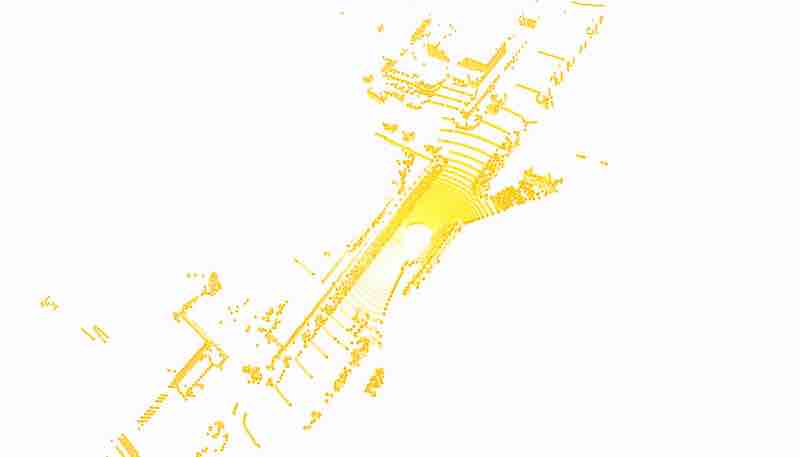}
        \end{overpic} &  
        \begin{overpic}[width=0.32\textwidth]{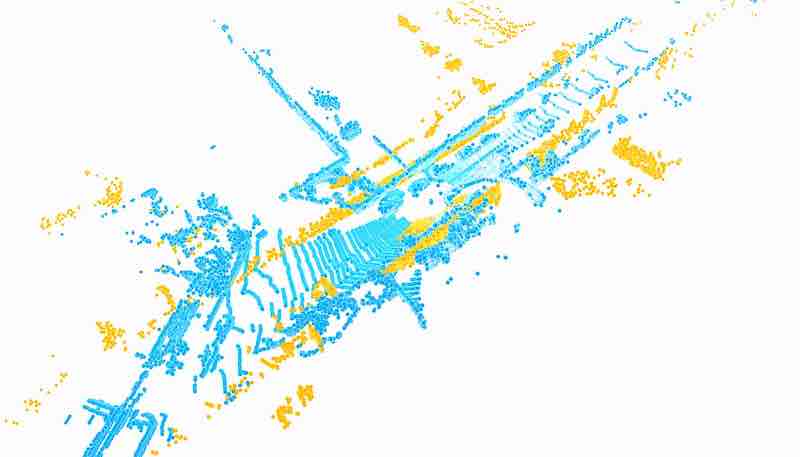}
        \end{overpic} &
        \begin{overpic}[width=0.32\textwidth]{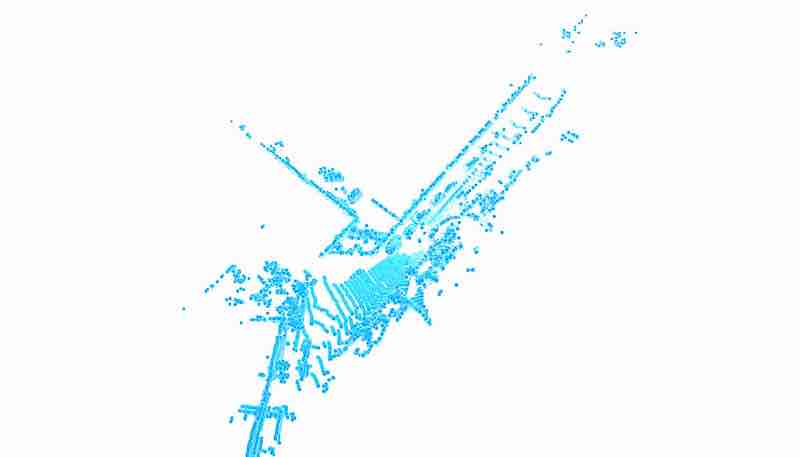}
        \end{overpic}\\
        \midrule
        \begin{overpic}[width=0.32\textwidth]{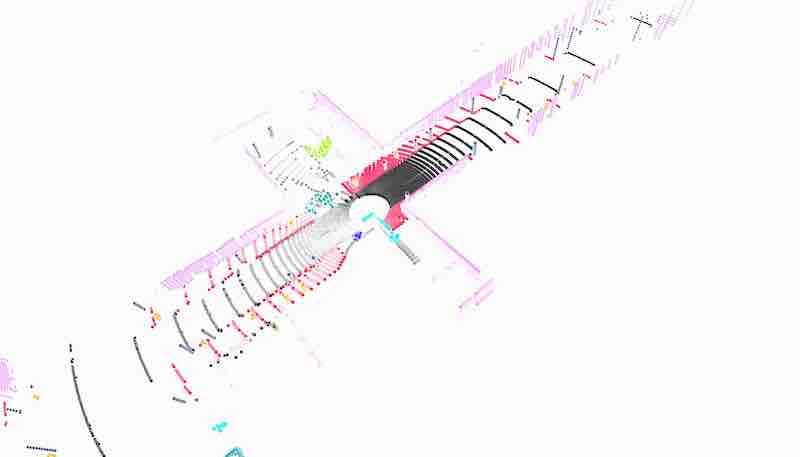}
        \end{overpic} &  
        \begin{overpic}[width=0.32\textwidth]{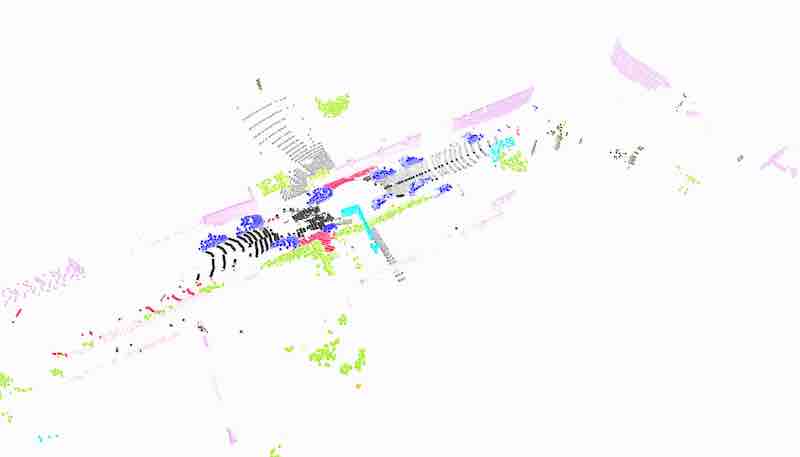}
        \end{overpic} &
        \begin{overpic}[width=0.32\textwidth]{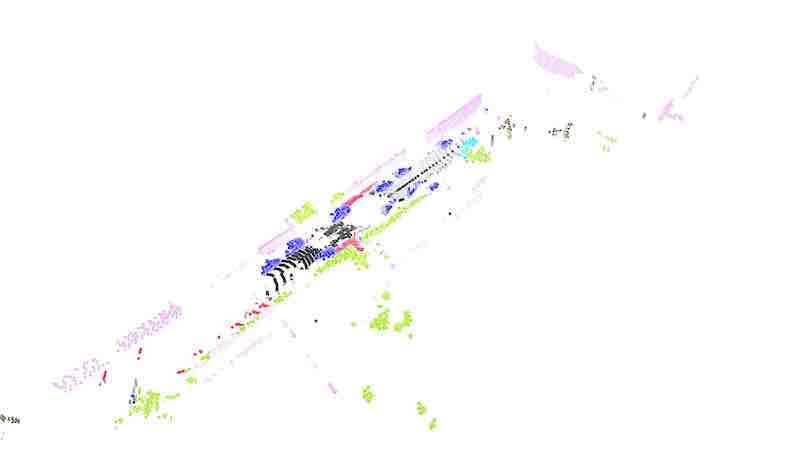}
        \end{overpic}\\
        \begin{overpic}[width=0.32\textwidth]{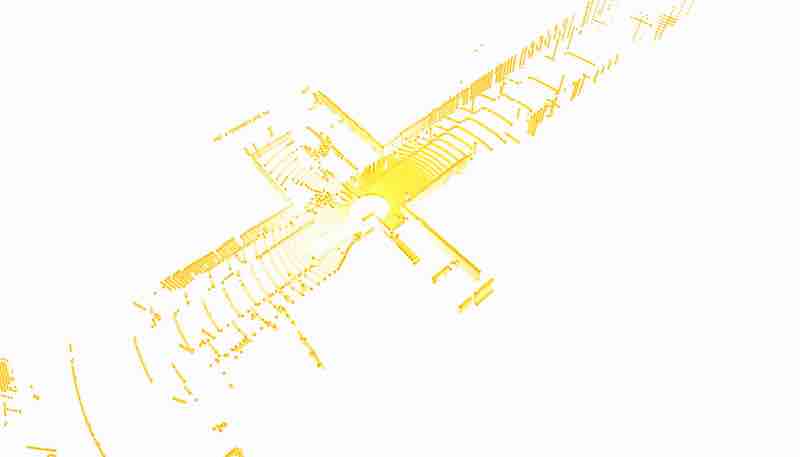}
        \end{overpic} &  
        \begin{overpic}[width=0.32\textwidth]{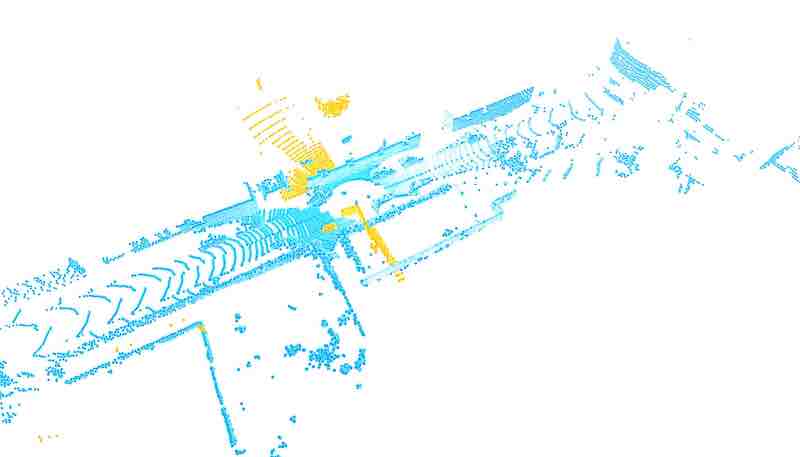}
        \end{overpic} &
        \begin{overpic}[width=0.32\textwidth]{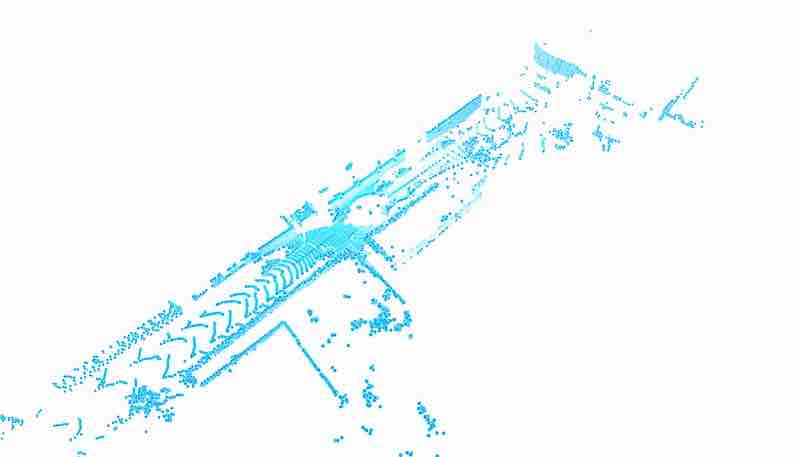}
        \end{overpic}\\
        
        % \begin{overpic}[width=0.32\textwidth]{images/supplementary/mix/source_3.jpg}
        % \end{overpic} &  
        % \begin{overpic}[width=0.32\textwidth]{images/supplementary/mix/s2t_3.jpg}
        % \end{overpic} &
        % \begin{overpic}[width=0.32\textwidth]{images/supplementary/mix/target_3.jpg}
        % \end{overpic}\\
        % \begin{overpic}[width=0.32\textwidth]{images/supplementary/mix/source_mask_3.jpg}
        % \end{overpic} &  
        % \begin{overpic}[width=0.32\textwidth]{images/supplementary/mix/s2t_mask_3.jpg}
        % \end{overpic} &
        % \begin{overpic}[width=0.32\textwidth]{images/supplementary/mix/target_mask_3.jpg}
        % \end{overpic}\\
    \end{tabular}
    \caption{Example of mixed point clouds in the s$\rightarrow$t branch on SynLiDAR $\rightarrow$ SemanticKITTI. We report scenes with annotations (top rows) and binary masks (bottom rows).}
    \label{fig:qualitative_mix_s2t}
\end{figure}

\begin{figure}[t]
\centering
    \setlength\tabcolsep{1.pt}
    \begin{tabular}{ccc}
    \raggedright
        \begin{overpic}[width=0.32\textwidth]{images/supplementary/mix/source_5.jpg}
        \put(40,67){\color{black}\footnotesize \textbf{source}}
        \put(160,67){\color{black}\footnotesize \textbf{$t \rightarrow s$}}
        \put(285,68){\color{black}\footnotesize \textbf{target}}
        \end{overpic} &  
        \begin{overpic}[width=0.32\textwidth]{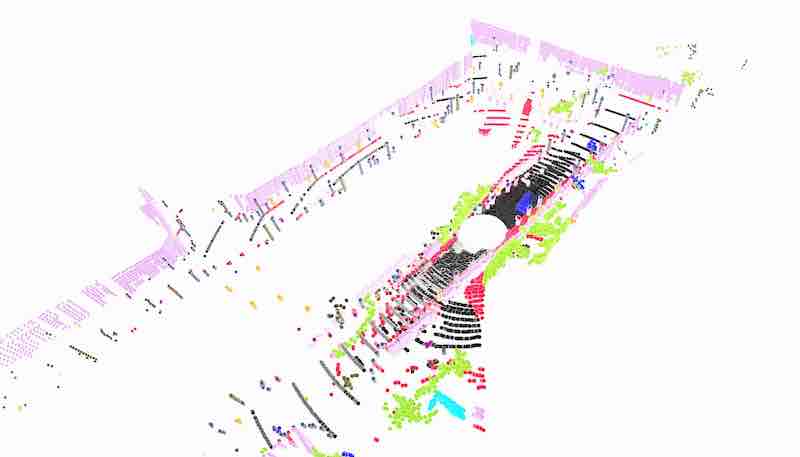}
        \end{overpic} &
        \begin{overpic}[width=0.32\textwidth]{images/supplementary/mix/target_5.jpg}
        \end{overpic}\\
        \begin{overpic}[width=0.32\textwidth]{images/supplementary/mix/source_mask_5.jpg}
        \end{overpic} &  
        \begin{overpic}[width=0.32\textwidth]{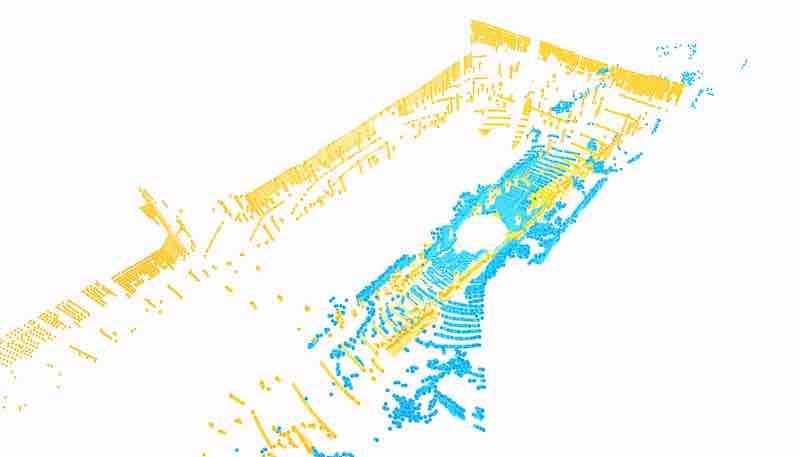}
        \end{overpic} &
        \begin{overpic}[width=0.32\textwidth]{images/supplementary/mix/target_mask_5.jpg}
        \end{overpic}\\
        \midrule
        \begin{overpic}[width=0.32\textwidth]{images/supplementary/mix/source_60.jpg}
        \end{overpic} &  
        \begin{overpic}[width=0.32\textwidth]{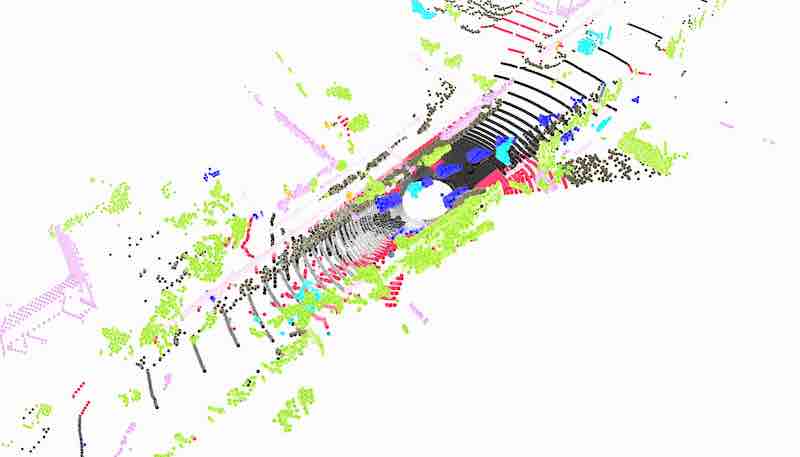}
        \end{overpic} &
        \begin{overpic}[width=0.32\textwidth]{images/supplementary/mix/target_60.jpg}
        \end{overpic}\\
        \begin{overpic}[width=0.32\textwidth]{images/supplementary/mix/source_mask_60.jpg}
        \end{overpic} &  
        \begin{overpic}[width=0.32\textwidth]{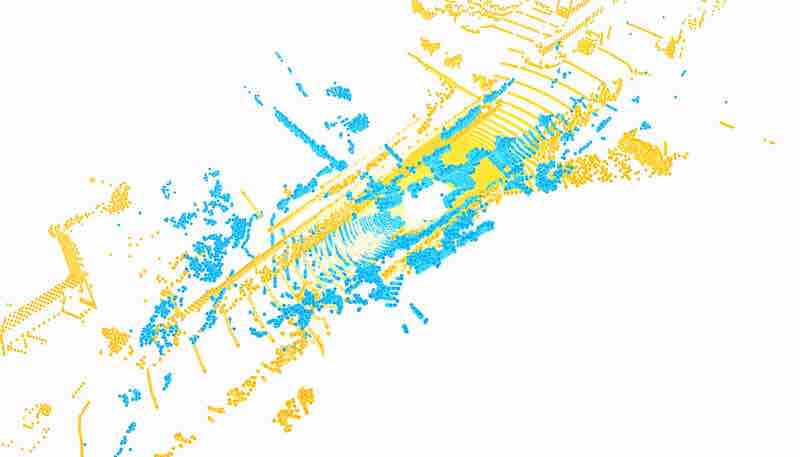}
        \end{overpic} &
        \begin{overpic}[width=0.32\textwidth]{images/supplementary/mix/target_mask_60.jpg}
        \end{overpic}\\
        \midrule
        \begin{overpic}[width=0.32\textwidth]{images/supplementary/mix/source_130.jpg}
        \end{overpic} &  
        \begin{overpic}[width=0.32\textwidth]{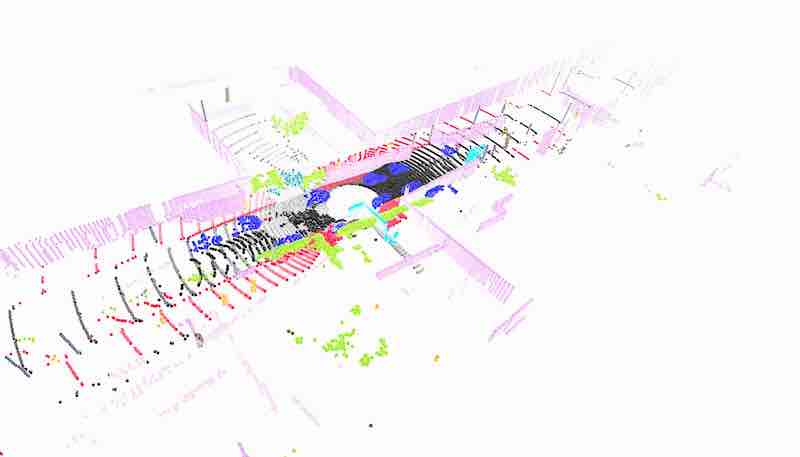}
        \end{overpic} &
        \begin{overpic}[width=0.32\textwidth]{images/supplementary/mix/target_130.jpg}
        \end{overpic}\\
        \begin{overpic}[width=0.32\textwidth]{images/supplementary/mix/source_mask_130.jpg}
        \end{overpic} &  
        \begin{overpic}[width=0.32\textwidth]{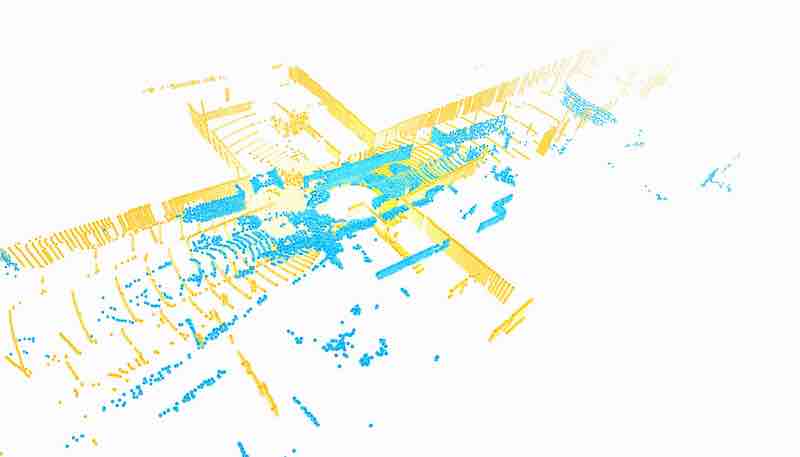}
        \end{overpic} &
        \begin{overpic}[width=0.32\textwidth]{images/supplementary/mix/target_mask_130.jpg}
        \end{overpic}\\
        % \begin{overpic}[width=0.32\textwidth]{images/supplementary/mix/source_3.jpg}
        % \end{overpic} &  
        % \begin{overpic}[width=0.32\textwidth]{images/supplementary/mix/t2s_3.jpg}
        % \end{overpic} &
        % \begin{overpic}[width=0.32\textwidth]{images/supplementary/mix/target_3.jpg}
        % \end{overpic}\\
        % \begin{overpic}[width=0.32\textwidth]{images/supplementary/mix/source_mask_3.jpg}
        % \end{overpic} &  
        % \begin{overpic}[width=0.32\textwidth]{images/supplementary/mix/t2s_mask_3.jpg}
        % \end{overpic} &
        % \begin{overpic}[width=0.32\textwidth]{images/supplementary/mix/target_mask_3.jpg}
        % \end{overpic}\\
    \end{tabular}
    \caption{Example of mixed point clouds in the t$\rightarrow$s branch on SynLiDAR $\rightarrow$ SemanticKITTI.  We report scenes with annotations (top rows) and binary masks (bottom rows).}
    \label{fig:qualitative_mix_t2s}
\end{figure}

We provide qualitative examples of the mixed input point clouds $\mathcal{X}^{s\rightarrow t}$ and $\mathcal{X}^{t\rightarrow s}$ on SynLiDAR$\rightarrow$SemanticKITTI. Each point cloud is randomly sampled during adaptation with local $h$ and global $r$ augmentations activated, $\zeta=0.9$ and $\alpha=0.5$. In Fig.~\ref{fig:qualitative_mix_s2t}, we report $\mathcal{X}^{s\rightarrow t}$ taken from the $s \rightarrow t$ branch while in Fig.~\ref{fig:qualitative_mix_t2s} we report $\mathcal{X}^{t\rightarrow s}$ taken from the $t \rightarrow s$ branch. We provide paired source, target and mixed ($s \rightarrow t$ and $t \rightarrow s$) point clouds by reporting labels (top row) and binary masks (bottom row). 
Source labels are the ground-truth labels while target labels are the pseudo-labels filtered with $\zeta=0.9$.
Mixed samples show hybrid scenes with both source and target components. Although classes with points distributed over wide regions may be mixed (\textit{e.g.} road and especially in the case of $t \rightarrow s$), mixed point clouds often include complementary elements among domains.

\section{Qualitative adaptation results}\label{sec:supplementary_qualitative}
\begin{figure}[t]
\centering
    \setlength\tabcolsep{1.pt}
    \begin{tabular}{ccc}
    \raggedright
        
        \begin{overpic}[width=0.32\textwidth]{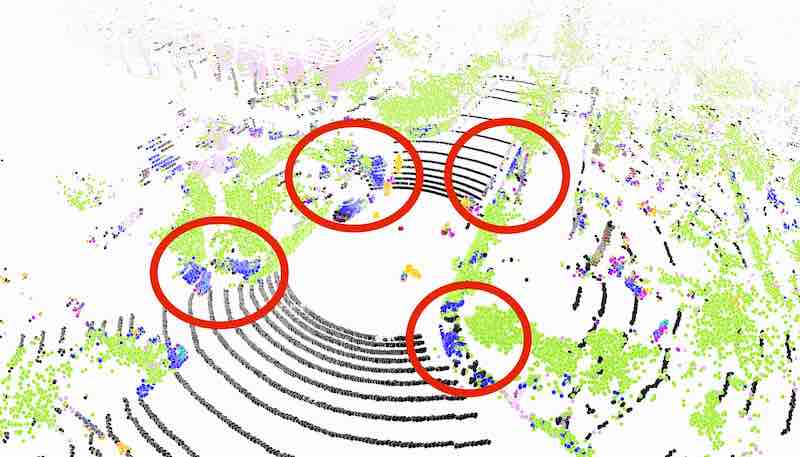}
        \put(40,67){\color{black}\footnotesize \textbf{source}}
        \put(160,67){\color{black}\footnotesize \textbf{ours}}
        \put(285,68){\color{black}\footnotesize \textbf{gt}}
        \end{overpic} &  
        \begin{overpic}[width=0.32\textwidth]{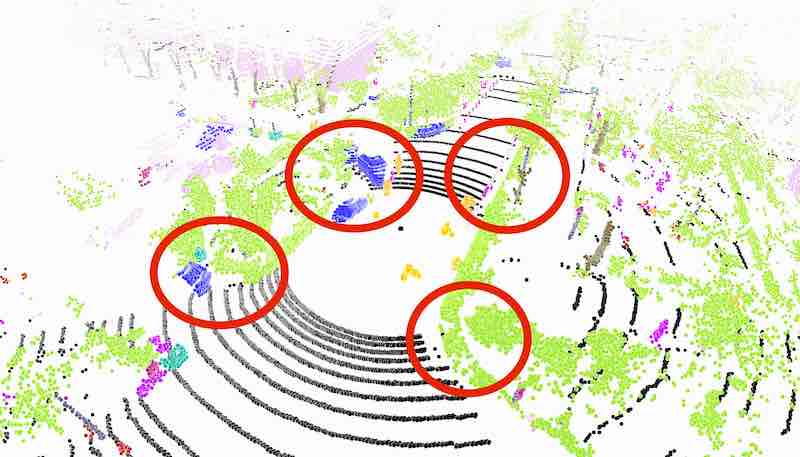}
        \end{overpic} &
        \begin{overpic}[width=0.32\textwidth]{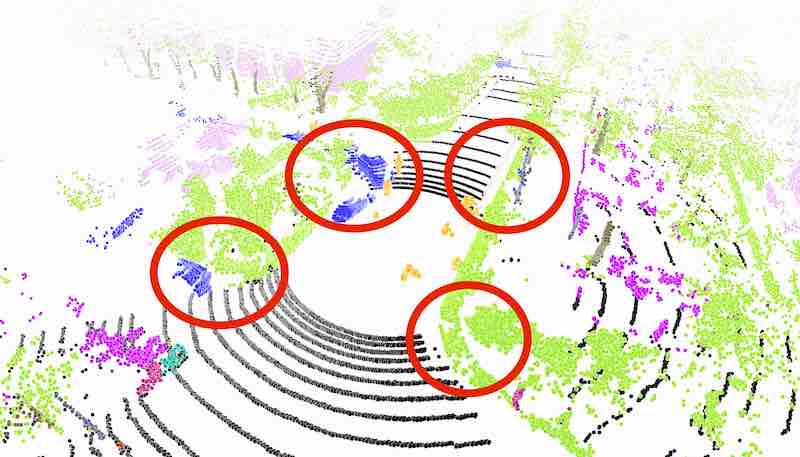}
        \end{overpic}\\
        \begin{overpic}[width=0.32\textwidth]{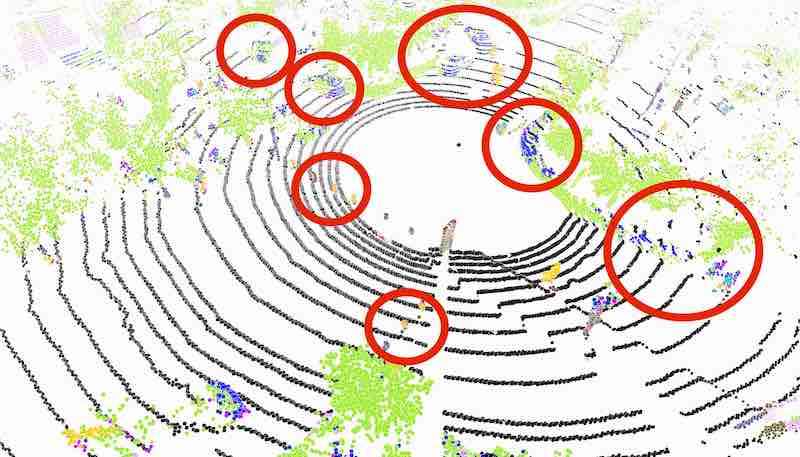}
        \end{overpic} &  
        \begin{overpic}[width=0.32\textwidth]{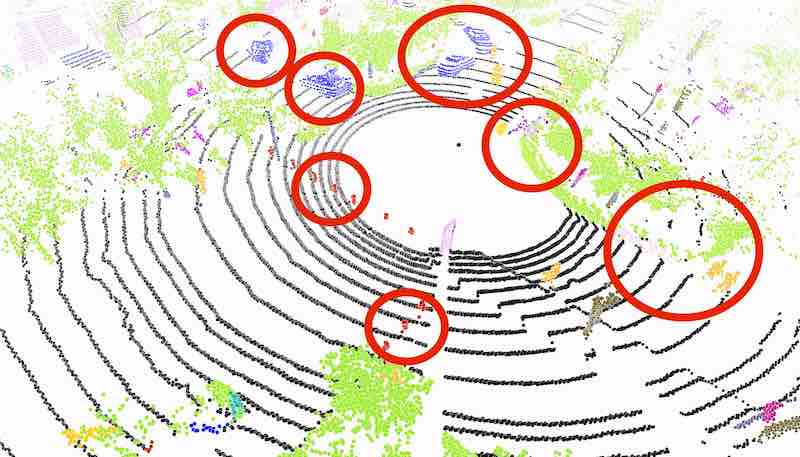}
        \end{overpic} &
        \begin{overpic}[width=0.32\textwidth]{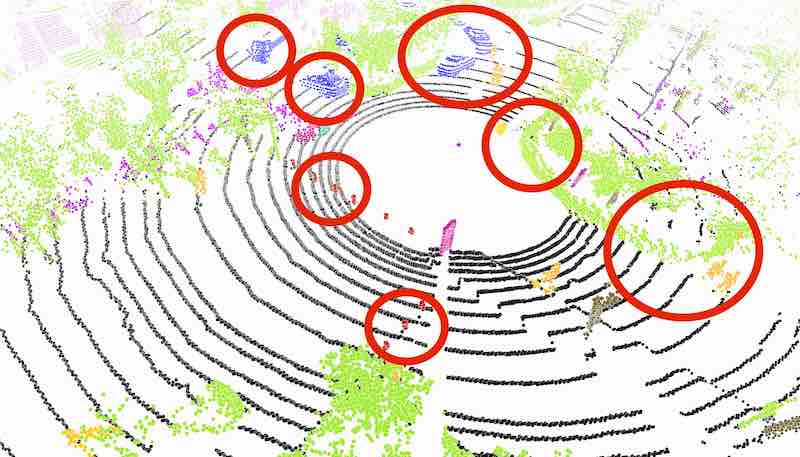}
        \end{overpic}\\
        \begin{overpic}[width=0.32\textwidth]{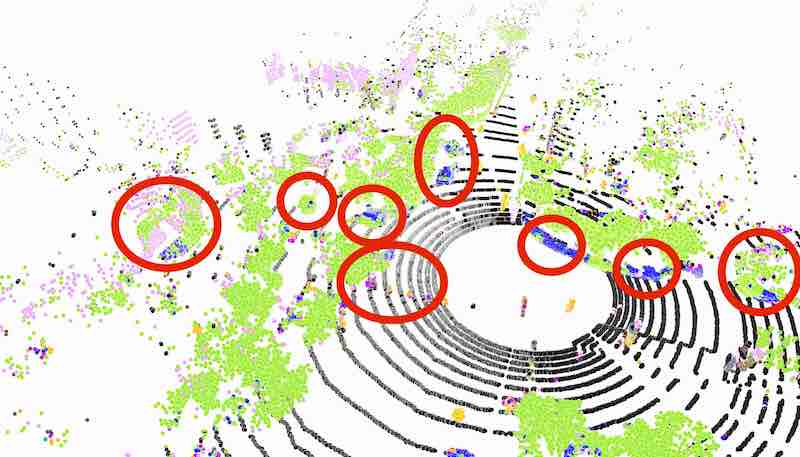}
        \end{overpic} &  
        \begin{overpic}[width=0.32\textwidth]{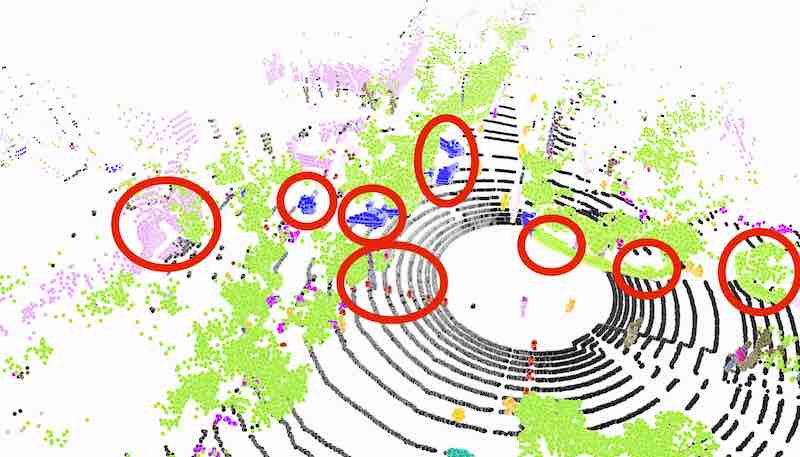}
        \end{overpic} &
        \begin{overpic}[width=0.32\textwidth]{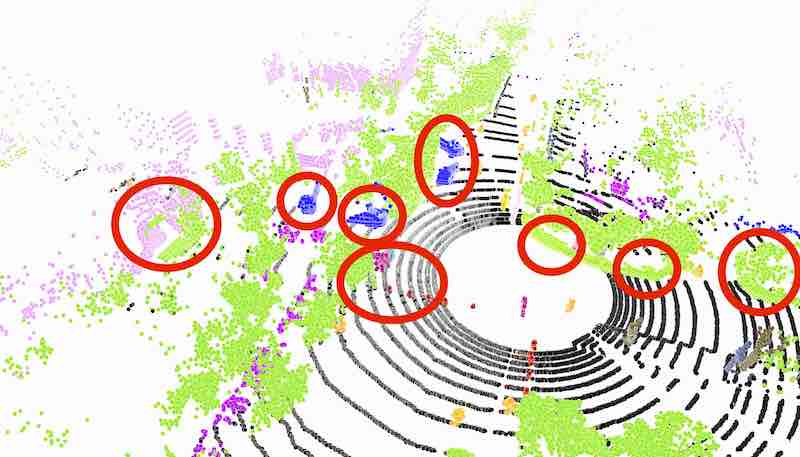}
        \end{overpic}\\
        \begin{overpic}[width=0.32\textwidth]{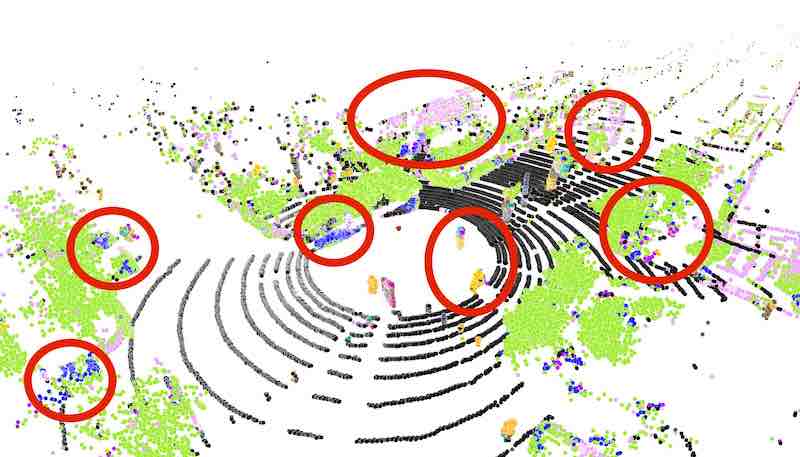}
        \end{overpic} &  
        \begin{overpic}[width=0.32\textwidth]{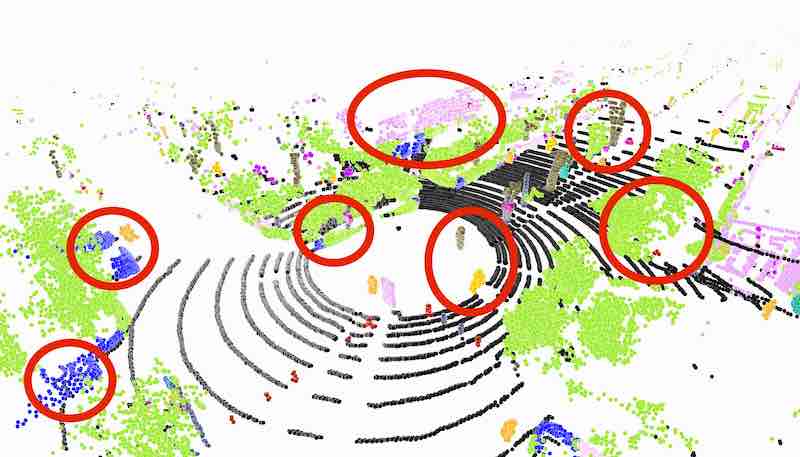}
        \end{overpic} &
        \begin{overpic}[width=0.32\textwidth]{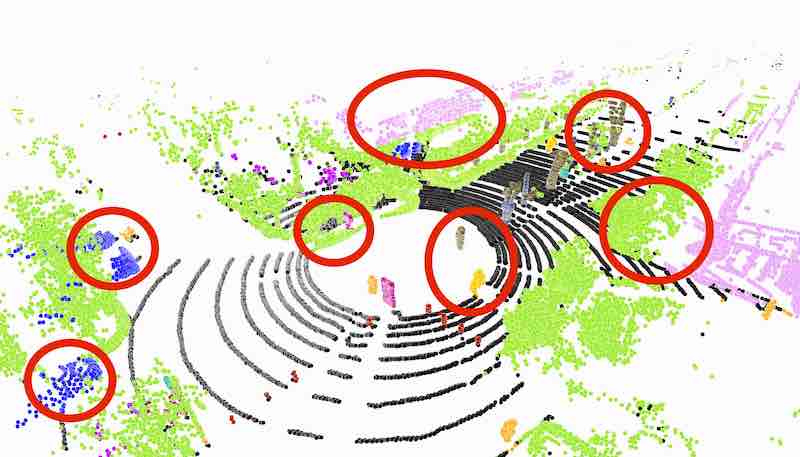}
        \end{overpic}\\
        \begin{overpic}[width=0.32\textwidth]{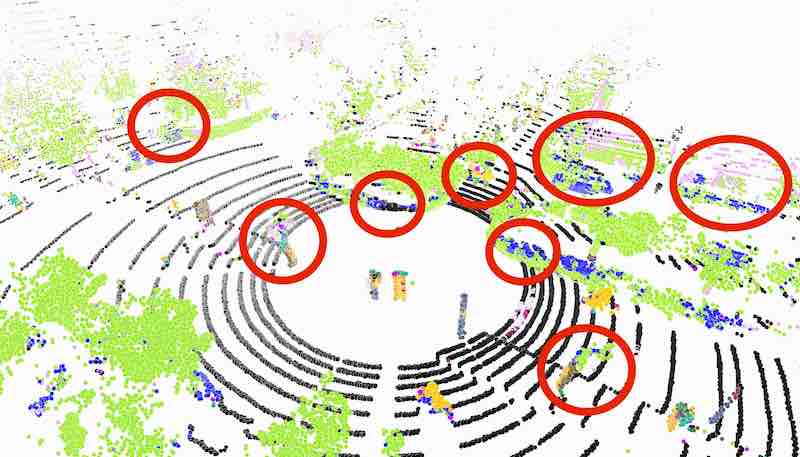}
        \end{overpic} &  
        \begin{overpic}[width=0.32\textwidth]{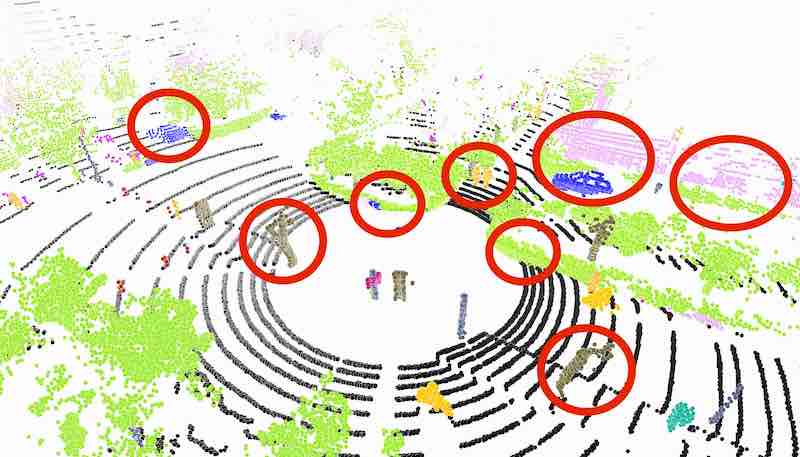}
        \end{overpic} &
        \begin{overpic}[width=0.32\textwidth]{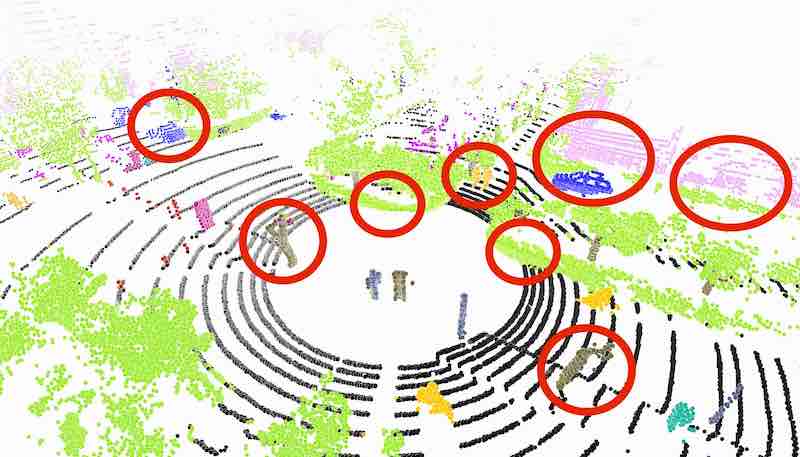}
        \end{overpic}\\
        \multicolumn{3}{c}{
        \begin{overpic}[width=0.99\textwidth]{images/qualitative/legend_semposs.pdf}
        \end{overpic}}
    \end{tabular}
    \caption{Results on SynLiDAR$\rightarrow$SemanticPOSS. Source predictions are often wrong and mingled in the same region. After adaptation, \ourmethod improves the segmentation accuracy with homogeneous predictions and correctly assigned classes. The red circles highlight regions with interesting results.}
    \label{fig:qualitative_poss_sup_0}
\end{figure}

\begin{figure}[t]
\centering
    \setlength\tabcolsep{1.pt}
    \begin{tabular}{ccc}
    \raggedright
        \begin{overpic}[width=0.32\textwidth]{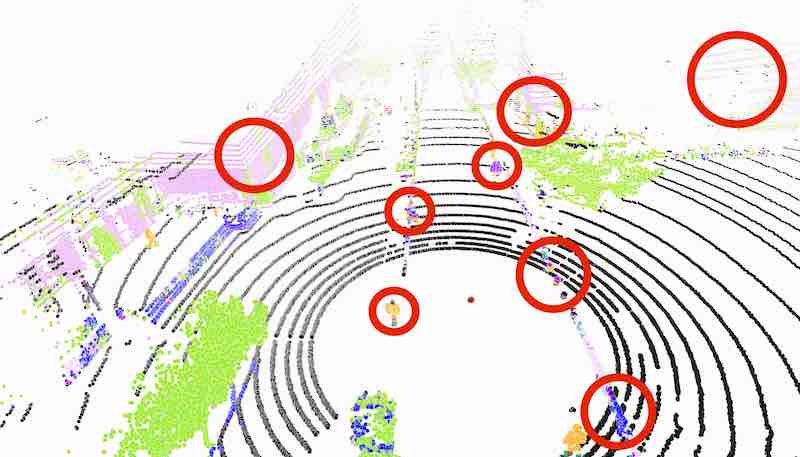}
        \put(40,67){\color{black}\footnotesize \textbf{source}}
        \put(160,67){\color{black}\footnotesize \textbf{ours}}
        \put(285,68){\color{black}\footnotesize \textbf{gt}}
        \end{overpic} &  
        \begin{overpic}[width=0.32\textwidth]{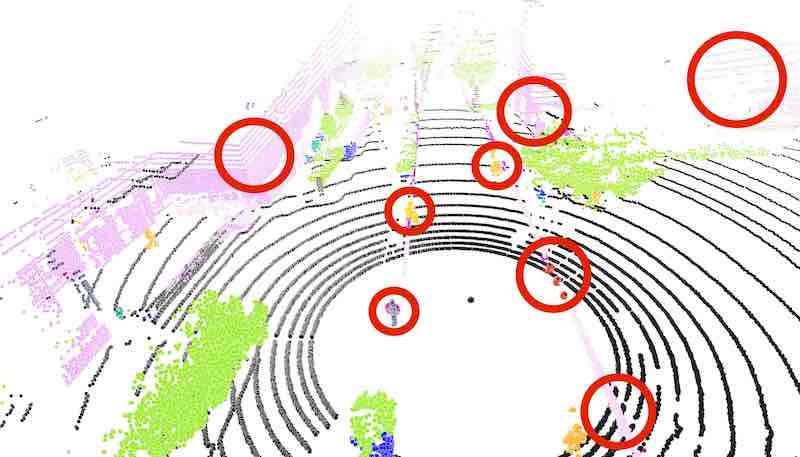}
        \end{overpic} &
        \begin{overpic}[width=0.32\textwidth]{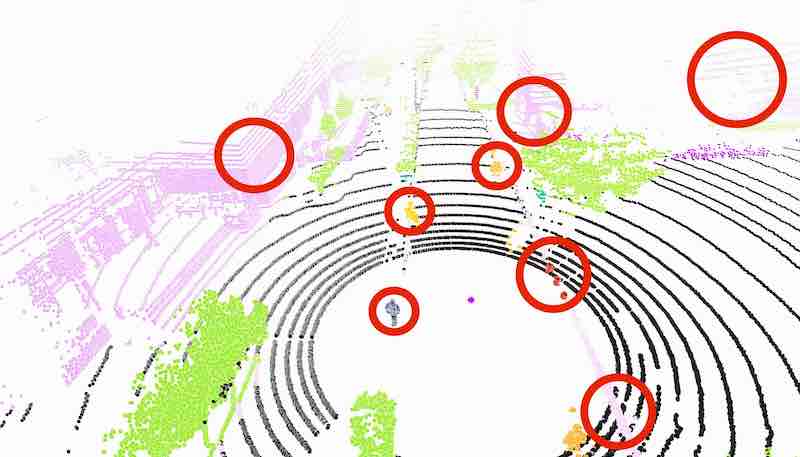}
        \end{overpic}\\
        \begin{overpic}[width=0.32\textwidth]{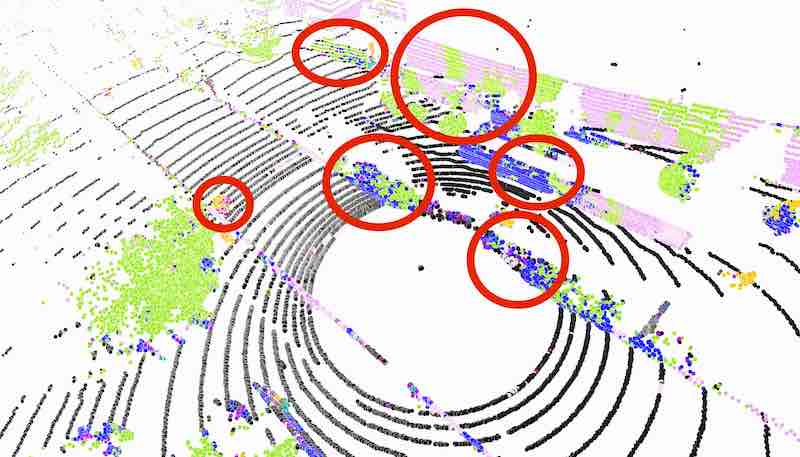}
        \end{overpic} &  
        \begin{overpic}[width=0.32\textwidth]{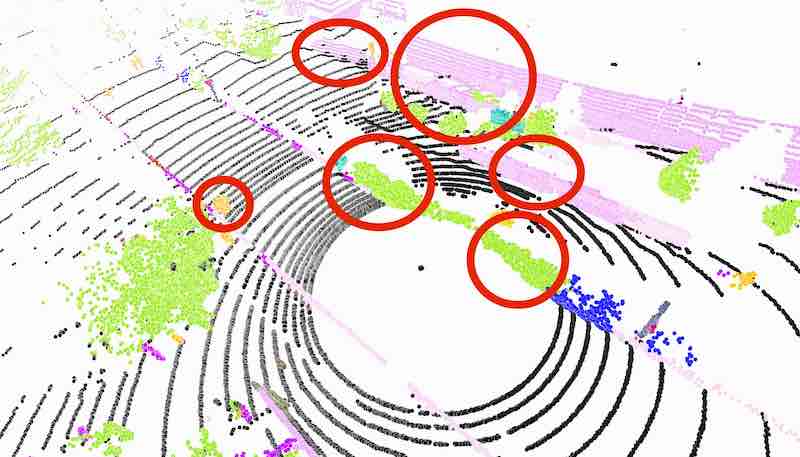}
        \end{overpic} &
        \begin{overpic}[width=0.32\textwidth]{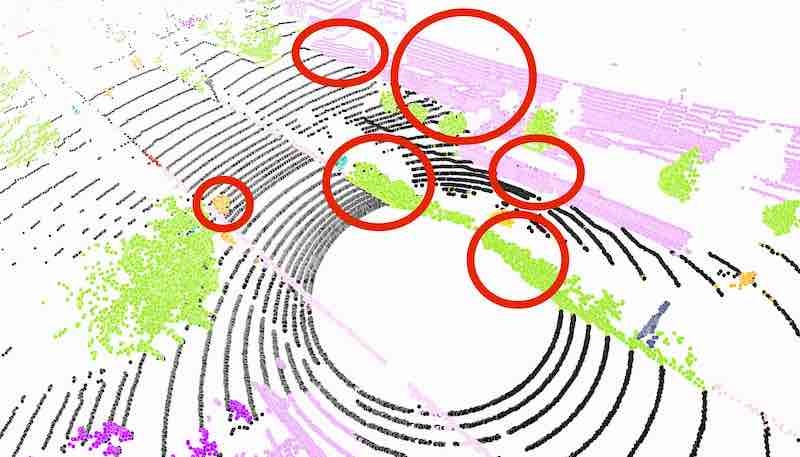}
        \end{overpic}\\
        \begin{overpic}[width=0.32\textwidth]{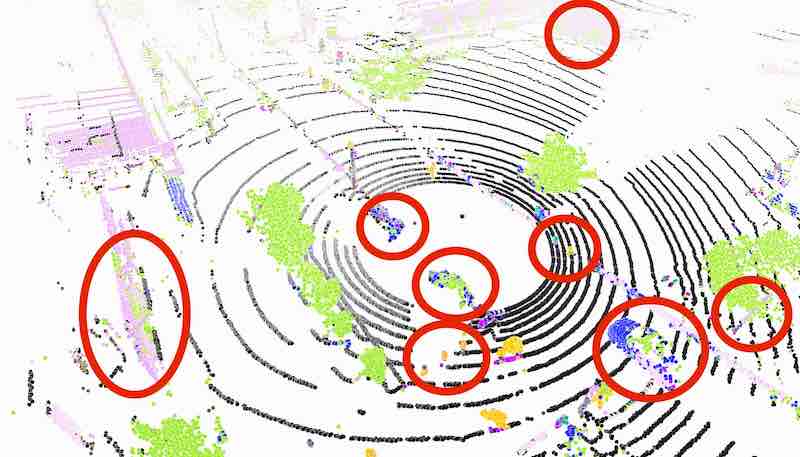}
        \end{overpic} &  
        \begin{overpic}[width=0.32\textwidth]{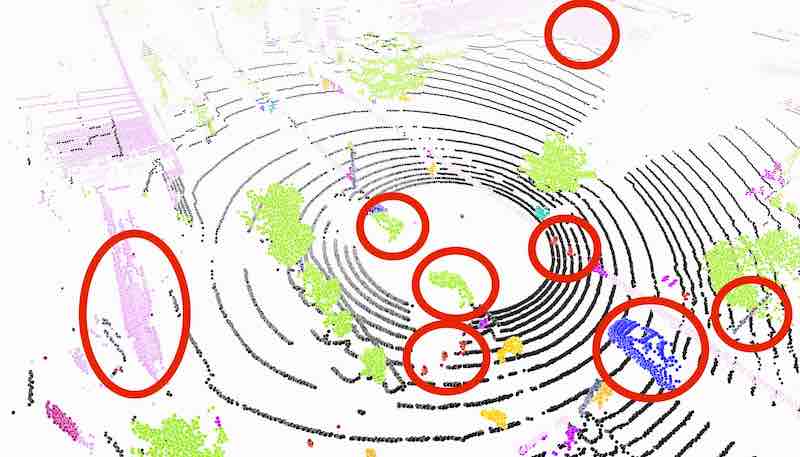}
        \end{overpic} &
        \begin{overpic}[width=0.32\textwidth]{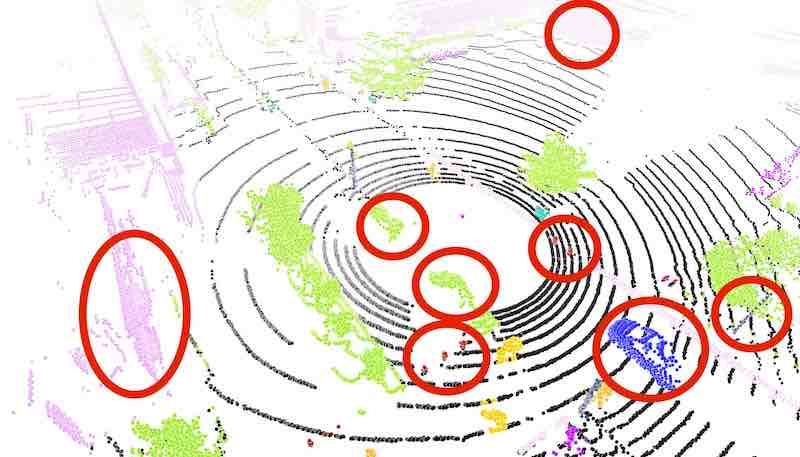}
        \end{overpic}\\
        \begin{overpic}[width=0.32\textwidth]{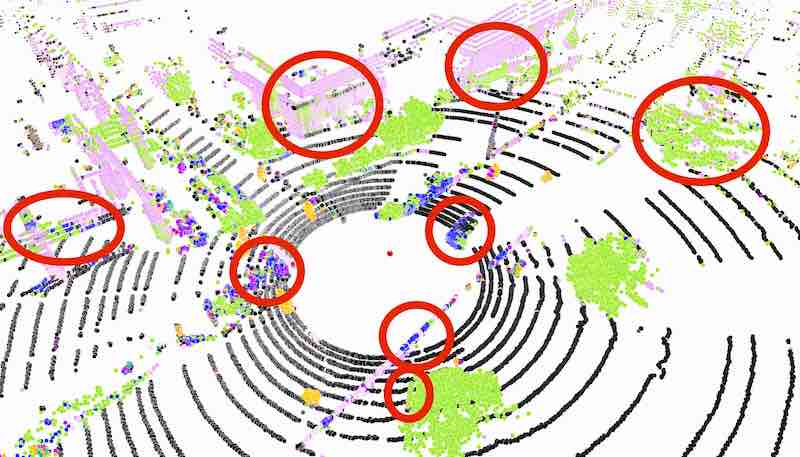}
        \end{overpic} &  
        \begin{overpic}[width=0.32\textwidth]{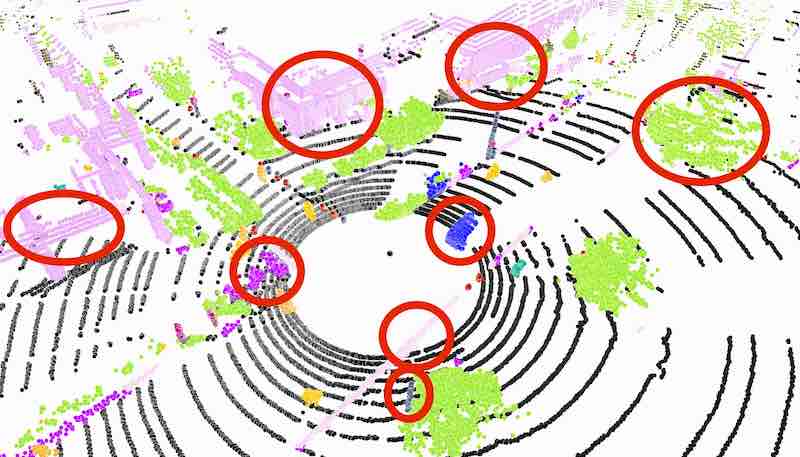}
        \end{overpic} &
        \begin{overpic}[width=0.32\textwidth]{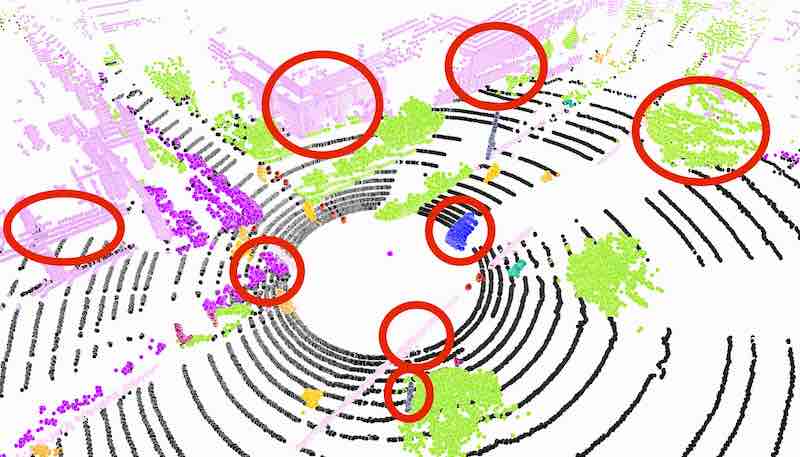}
        \end{overpic}\\
        \begin{overpic}[width=0.32\textwidth]{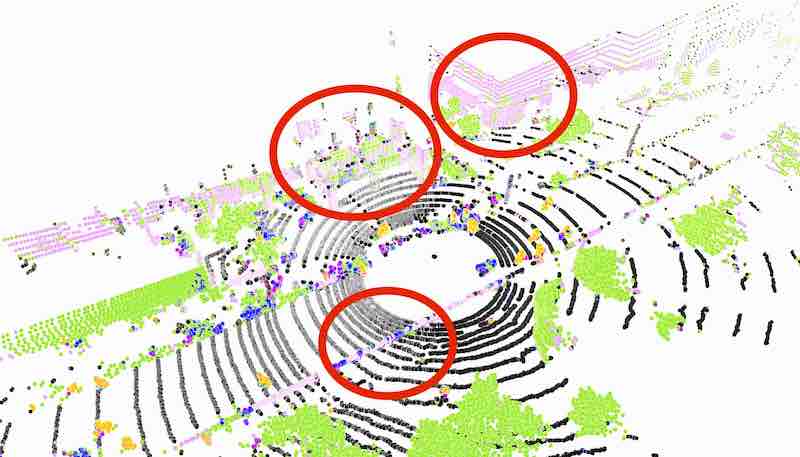}
        \end{overpic} &  
        \begin{overpic}[width=0.32\textwidth]{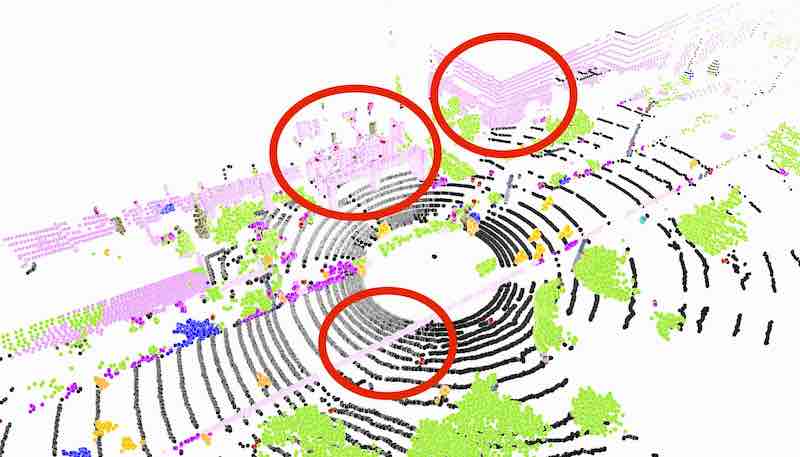}
        \end{overpic} &
        \begin{overpic}[width=0.32\textwidth]{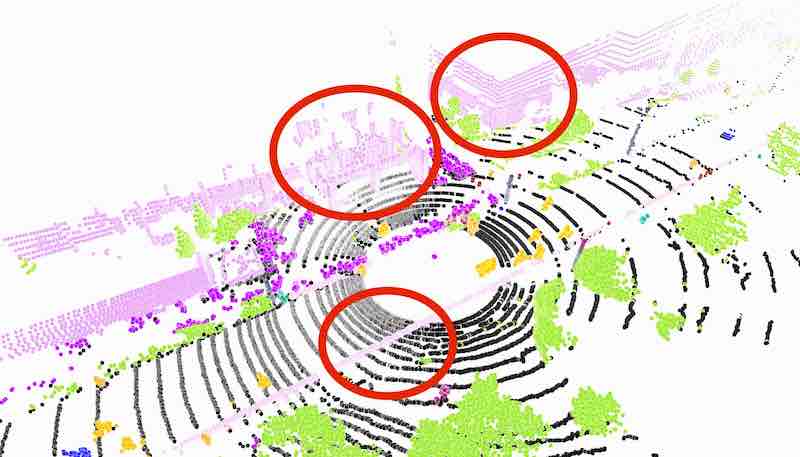}
        \end{overpic}\\
        \multicolumn{3}{c}{
        \begin{overpic}[width=0.99\textwidth]{images/qualitative/legend_semposs.pdf}
        \end{overpic}}
    \end{tabular}
    \caption{Results on SynLiDAR$\rightarrow$SemanticPOSS. Source predictions are often wrong and mingled in the same region. After adaptation, \ourmethod improves the segmentation accuracy with homogeneous predictions and correctly assigned classes. The red circles highlight regions with interesting results.}
    \label{fig:qualitative_poss_sup_1}
\end{figure}

\begin{figure}[t]
\centering
    \setlength\tabcolsep{1.pt}
    \begin{tabular}{ccc}
    \raggedright
        \begin{overpic}[width=0.32\textwidth]{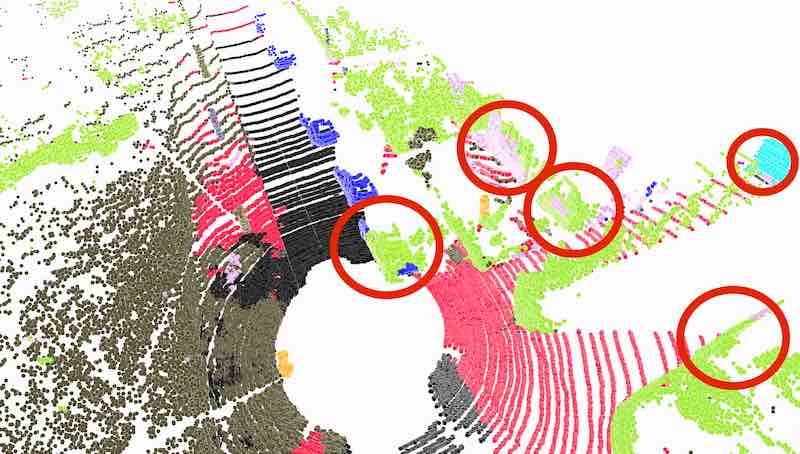}
        \put(40,67){\color{black}\footnotesize \textbf{source}}
        \put(160,67){\color{black}\footnotesize \textbf{ours}}
        \put(285,68){\color{black}\footnotesize \textbf{gt}}
        \end{overpic} &  
        \begin{overpic}[width=0.32\textwidth]{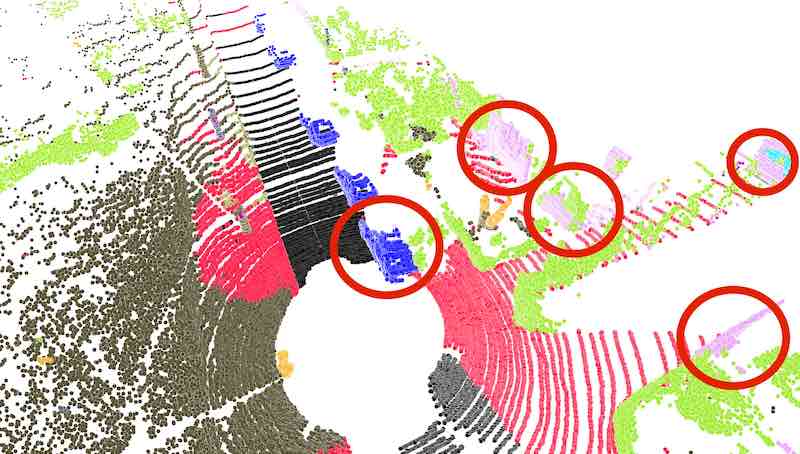}
        \end{overpic} &
        \begin{overpic}[width=0.32\textwidth]{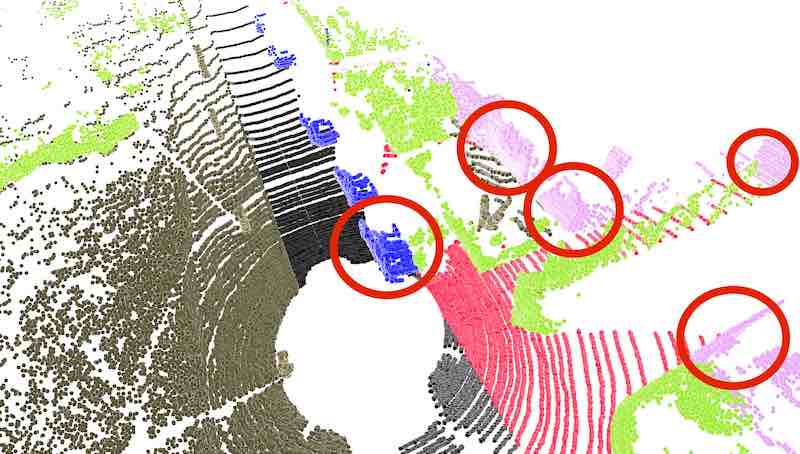}
        \end{overpic}\\
        \begin{overpic}[width=0.32\textwidth]{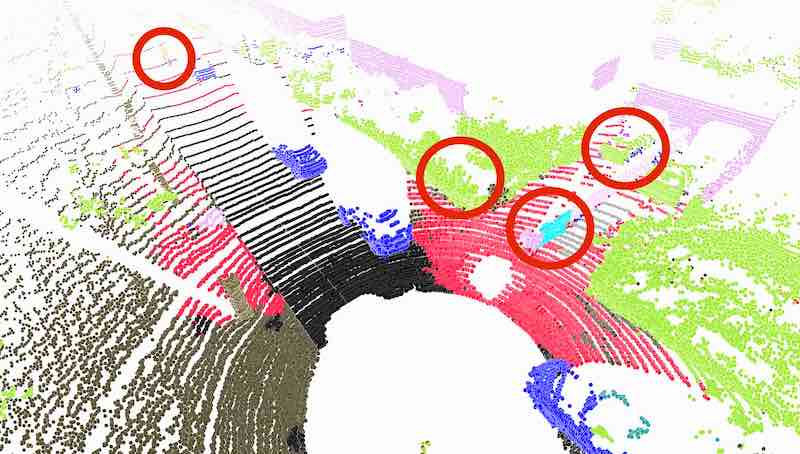}
        \end{overpic} &  
        \begin{overpic}[width=0.32\textwidth]{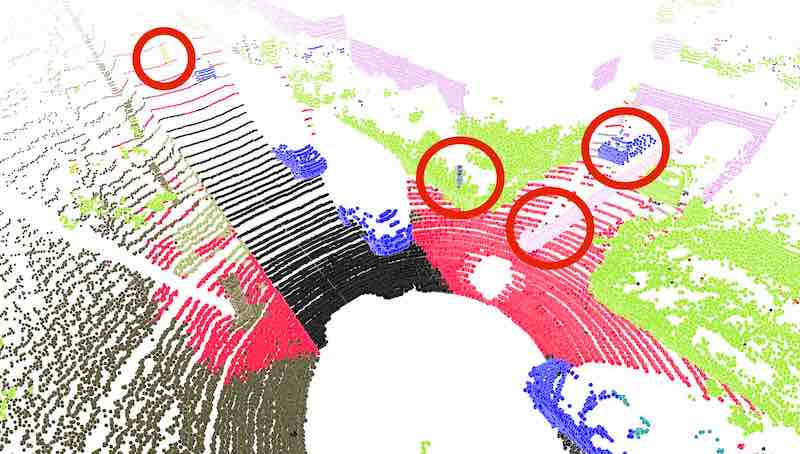}
        \end{overpic} &
        \begin{overpic}[width=0.32\textwidth]{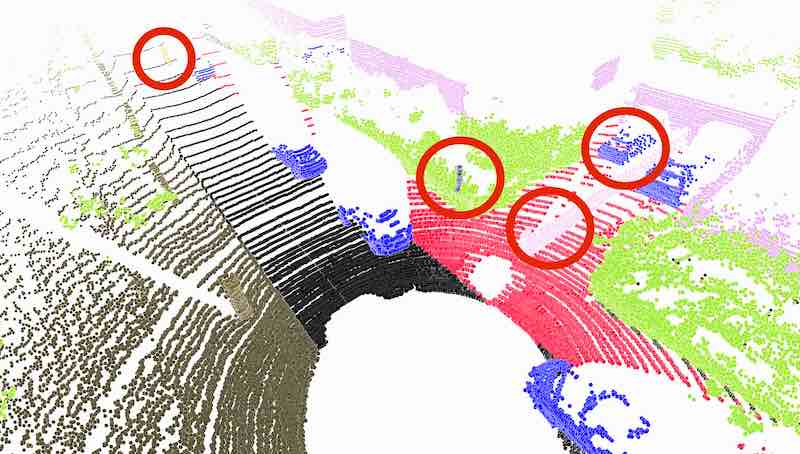}
        \end{overpic}\\
        \begin{overpic}[width=0.32\textwidth]{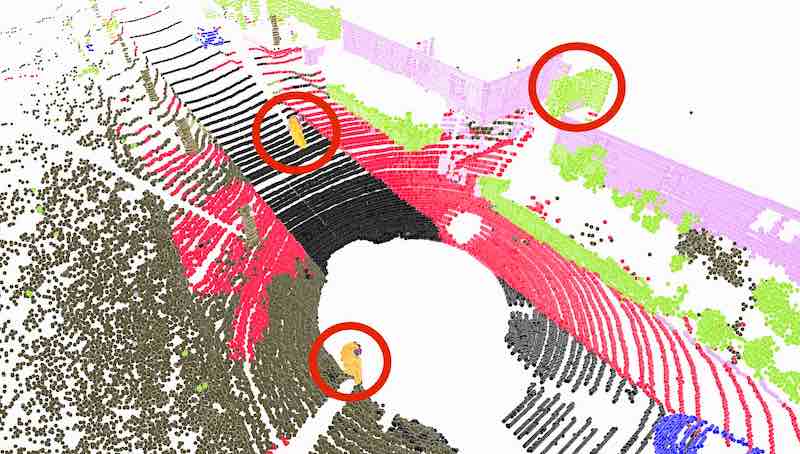}
        \end{overpic} &  
        \begin{overpic}[width=0.32\textwidth]{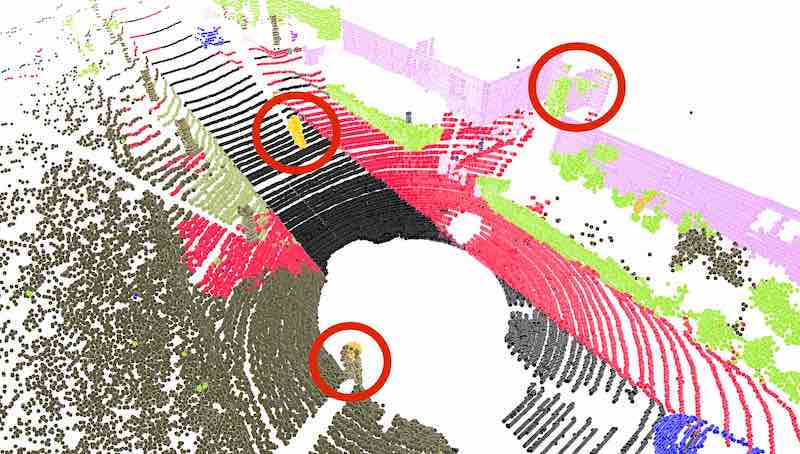}
        \end{overpic} &
        \begin{overpic}[width=0.32\textwidth]{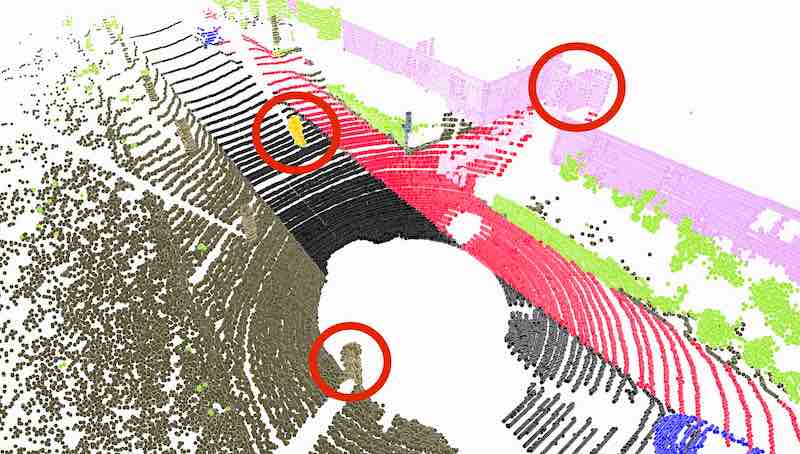}
        \end{overpic}\\
        \begin{overpic}[width=0.32\textwidth]{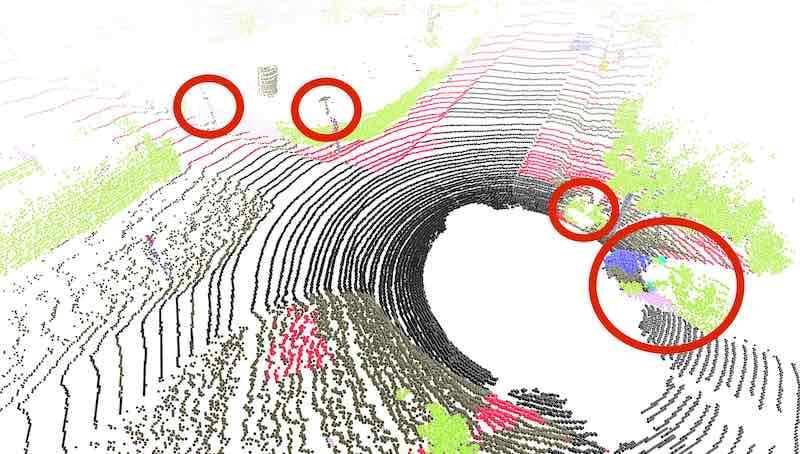}
        \end{overpic} &  
        \begin{overpic}[width=0.32\textwidth]{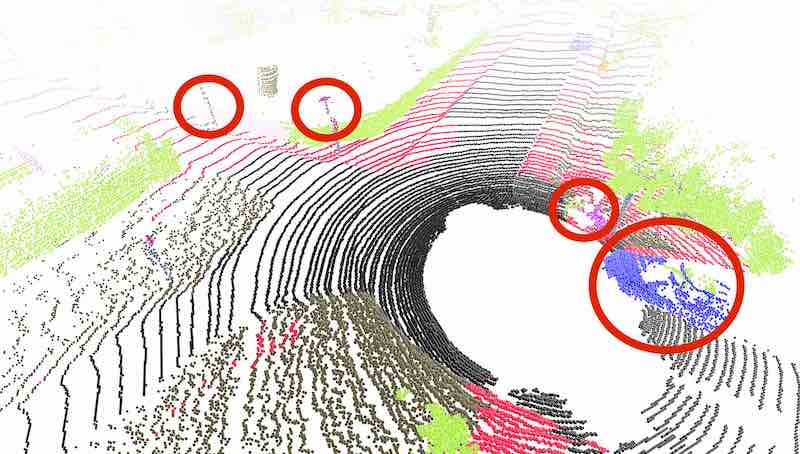}
        \end{overpic} &
        \begin{overpic}[width=0.32\textwidth]{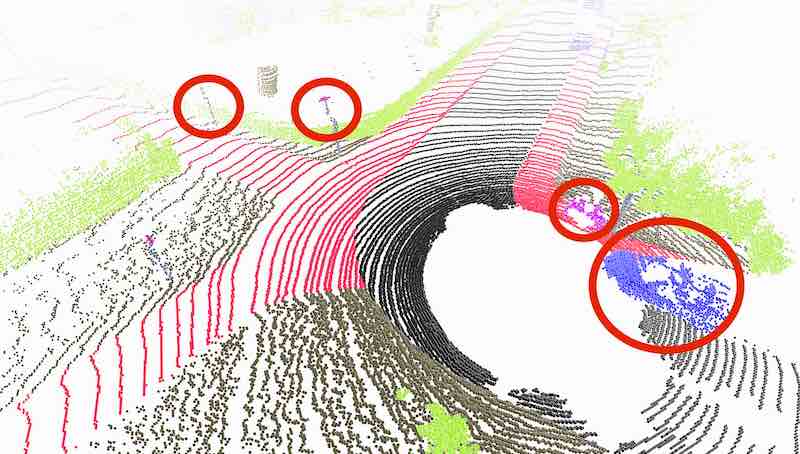}
        \end{overpic}\\
        \begin{overpic}[width=0.32\textwidth]{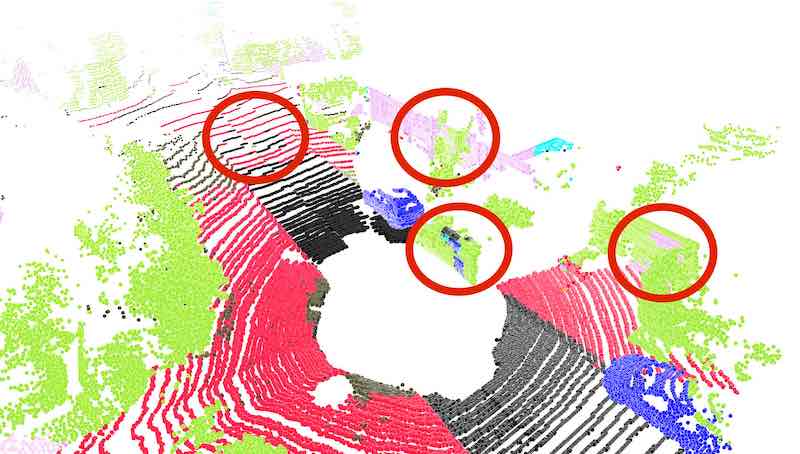}
        \end{overpic} &  
        \begin{overpic}[width=0.32\textwidth]{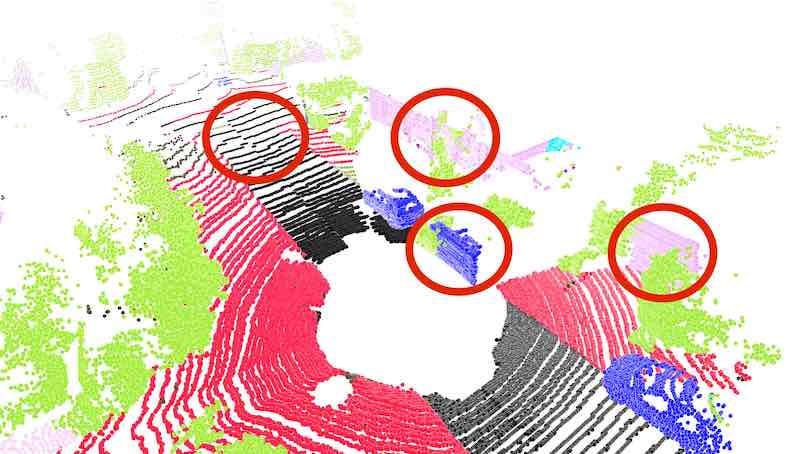}
        \end{overpic} &
        \begin{overpic}[width=0.32\textwidth]{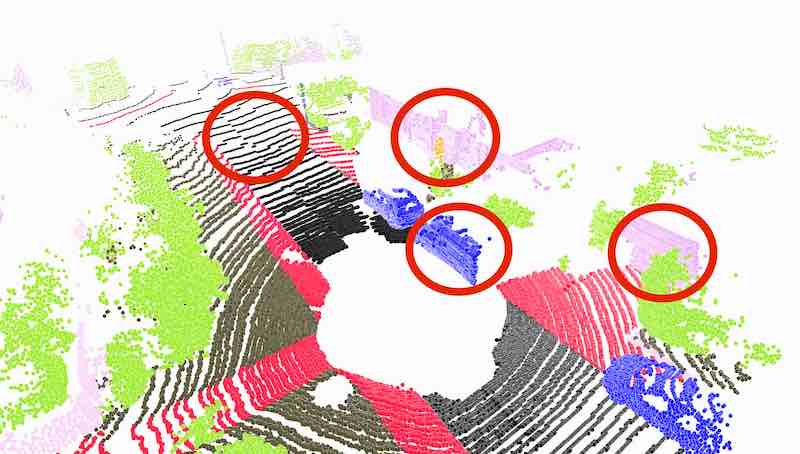}
        \end{overpic}\\
        \multicolumn{3}{c}{
        \begin{overpic}[width=0.99\textwidth]{images/qualitative_2/legend_semkitti.jpg}
        \end{overpic}}
    \end{tabular}

    \caption{Results on SynLiDAR$\rightarrow$SemanticKITTI. Source predictions are often wrong and mingled in the same region. After adaptation, \ourmethod improves the segmentation accuracy with homogeneous predictions and correctly assigned classes. The red circles highlight regions with interesting results.}
    \label{fig:qualitative_kitti_sup_0}
\end{figure}

\begin{figure}[t]
\centering
    \setlength\tabcolsep{1.pt}
    \begin{tabular}{ccc}
    \raggedright
        \begin{overpic}[width=0.32\textwidth]{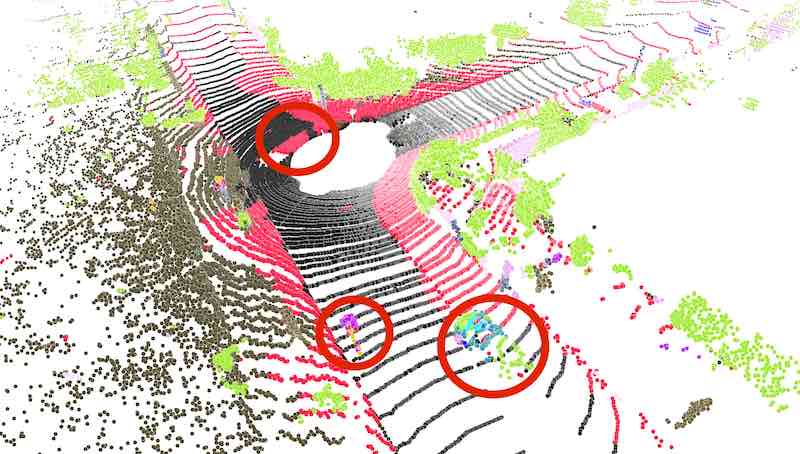}
        \put(40,67){\color{black}\footnotesize \textbf{source}}
        \put(160,67){\color{black}\footnotesize \textbf{ours}}
        \put(285,68){\color{black}\footnotesize \textbf{gt}}
        \end{overpic} &  
        \begin{overpic}[width=0.32\textwidth]{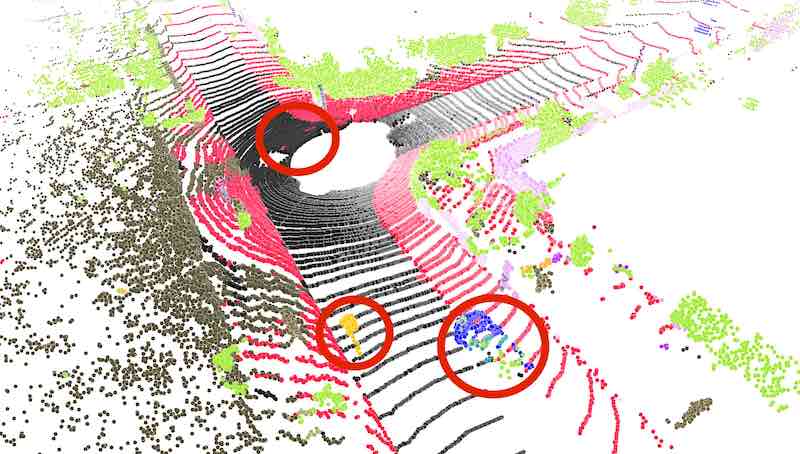}
        \end{overpic} &
        \begin{overpic}[width=0.32\textwidth]{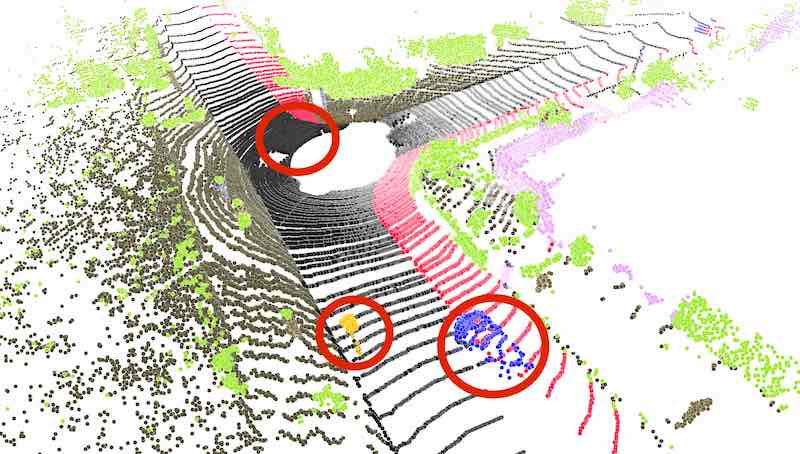}
        \end{overpic}\\
        \begin{overpic}[width=0.32\textwidth]{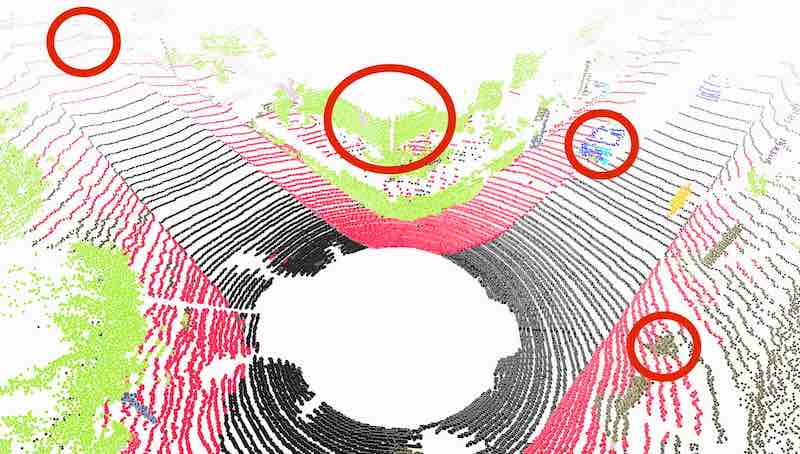}
        \end{overpic} &  
        \begin{overpic}[width=0.32\textwidth]{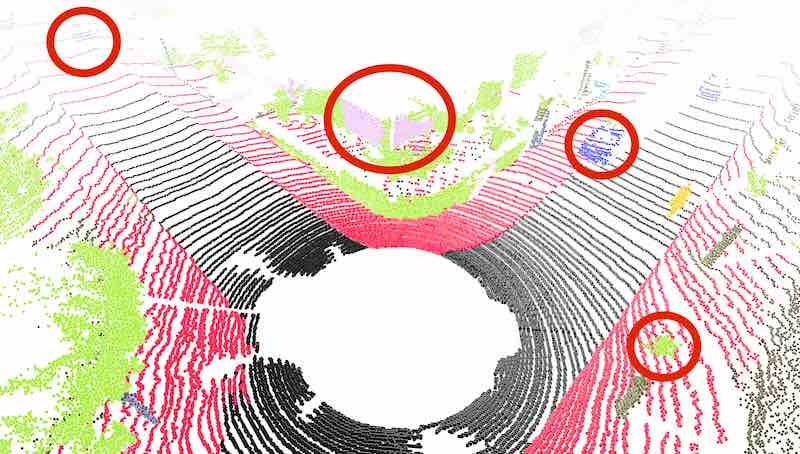}
        \end{overpic} &
        \begin{overpic}[width=0.32\textwidth]{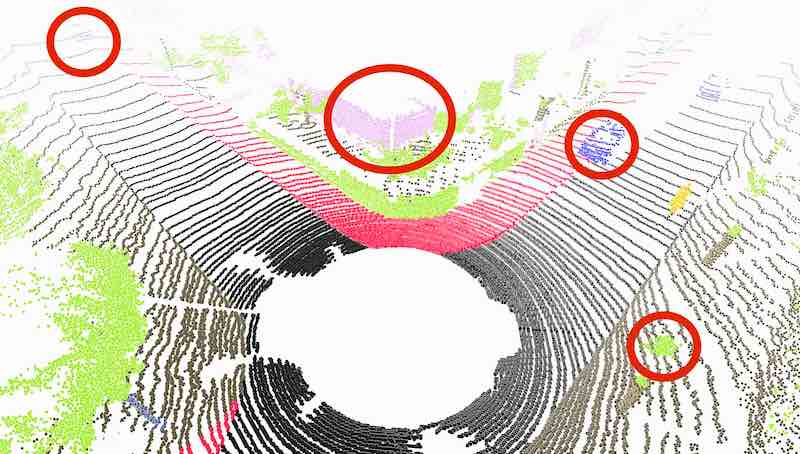}
        \end{overpic}\\
        \begin{overpic}[width=0.32\textwidth]{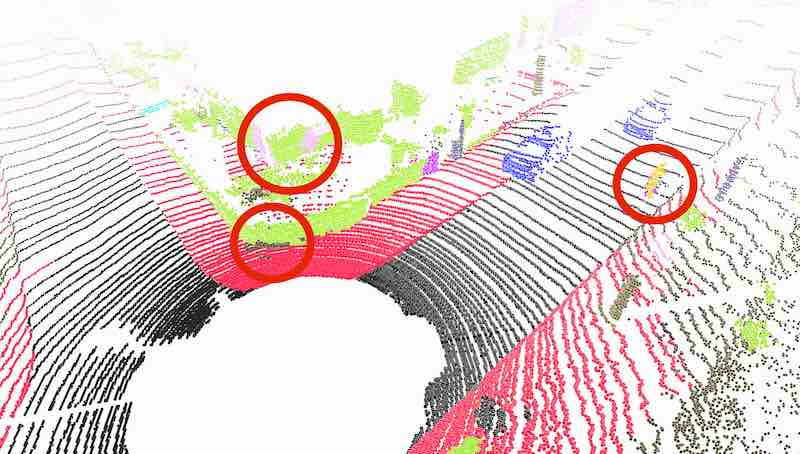}
        \end{overpic} &  
        \begin{overpic}[width=0.32\textwidth]{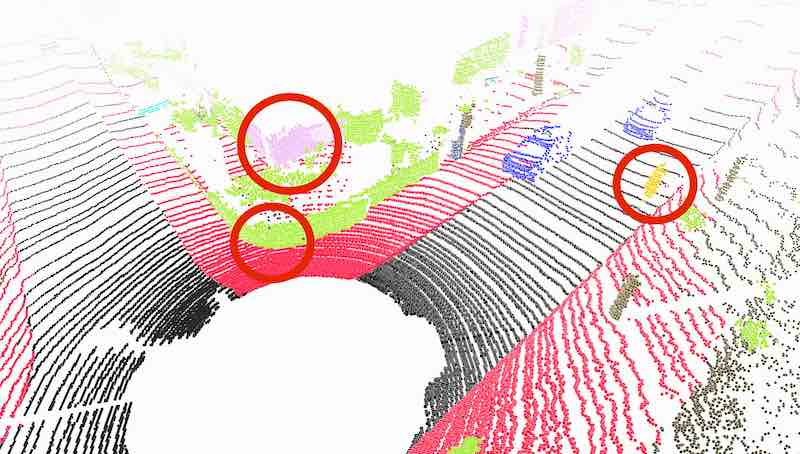}
        \end{overpic} &
        \begin{overpic}[width=0.32\textwidth]{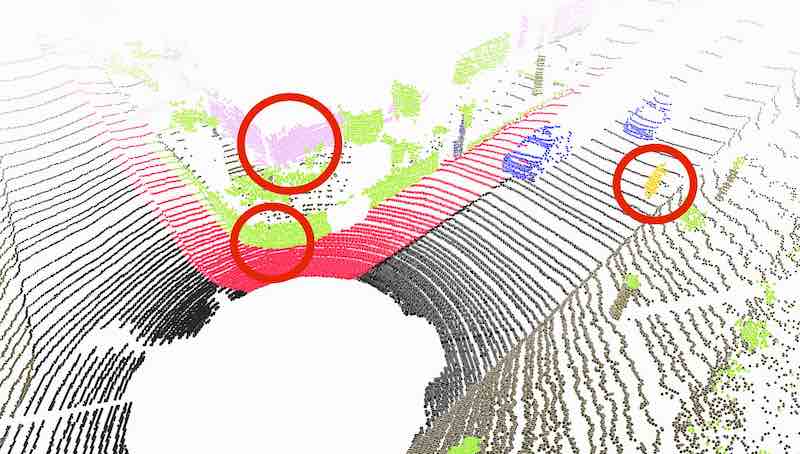}
        \end{overpic}\\
        \begin{overpic}[width=0.32\textwidth]{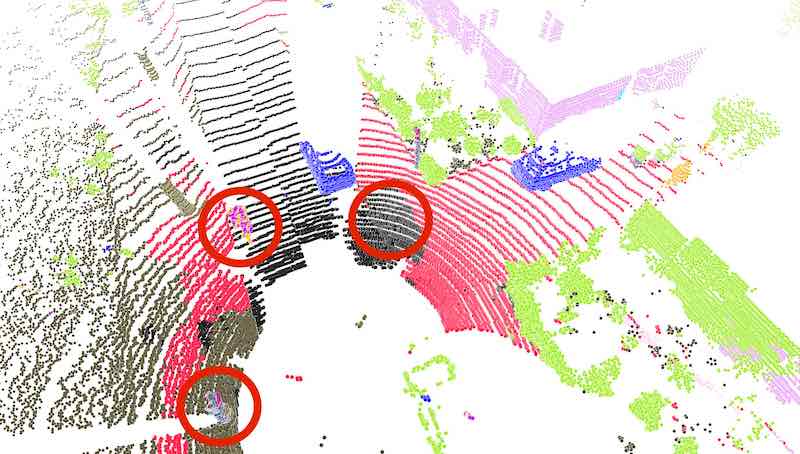}
        \end{overpic} &  
        \begin{overpic}[width=0.32\textwidth]{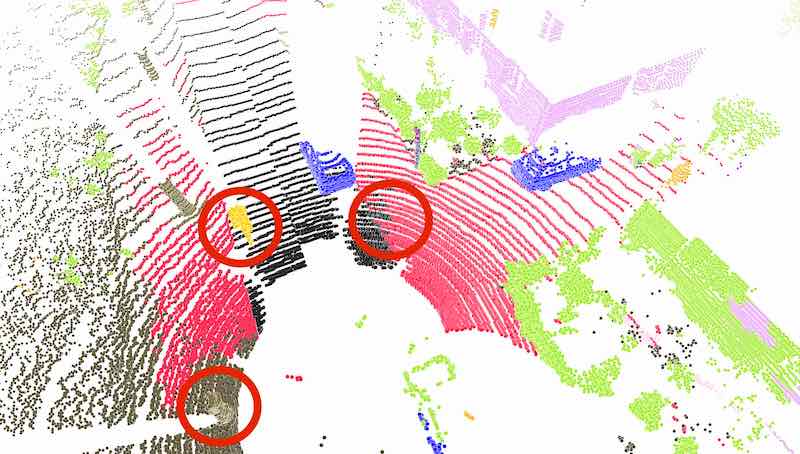}
        \end{overpic} &
        \begin{overpic}[width=0.32\textwidth]{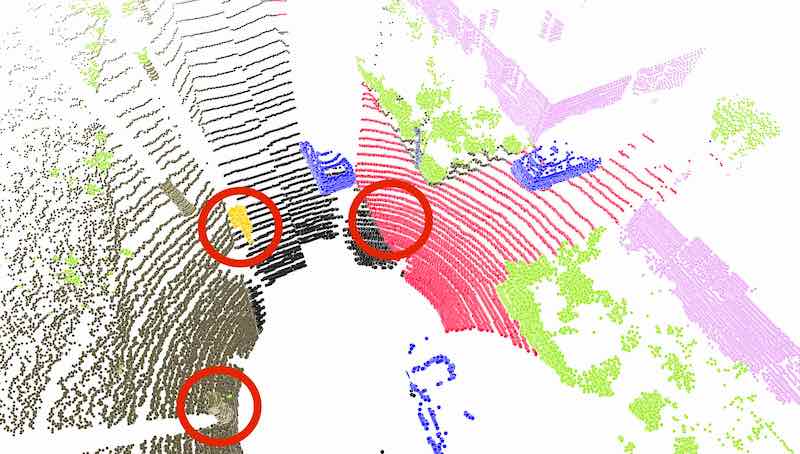}
        \end{overpic}\\
        \begin{overpic}[width=0.32\textwidth]{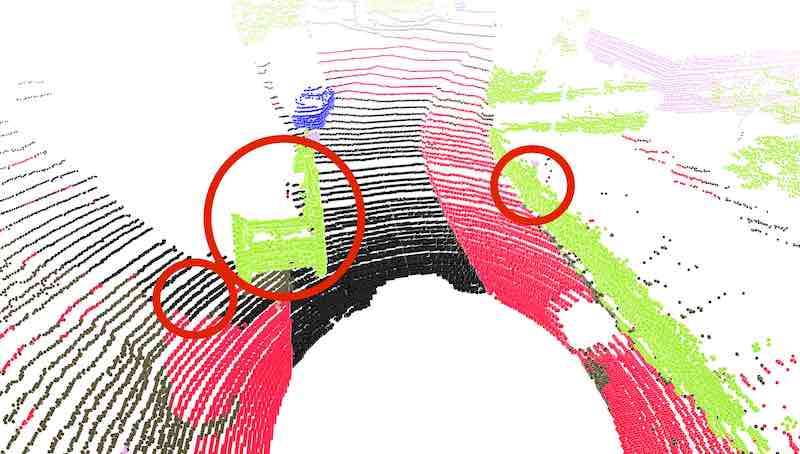}
        \end{overpic} &  
        \begin{overpic}[width=0.32\textwidth]{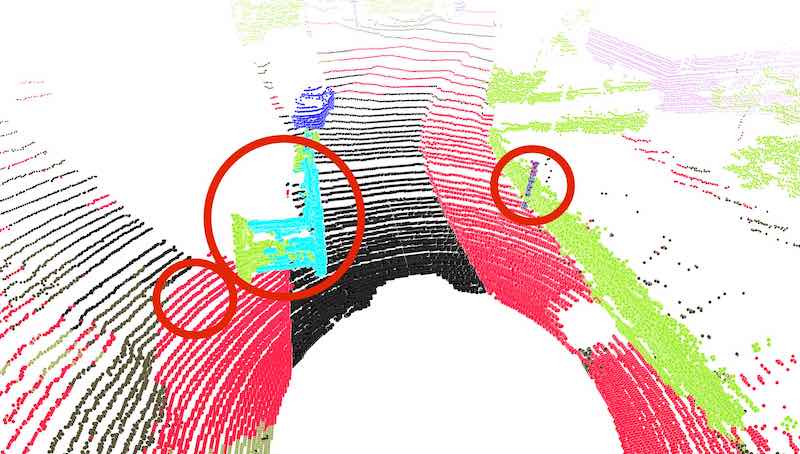}
        \end{overpic} &
        \begin{overpic}[width=0.32\textwidth]{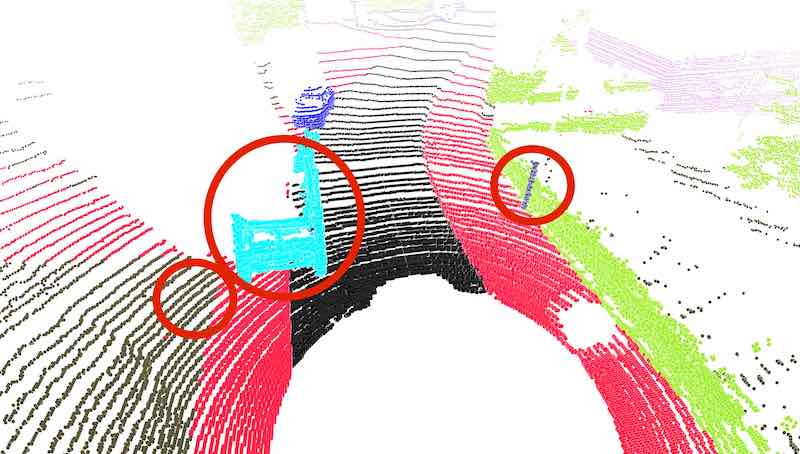}
        \end{overpic}\\
        \multicolumn{3}{c}{
        \begin{overpic}[width=0.99\textwidth]{images/qualitative_2/legend_semkitti.jpg}
        \end{overpic}}
    \end{tabular}
    \caption{Results on SynLiDAR$\rightarrow$SemanticKITTI. Source predictions are often wrong and mingled in the same region. After adaptation, \ourmethod improves the segmentation accuracy with homogeneous predictions and correctly assigned classes. The red circles highlight regions with interesting results.}
    \label{fig:qualitative_kitti_sup_1}
\end{figure}

Fig.~\ref{fig:qualitative_poss_sup_0}-\ref{fig:qualitative_poss_sup_1} show additional qualitative examples after adaptation on SynLiDAR $\rightarrow$ SemanticPOSS while Fig.~\ref{fig:qualitative_kitti_sup_0}-\ref{fig:qualitative_kitti_sup_1} show additional qualitative examples after adaptation on SynLiDAR$\rightarrow$SemanticKITTI. We report results before adaptation (source), after adaptation with CoSMix (ours) and add ground-truth labels (gt) for comparison.
We highlight regions with interesting results using red circles.
% However, we can observe that improvements are also visible across other regions of the scene.

% In SynLiDAR$\rightarrow$SemanticPOSS, improvements are visible in many regions covering all the frames. Source often fails in segmenting the classes \textit{car}, \textit{building}, \textit{vegetation} and small classes such as \textit{pole} and \textit{pedestrian}. After adaptation these classes are better segmented with a visible improvement.
% In SynLiDAR$\rightarrow$SemanticKITTI, visible improvements belongs to the class \textit{car}, \textit{truck}, \textit{building}, \textit{vegetation} and \textit{terrain}. Interestingly, source model confuses often \textit{road} with \textit{other-ground} and \textit{car} with \textit{vegetation}. After adaptation, these classes are better segmented.

\clearpage
% ---- Bibliography ----
%
% BibTeX users should specify bibliography style 'splncs04'.
% References will then be sorted and formatted in the correct style.
%
\bibliographystyle{splncs04}
\bibliography{egbib}